\documentclass{article}

\usepackage[accepted]{icml2026}

\usepackage{titletoc}
\usepackage{microtype}
\usepackage{graphicx}
\usepackage{subcaption}
\usepackage{booktabs}
\usepackage{amsmath}
\usepackage{amssymb}
\usepackage{mathtools}
\usepackage{amsthm}
\usepackage{caption}

\usepackage{url}            
\usepackage{booktabs}       
\usepackage{amsfonts}       
\usepackage{nicefrac}       
\usepackage{microtype}      
\usepackage{caption} 
\usepackage{soul}
\usepackage{amsmath}
\usepackage{amssymb}
\usepackage{graphicx} 
\usepackage[english]{babel}
\usepackage{amsthm}
\usepackage{mathrsfs}
\usepackage{multirow}
\usepackage{dsfont}
\usepackage{subcaption}
\usepackage{wrapfig}
\usepackage{etoolbox}
\usepackage{upgreek}
\usepackage{etoc}
\usepackage{tikz}

\PassOptionsToPackage{square,numbers,sort&compress}{natbib}
\usepackage[square,numbers,sort&compress]{natbib}

\definecolor{darkblue}{rgb}{0,0.08,0.45}
\usepackage[pagebackref=true,breaklinks=true,unicode=true,colorlinks=true,citecolor=darkblue,linkcolor=darkblue,urlcolor=darkblue]{hyperref}
\usepackage[capitalize,noabbrev]{cleveref}
\usepackage[textsize=tiny]{todonotes}

\PassOptionsToPackage{a4paper}{geometry}

\usepackage[utf8]{inputenc}
\usepackage[T1]{fontenc}    
\usepackage{xcolor}
\newcommand{\cX}{\mathcal{X}}
\newcommand{\cC}{\mathcal{C}}

\newcommand{\statk}{\pi_{k}}

\renewcommand{\mid}{\,|\,}

\newtheorem{theorem}{Theorem}
\newtheorem{prop}{Proposition}

\newtheorem{rmk}{Remark}
\newtheorem{corr}{Corollary}

\newtheorem{asm}[theorem]{Assumption}

\usetikzlibrary{
  shapes.geometric,
  arrows.meta,        
  positioning,        
  calc,              
  fit,                
  backgrounds,        
  decorations.pathreplacing
}
\usetikzlibrary{arrows.meta}
\definecolor{mylightred}{RGB}{245,193,192}
\definecolor{mylightblue}{RGB}{187,187,250}
\definecolor{mygray}{RGB}{227,227,227}
\definecolor{mypurple}{RGB}{171,110,145}

\usetikzlibrary{backgrounds}
\pgfdeclarelayer{bg}
\pgfsetlayers{bg,main}
\usetikzlibrary{positioning,calc,arrows.meta,shapes.geometric,decorations.pathreplacing,fit,backgrounds}
\pgfdeclarelayer{bg}
\pgfsetlayers{bg,main}

\icmltitlerunning{Multiple Choice Learning of Low-Rank Adapters for Language Modeling}

\begin{document}

\twocolumn[
  \icmltitle{Multiple Choice Learning of Low-Rank Adapters for Language Modeling}
  \icmlsetsymbol{equal}{*}

  \begin{icmlauthorlist}
    \icmlauthor{Victor Letzelter}{equal,telecom,valeo}
    \icmlauthor{Hugo Malard}{equal,telecom}

    \icmlauthor{Mathieu Fontaine}{telecom}
    \icmlauthor{Gaël Richard}{telecom}
    \icmlauthor{Slim Essid}{telecom}
    \icmlauthor{Andrei Bursuc}{valeo}
    \icmlauthor{Patrick Pérez}{kyutai}
  \end{icmlauthorlist}

  \icmlaffiliation{telecom}{LTCI, Télécom Paris, Institut Polytechnique de Paris}
  \icmlaffiliation{valeo}{Valeo.ai}
  \icmlaffiliation{kyutai}{Kyutai}

  \icmlcorrespondingauthor{Victor Letzelter}{letzelter.victor@hotmail.fr}
  \icmlkeywords{Winner-takes-all, Diversity, Ambiguity, Multiple Choice Learning, Language Modeling, Low Rank Adapters}

  \vskip 0.3in
]

\printAffiliationsAndNotice{\icmlEqualContribution}

\begin{abstract}
We propose $\texttt{LoRA-MCL}$, a training scheme that extends next-token prediction in language models with a method designed to decode diverse, plausible sentence continuations at inference time. Traditional language modeling is an intrinsically ill-posed problem: given a context, multiple ``futures'' may be equally plausible. Our approach leverages Multiple Choice Learning (MCL) and the winner-takes-all loss to efficiently handle ambiguity through Low-Rank Adaptation. We provide a theoretical interpretation of applying MCL to language modeling, assuming the data is generated from a mixture of distributions. We illustrate the proposed approach using mixtures of Markov chains. We then demonstrate with experiments on audio and visual captioning, as well as machine translation, that our method achieves high diversity and relevance in generated outputs. We release the code for applying \texttt{LoRA-MCL} to a wide range of language models.\footnote{\href{https://github.com/Victorletzelter/LoRA-MCL}{https://github.com/Victorletzelter/LoRA-MCL}}
\end{abstract}

\section{Introduction}

Predicting what a person will say next or describing the content of an audio or visual scene with text is difficult, if not impossible, to do with perfect accuracy. When the context is not informative enough, external factors may lead to different scenarios or \textit{modes} of plausible text continuations \citep{yang2018breaking, dieng2020topic, mei2022diverse}. In such ambiguous tasks, the conditional distribution over the space of output sentences given the input context may be multi-modal due to the underlying inherent uncertainty \citep{malinin2020uncertainty}.

Initially proposed for text processing, transformer-based auto-regressive language models \citep{radford2018improving, radford2019language} have quickly become a general framework for modeling streams of tokens, which can also represent, for instance, images and audio signals \citep{touvron2023llama, chu2024qwen2, liu2023visual}. Such models are trained as next-token predictors and allow, by nature, for addressing this uncertainty, by generating plausible output sentences through a wide body of maximum a posteriori (MAP) or sampling-based decoding approaches \citep{welleckdecoding}. While sampling allows exploration and diversity, it may lead to unreliable responses and requires truncation to avoid unexpected answers, partly due to the overestimation of low-probability tokens \citep{zhang2021trading, hewitt2022truncation}. When seeking reliable and expected answers, MAP estimation techniques, like Beam Search \citep{beamSearch}, look for sentences that maximize the model's likelihood.
However, these alternatives have drawbacks as they may lack diversity, be prone to repetition loops \citep{keskar2019ctrl}, and may sound unnatural \citep{holtzmancurious}.
Some approaches, e.g., Diverse Beam Search \citep{vijayakumar2018diverse}, were therefore proposed to artificially increase the diversity at inference, e.g., through a diversity penalty parameter $\lambda$, to find a tradeoff between generation quality and sample diversity \citep{sucontrastive}. In contrast with these methods, our approach aims to \textit{predict} diverse sentences reflecting the ambiguity of the input context.

Multiple Choice Learning (MCL) \citep{guzman2012multiple, lee2016stochastic} has emerged as a paradigm for addressing ambiguous tasks. It generally consists of a network with a shared backbone and multiple output heads. During training, it utilizes the winner-takes-all loss for adaptively updating the head that performs the best for each example. This is a competitive training scheme that specializes each model to subsets of the conditional output distribution \citep{rupprecht2017learning}. In this paper, we propose incorporating this idea for language model fine-tuning, leveraging multiple Low-Rank Adapters \citep{hu2022lora} instead of multiple heads, which may be impractical due to computational requirements and architectural constraints. Our method natively generates diverse and plausible sequences in a single forward pass, aiming to best approximate the conditional output distribution. Our main contributions are as follows.

\textbf{We propose a new paradigm that adapts MCL for token sequence modeling} with \texttt{LoRA-MCL}, that is particularly suited for efficient fine-tuning of language models.

\textbf{We provide a theoretical analysis of our approach.} Assuming the sequences are sampled from a mixture of distributions, we explain why \texttt{LoRA-MCL} should capture the data distribution modes, and validate the claims on Markov chains with a well-designed toy example.

\textbf{We conduct extensive experiments on audio and vision captioning, as well as machine translation}, showing wide applicability and an excellent diversity--quality trade-off.

\section{Problem setup}

 Let $x \triangleq (x_{t})_{t=1}^{T} \in \mathcal{V}^{T}$ be a sequence of $T$ tokens belonging to a finite vocabulary $\mathcal{V} = \{1,\dots,|\mathcal{V}|\}$, and $c \triangleq (c_{t})_{t=1}^{\tau} \in (\mathbb{R}^{d})^{\tau}$ be a sequence of $\tau$ context vector embeddings of dimension $d$.
 Language modeling aims at learning the law $p(x \mid c) = \prod_{t=1}^{T} p(x_t \mid x_{<t}, c)$ using a model $p_{\theta}$ with parameters $\theta$,
 by minimizing the following negative log-likelihood loss, which is equivalent to the maximum likelihood estimation (MLE):
 {%
\setlength{\abovedisplayskip}{4pt}%
 \setlength{\belowdisplayskip}{4pt}%
 \begin{equation}
 \mathcal{L}(\theta)
 = \mathbb{E}_{c,x}[- \sum_{t=1}^{T} \log p_{\theta}(x_t \mid x_{<t},c) ]\;,
 \label{eq:teacher_forcing}
 \end{equation}
 }% 
 where $x_{< t}$ denotes the sequence of tokens prior to time $t$. Optimizing \eqref{eq:teacher_forcing} is referred to as \textit{teacher-forcing} \citep{williams1989learning}, where $p_{\theta}$ is fed with target (instead of predicted) tokens $x_{<t}$ during training \citep{sutskever2014sequence}. When using a transformer architecture \citep{vaswani2017attention}, this is implemented via causal attention modules, which allow for computing the conditional distributions in parallel through all time steps within a single forward.

During inference, decoding methods \citep{welleckdecoding} proceed to generating sequences $\hat{x}$ from the trained model $p_{\theta}$ in an auto-regressive fashion. First, they start with a conditional distribution for the first token from the context: $p_{\theta}(x_1 \mid c)$, which allows selecting $\hat{x}_1$. Then, for $t \geq 2$ they predict $p_{\theta}(x_t \mid \hat{x}_{<t}, c)$, and select $\hat{x}_t$, until reaching either the sequence length limit or an end-of-sentence token. The choice of the decoding method to generate $K$ candidate sequences $\hat{x}^1, \dots, \hat{x}^K$ depends on the purpose of the task, but the general goal is $(i)$ to get highly likely sentences, i.e., ones that maximize $p_{\theta}(\hat{x} \mid c)$; and $(ii)$ to get diverse sentences, as can be measured by $n$-gram similarity \citep{ippolito2019comparison}. Although this is a widely adopted paradigm, we show next how this training and decoding pipeline can be improved: instead of \textit{artificially} generating diversity at inference time, we aim at \textit{learning to predict} sequences that cover well the modes of the target distribution $p(x \mid c)$.

\section{Methodology}
\label{sec:methodo}

\subsection{Motivation}
\label{sec:motivation}
In language modeling, topic models \citep{nigam2000text, blei2003latent, dieng2020topic} are data-generating processes in which the ground-truth probability distribution of sequences is modeled as a mixture of latent components or topics.
For example, the sequence “I am eating ...” may have multiple plausible continuations, but the likelihood of each depends heavily on contextual factors such as the speaker’s location, which influences their culinary habits. Each location (or context) can thus be associated with a distinct word distribution. In topic models, data generation proceeds by first sampling a topic $z \in \mathcal{Z}$ for each sentence (usually referred to as a \textit{document} in the literature), and then sampling words (or $n$-grams) from the distribution associated with that topic. 

With this in mind, MLE in \eqref{eq:teacher_forcing} may not be suitable \citep{yang2018breaking}. While MLE is effective for estimating the distribution $p(x)$, it does not capture the components when it is expressed as a mixture, i.e., $p(x) = \sum_{k} p(z_k) p(x \mid z_k)$. In such cases, MLE tends to model the aggregate rather than distinguish topic-specific distributions $p(x \mid z_k)$.

\subsection{Applying MCL to language modeling}
\label{sec:mcl_language}
Our approach is inspired by the multiple choice learning (MCL) literature \citep{guzman2012multiple, lee2016stochastic}. We propose the following training scheme, intending to enable the recovery of the different topics $z_k$. Instead of a single model, we consider a \textit{set} of models $(\theta_1, \dots, \theta_{K})$. Then the objective \eqref{eq:teacher_forcing} is replaced by one consisting of iterating between the following two steps:
\begin{enumerate}
    \item For each training sample $(c,x)$ in the batch: Compute $p(x \mid c; \theta_k)$ for $k \in \{1,\dots,K\}$, and choose the best model $k^{\star}(x,c) = \mathrm{argmax}_k \; p(x \mid c; \theta_k)$.
    \item Compute the winner-takes-all (WTA) loss as: \begin{equation}
    \label{eq:wta_loss}
        \mathcal{L}^{\mathrm{WTA}}(\theta_1, \dots, \theta_{K}) = - \mathbb{E}_{c,x}\Big[ \max_{k = 1,\dots,K} \log p(x \mid c ; \theta_k) \Big]
    \end{equation}
    where $\log p(x \mid c ; \theta_k) = \sum_{t=1}^{T} \log p(x_t \mid x_{<t},c ; \theta_k)$, and perform an optimization step.
\end{enumerate}

This training procedure, similar to a hard-EM style optimization \citep{min2019discrete, wen2023an}, is a 
competitive training scheme that encourages the different models to explore different areas of the data distribution. However, it is subject to two main issues: First, using $K$ models instead of a single one drastically increases the training time and memory cost, which may be intractable for large language models (LLMs). Second, the optimization may be subject to collapse, where the same models are chosen as winners through the iterations, leaving the other models untrained. In the next section, we describe how we solve these issues with our approach \texttt{LoRA-MCL}.

\subsection{\texttt{LoRA-MCL} method}
\label{sec:method_lora_mcl}
Multiple choice learning typically alleviates the high training cost issue of $K$ models by training a single model with several heads \citep{lee2016stochastic, lee2017confident}. However, we argue that such an approach is not well-suited for fine-tuning language models. First, the heads of most language models are quite large (for example, in Qwen2-Audio \citep{chu2024qwen2} the \texttt{lm\_head} has $d \times | \mathcal{V} | = 4096 \times 156032 \simeq 640 \mathrm{M}$ parameters), and standard MCL would not scale easily with the number of heads. 
Second, the initialization of the heads poses several challenges. Initializing each head with the parameters of the head of the pretrained model 
requires special care, as the collapse of the predictions is likely given the very similar hypotheses (same parameters). A complete re-initialization of the heads is detrimental to performance, as numerous training iterations would be necessary to reach the same level of knowledge as in the pretrained head. These effects are empirically verified in Apx.~\ref{apx:multi-head}. For these reasons, we consider a Low Rank Adapter (LoRA) approach \citep{hu2022lora} due to its excellent trade-off between performance and computational requirements, as well as its wide adoption in the context of large model fine-tuning. 

Let $\theta$ be the parameters of the pretrained base model. At each layer $\ell$ where LoRA is enabled, we use a family of adapters $(A^{k}_{\ell},B^{k}_{\ell}) \in \mathbb{R}^{d \times r} \!\times\! \mathbb{R}^{r \times d}$ for $k \in \{1,\dots,K\}$.
Let 
\begin{equation}
    \theta_k = \theta \cup \left\{ (A_{\ell}^{k}, B_{\ell}^{k}) \;|\; \ell \in \{1,\dots,L\} \right\}\;,
\end{equation}
be the set of parameters that are involved in hypothesis $k$, with $L$ being the total number of layers where LoRA is used. Training in the WTA fashion involves computing $p(x \mid c; \theta_k)$ for $k = 1,\dots,K$. To avoid situations where some heads may be under-trained, including the \textit{collapse} when a single head is trained, we use the relaxation of the winner-takes-all training objective of the form
\begin{equation}
    \mathcal{L}^{\mathrm{WTA}}(\theta_{1},\dots,\theta_{K}) = - \mathbb{E}_{c,x}\Big[  \sum_{k=1}^{K} q_k \log p(x \mid c ; \theta_k) \Big]\;.
\label{eq:relaxation_wta_loss}
\end{equation}
where $\{q_k\}$ is a set of positive coefficients that sum to 1. These coefficients assign higher weight to the winning head $q_{k^{\star}}$ while still providing nonzero gradient contributions to the other heads, thereby mitigating collapse.
We experimented with two relaxation techniques. First, \textit{Relaxed-WTA} \citep{rupprecht2017learning} where $q_{k^{\star}} = 1 - \varepsilon$ and $q_k = \frac{\varepsilon}{K-1}$ for $k \neq k^{\star}$, with $\varepsilon > 0$ a small constant. We also considered the \emph{annealed MCL} method \citep{amcl}, which introduces a temperature parameter $\uptau$:
\begin{equation}
    q_k(x,c ; \uptau) = \frac{p(x \mid c ; \theta_k)^{\frac{1}{\uptau}}}{Z_{x,c}(\uptau)}\,, \; Z_{x,c}(\uptau) = \sum_{s=1}^{K} p(x \mid c ; \theta_s)^{\frac{1}{\uptau}}\;.
    \label{eq:annealed_coeff}
\end{equation} 
Here the temperature $t \mapsto \uptau(t)$ follows a decreasing schedule, typically $\uptau(t) = \uptau(0)\rho^t$ with $\rho < 1$ and $\uptau(0) > 0$. At high temperatures, training is distributed more evenly across all hypotheses, preventing collapse; as $\uptau \to 0$, the method converges to the greedy WTA regime.

\begin{figure}
\centering
\hspace*{-0.3cm}
\begin{tikzpicture}[font=\small, node distance=6mm, xshift=-12cm]
\definecolor{OIblue}{HTML}{0072B2}
\definecolor{OIpurple}{HTML}{CC79A7}

\tikzset{
  vec/.style = {
  draw=white,
  fill=white,
  rounded corners=2pt,
  minimum width=20mm,
  minimum height=6mm
},
  W/.style         ={draw, fill=blue!15, rounded corners=2pt, minimum width=17.6mm, minimum height=16mm, opacity=0.9},
  loraDown/.style  ={draw, fill=red!15, trapezium, trapezium angle=70, shape border rotate=180,
                     minimum width=17.6mm, minimum height=4.8mm, line width=0.8pt, text width=5mm},
  loraUp/.style    ={draw, fill=red!15, trapezium, trapezium angle=70, minimum width=17.6mm, minimum height=4.8mm, line width=0.8pt, text width=5mm},
  loraDownWin/.style={draw=black, fill=red!20, trapezium, trapezium angle=70, shape border rotate=180, minimum width=17.6mm, minimum height=4.8mm, line width=1.4pt, text width=5mm},
  loraUpWin/.style  ={draw=black, fill=red!20, trapezium, trapezium angle=70,minimum width=17.6mm, minimum height=4.8mm, line width=1.4pt, text width=5mm},
  bwdWin/.style    ={-{Latex[length=3.5mm,width=2.3mm]}, very thick, OIpurple!90!black},
  bwdLose/.style   ={-{Latex[length=2.5mm,width=1.5mm]}, thick,OIpurple!55!white},
fwdArrow/.style    ={-{Latex[length=2.4mm,width=1.5mm]}, line width=0.9pt, draw=gray!55, opacity=0.35},
  frozenBus/.style ={line width=0.6pt, gray!70},
  mergepoint/.style={circle, draw, fill=white, inner sep=0.5pt, minimum size=3mm, line width=0.5pt, font=\footnotesize},
  context/.style   ={draw, dashed, gray!60, rounded corners=3pt, inner sep=5pt},
  note/.style      ={font=\scriptsize, text=gray!60},
  ellipsis/.style  ={font=\normalsize, text=gray!70},
  cond/.style      ={inner sep=0pt, outer sep=0pt}
}

\tikzset{
  perspTip/.tip    = {Latex[round, length=2.4mm, width=1.2mm]},
}
\tikzset{
  halo/.style={preaction={draw, line width=6pt, white}},
}
\def\loraSep{7mm}      
\def\hspaceLora{20mm}    
\def\gapBaseA{12mm} 
\def\xDrop{11mm} 
\def\mergeShiftC{-1.5mm}
\def\mergeShiftR{-3mm}

\coordinate (levelA) at (0,14mm);          
\coordinate (levelB) at ($(levelA)+(0,\loraSep)$);
\coordinate (bneck)  at ($(levelA)!0.5!(levelB)$);

\coordinate (colC) at (0,0 |- levelA);
\coordinate (colL) at ($ (colC) + (-\hspaceLora,0) $);
\coordinate (colR) at ($ (colC) + (\hspaceLora,0) $); 

\node[loraUp]      (Aleft)   at (colL) {\large $A_{\ell}^{1}$};
\node[loraDown]    (Bleft)   at ($(Aleft.north)+(0,\loraSep)$) {\large $B_{\ell}^{1}$};

\node[loraUpWin]   (Acenter) at (colC) {\large $A_{\ell}^{k^{\star}}$};
\node[loraDownWin] (Bcenter) at ($(Acenter.north)+(0,\loraSep)$) {\large $B_{\ell}^{k^{\star}}$};

\node[loraUp]      (Aright)  at (colR) {\large $A_{\ell}^{K}$};
\node[loraDown]    (Bright)  at ($(Aright.north)+(0,\loraSep)$) {\large $B_{\ell}^{K}$};

\node[W, anchor=east] (Wl) at ($(Aleft)!0.5!(Bleft)+(-\gapBaseA,0)$) {\large $W_{\ell}$};

\node[cond] (xL) at ($(colL)+(0,-\xDrop)$) {$h^{(1)}$};
\node[cond] (xC) at ($(colC)+(0,-\xDrop)$) {$h^{(k^{\star})}$};
\node[cond] (xR) at ($(colR)+(0,-\xDrop)$) {$h^{(K)}$};
\node[cond, below=8.5mm of xC] (xPrev) {$x$};

\node[mergepoint] (mC) at ($(Bcenter.north)+(0,7mm)$) {$+$};
\node[mergepoint] (mL) at ($(Bleft.north |- Bcenter.north)+(0,7mm)$) {$+$};
\node[mergepoint] (mR) at ($(Bright.north)+(0,7mm)$) {$+$};

\node[ellipsis, above=3.5mm of mL, yshift=2.5mm]     (dotsL) {$\vdots$};
\node[ellipsis, above=3.5mm of mC, yshift=2.5mm]     (dotsC) {$\vdots$};
\node[ellipsis, above=3.5mm of mR, yshift=2.5mm]     (dotsR) {$\vdots$};

\node[cond, above=-2mm of dotsL, anchor=south] (p1)   {$p(x \mid \theta_1)$};
\node[cond, above=-2mm of dotsC, anchor=south] (pmid) {\textbf{$p(x \mid \theta_{k^{\star}})$}};
\node[cond, above=-2mm of dotsR, anchor=south] (pK)   {$p(x \mid \theta_{K})$};

\node[cond, above=8mm of pmid, anchor=south] (loss) {$\mathcal{L}^{\mathrm{WTA}}(\theta_1,\dots,\theta_K)$};

\coordinate (WinL) at ($(Wl.south west)!0.2!(Wl.south east)$);
\coordinate (WinC) at ($(Wl.south west)!0.5!(Wl.south east)$);
\coordinate (WinR) at ($(Wl.south west)!0.8!(Wl.south east)$);

\coordinate (OutL) at ($(Wl.north west)!0.2!(Wl.north east)$);
\coordinate (OutC) at ($(Wl.north west)!0.5!(Wl.north east)$);
\coordinate (OutR) at ($(Wl.north west)!0.8!(Wl.north east)$);

\begin{pgfonlayer}{bg}
\coordinate (xLdot) at ($(xL.south) + (0,-0.4cm)$);
\coordinate (xCdot) at ($(xC.south) + (0,-0.4cm)$);
\coordinate (xRdot) at ($(xR.south) + (0,-0.4cm)$);

\node[ellipsis, below=-2.4mm of xL] (dotsXL) {$\vdots$};
\node[ellipsis, below=-2.4mm of xC] (dotsXC) {$\vdots$};
\node[ellipsis, below=-2.4mm of xR] (dotsXR) {$\vdots$};

\draw[fwdArrow, -{Latex[length=1.9mm,width=1.5mm]}]
  (xPrev.north) to[out=90, in=-90, looseness=0.9] (xLdot);
\draw[fwdArrow, -{Latex[length=1.9mm,width=1.5mm]}]
  (xPrev.north) to[out=90, in=-90, looseness=0.9] (xCdot);
\draw[fwdArrow, -{Latex[length=1.9mm,width=1.5mm]}]
  (xPrev.north) to[out=90, in=-90, looseness=0.9] (xRdot);

\draw[fwdArrow, -{perspTip}] ($ (xL.north) + (0,0) $) -- (WinL);
\draw[fwdArrow, -{perspTip}] ($ (xC.north) + (0,0) $) -- (WinC);
\draw[fwdArrow, -{perspTip}] ($ (xR.north) + (0,0) $) -- (WinR);
\draw[fwdArrow,    -{perspTip},halo]    (OutL) -- (mL.west);
\draw[fwdArrow, -{perspTip},halo] (OutC) -- (mC.west);
\draw[fwdArrow,    -{perspTip},halo]    (OutR) -- (mR.west);

    \draw[fwdArrow, halo]    (xL.north) -- (Aleft.south);
    \draw[fwdArrow, halo] (xC.north) -- (Acenter.south);
    \draw[fwdArrow, halo]    (xR.north) -- (Aright.south);

  \begin{scope}[transform canvas={xshift=-0.2cm}]
    \draw[fwdArrow, halo]    (Bleft.north)   -- (mL.south);
    \draw[fwdArrow, halo] (Bcenter.north) -- (mC.south);
    \draw[fwdArrow, halo]    (Bright.north)  -- (mR.south);
    \draw[fwdArrow, halo]    (mL.north) -- (dotsL.south);
    \draw[fwdArrow, halo] (mC.north) -- (dotsC.south);
    \draw[fwdArrow, halo]    (mR.north) -- (dotsR.south);
    \draw[fwdArrow]    (p1.north)   to[out=90,in=210] (loss.south);
    \draw[fwdArrow] (pmid.north) to[out=90,in=270] (loss.south);
    \draw[fwdArrow]    (pK.north)   to[out=90,in=330] (loss.south);
  \end{scope}
\end{pgfonlayer}

\draw[bwdLose] (loss.south) to[out=-110,in=90] (p1.north);
\draw[bwdWin]  (loss.south) to[out=-90, in=90] (pmid.north);
\draw[bwdLose] (loss.south) to[out=-70, in=90] (pK.north);

\draw[bwdLose, halo] (dotsL.south) -- (mL.north);
\draw[bwdWin, halo]  (dotsC.south) -- (mC.north);
\draw[bwdLose, halo] (dotsR.south) -- (mR.north);

\draw[bwdLose, halo] (mL.south) -- (Bleft.north);
\draw[bwdWin, halo]  (mC.south) -- (Bcenter.north);
\draw[bwdLose, halo] (mR.south) -- (Bright.north);

\coordinate (leftPad) at ($(Aleft.west) + (0mm,-7mm)$);
\coordinate (rightPad) at ($(Bright.east) + (3mm,18mm)$);
\node[context, fit=(Wl) (leftPad) (rightPad)] (ctx) {};
\node[anchor=south west, inner sep=3pt, xshift=-1.5mm, yshift=50mm] (legend) at (ctx.south west) {
  \begin{tikzpicture}[x=1cm,y=0.6cm, font=\scriptsize]
    \draw[bwdWin] (0,0) -- ++(0.8,0) node[right=2pt, black] {winning gradient};
    \draw[bwdLose] (0,-0.5) -- ++(0.8,0) node[right=2pt, black] {non-winning gradient};
    \draw[fwdArrow] (0,-1.0) -- ++(0.8,0) node[right=2pt, black] {forward pass};
  \end{tikzpicture}
};
\node[note, above=0mm of ctx.north, yshift=-1.0mm, xshift=-3.5cm] {\footnotesize layer $\ell$};
\end{tikzpicture}
\vspace{-10pt}
\caption{
\textbf{\texttt{LoRA-MCL}}. Components of a linear layer $\ell$ where LoRA is enabled, with context $c$ omitted. Frozen base weights $W_{\ell}$ are in \textcolor{mylightblue}{blue}; trainable LoRA adapters in \textcolor{mylightred}{light red}. The forward pass (in \textcolor{gray}{gray}) is computed independently for each hypothesis, where $h^{(1)}, \dots, h^{(K)}$ denote the hidden states as in \eqref{eq:group_lora}. Gradients (\textcolor{mypurple}{purple} arrows) are stronger for the winning hypothesis ($k^{\star}$).
}
    \label{fig:tikz_example}
\end{figure}

\subsection{Accelerating \texttt{LoRA-MCL} training with parallelization over the hypotheses}

A naive implementation of \texttt{LoRA-MCL} would require looping over the $K$ hypotheses in the batch to evaluate each candidate separately, which would slow down training. To avoid this, we process all hypotheses in parallel. Specifically, given an input sequence, we duplicate it $K$ times along the batch dimension. Each copy is then passed through a LoRA-adapted transformer, but crucially, we implement this in a \emph{grouped fashion}: instead of running $K$ independent forward passes, we combine them into a single batched operation where each group corresponds to one hypothesis. In practice, this can be achieved using a grouped 1D convolution (\texttt{nn.Conv1d} in PyTorch) with $K$ groups, so that each hypothesis uses its own LoRA weights while still sharing the frozen base model. This trick effectively multiplies the batch size by $K$ while keeping the memory overhead manageable (since the LoRA rank $r \ll d$). It removes the sequential loop, enabling parallelizing the forward of all hypotheses. We provide in Apx.~\ref{apx:acceleration} a quantitative comparison of the computational cost of \texttt{LoRA-MLE}, \texttt{LoRA-MCL}, and its grouped variant across varying numbers of hypotheses and training scenarios.

\section{Theoretical Analysis}

We justify the use of MCL for language modeling by assuming the sequence distribution is a mixture. Section~\ref{sec:hard_em} links our method to the Expectation Maximization (EM) algorithm \cite{dempster1977maximum} and derives lower and upper bounds on the optimal achievable test loss. We then apply this analysis to the case of Markov chains and simulate the method’s dynamics in a controlled setting. 

\subsection{Training dynamics and optimality conditions}
\label{sec:hard_em}
For the next-token prediction loss in \eqref{eq:teacher_forcing}, one can show that:
$ \mathrm{min}_{\theta} \; \mathcal{L}(\theta) = \mathcal{H}(x \mid c)$, where $\mathcal{H}(x \mid c) \triangleq - \mathbb{E}_{c,x} [ \mathrm{log} \; p(x \mid c)]$ is the entropy of $p$ \citep{mackay2003information}. Following the rationale of Sec.~\ref{sec:motivation}, let us now assume the data distribution can be written as a mixture. In the case of the WTA loss, we have the following proposition.

\begin{prop}[Proof in Apx.~\ref{sec:proof_prop:em}] Assume a data-generating process $p(x\,|\,c) = \sum_{k = 1}^{K} p(z_k\,|\,c) \; p(x\,|\,z_k, c)$ (Asm.\,\ref{asm:mixture}), perfect model expressiveness (Asm.\,\ref{asm:expressiveness}), and large enough batch size to approximate the true risk (Asm.\,\ref{asm:true_risk}). Then:
     \begin{itemize}
    \item[(i)] 
    \texttt{LoRA-MCL} acts as a conditional form of the hard-EM.
    \item[(ii)] Assuming disjoint components for the data mixture (Asm.~\ref{asm:disjoint}), and assuming $p(x \mid z_k, c) = p(x \mid c ; \theta_k)$ for each $k$, then 
    \begin{equation}
    \label{eq:opt}
    \mathcal{L}^{\mathrm{WTA}}(\theta) = \mathcal{H}(x \mid c,z) \,,\end{equation}
    where $\mathcal{H}(x \mid c,z) \triangleq \mathbb{E}_{c} \left[ \sum_{k=1}^{K} p(z_k \mid c) \mathcal{H}\big(x \mid c, z_k\big) \right]$ is the conditional entropy given the random variable $z$.
    \item[(iii)] We have the following inequalities:
    \begin{equation}
\label{eq:inequality1}
\begin{split}
\min_{\theta}\, \mathcal{L}(\theta) - \log K
\overset{(a)}{\leq} \min_{\theta}\, \mathcal{L}^{\mathrm{WTA}}(\theta) \\
\overset{(b)}{\leq} \mathcal{H}(x \mid c,z)
\overset{(c)}{\leq} \min_{\theta}\, \mathcal{L}(\theta)\;.
\end{split}
\end{equation}
    where $\mathrm{min}_{\theta} \; \mathcal{L}(\theta) = \mathcal{H}(x \mid c)$.
    \end{itemize}
    \label{prop:em} 
\end{prop}
$(i)$ in Prop.~\ref{prop:em} describes the relationship between \texttt{LoRA-MCL} and the hard-EM algorithm. $(ii)$ provides an expression for the WTA loss as a conditional entropy, $\mathcal{H}(x \mid c,z)$, assuming a perfect matching between the hypotheses and the modes. $(iii)$ establishes both a lower and an upper bound on the optimal achievable loss for \texttt{LoRA-MCL}, given in $(a)$ and $(b)$, respectively. Note that the gap between these bounds is
$\mathcal{H}(x \mid c,z) -  \mathrm{min}_{\theta} \, \mathcal{L}(\theta) + \mathrm{log}\,K = - \mathbb{E}_{c} [\mathcal{I}(x, z \mid c)] + \mathrm{log} \, K$ where $\mathcal{I}$ denotes the mutual information. Finally, $(c)$ shows in particular that $\mathrm{min}_{\theta} \, \mathcal{L}^{\mathrm{WTA}}(\theta) \leq \mathrm{min}_{\theta} \, \mathcal{L}(\theta)$.

\subsection{Case of Markov chains}
\label{sec:markov}

To make the analysis more concrete, we consider the case where the sequence of tokens is generated from Markov chains. Given a finite state space $\mathcal{V}$, a homogeneous Markov chain is defined by an initial distribution $\pi$ over $\mathcal{V}$ and a transition matrix $P \in [0,1]^{\mathcal{V} \times \mathcal{V}}$. A sequence $(x_1, \dots, x_T)$ is sampled from the Markov chain if $x_1 \sim \pi$ and, for each $t \geq 1$, $x_{t+1} \mid x_t \sim P_{x_t, \cdot}$, where $P_{i,j} = p(x_{t+1} = j \mid x_t = i)$. In the following, we ignore the initial warm-up phase, and we assume that $\pi$ is the stationary distribution of $P$. In this case, we will denote $x \sim \mathrm{MC}(P)$.

While the study of the training dynamics of transformers on Markov Chain data has been investigated in previous works \citep{edelman2024evolution, rajaraman2024analysis, makkuva2024attention, zekri2024large}, our setup instead considers a \textbf{mixture} of Markov chains \citep{gupta2016mixtures, kausik2023learning}. Assuming a uniform mixture, the data generating process is $x \sim \frac{1}{K} \sum_{k=1}^{K} \mathrm{MC}(P_{k})$; \begin{equation}
\label{eq:data_generating_process}
    k \sim \mathcal{U}(1,\dots,K),\quad x \mid z_k \sim \mathrm{MC}(P_{k})\;.
\end{equation}
When training a language model on such sequences, we have the following Corollary from Prop.\,\ref{prop:em}, where the context $c$ is ignored for simplicity.

\begin{corr} [Proof in Apx.~\ref{apx:proof_mc}] Assume the data-generating process is a uniform mixture of first-order Markov chains. Let $\hat{P}(\theta) \triangleq (p_{\theta}(x_{t+1} = j \mid x_{t} = i))_{i,j}$ be the predicted transition matrix. Under the same Asm. as in Prop.~\ref{prop:em}:
     \begin{itemize}
     \item[(i)] When the MLE estimator trained with \eqref{eq:teacher_forcing} reaches its optimal loss, we have
     \begin{equation}
         \hat{P}(\theta)_{i,j} 
     = \frac{1}{\sum_{s=1}^{K} (\pi_s)_i} \sum_{k=1}^{K} (\pi_k)_i (P_k)_{i,j}\;,
     \label{eq:stationary}
     \end{equation}
     where $\statk \in [0,1]^{\mathcal{V}}$ is the stationary distribution of $P_k$.
    \item[(ii)] The inequalities \eqref{eq:inequality1} hold, where $\mathcal{H}(x \mid z)$ is a weighted average of the entropy rate of each Markov Chain.
    \end{itemize}
    \label{corr:mc}
\end{corr}
The entropy $\mathcal{H}(x)$, which is the optimal loss of the MLE baseline, can be computed either exactly for short sequences or approximated, e.g., through Monte-Carlo integration. Our analysis considers first-order Markov chains, but we expect the results to extend to higher orders (see Apx.~\ref{apx:proof_mc}).

\subsection{Illustration with synthetic data}
\label{sec:synt_data}
\begin{figure*}[t]
    \centering
    \includegraphics[width=0.95\linewidth]{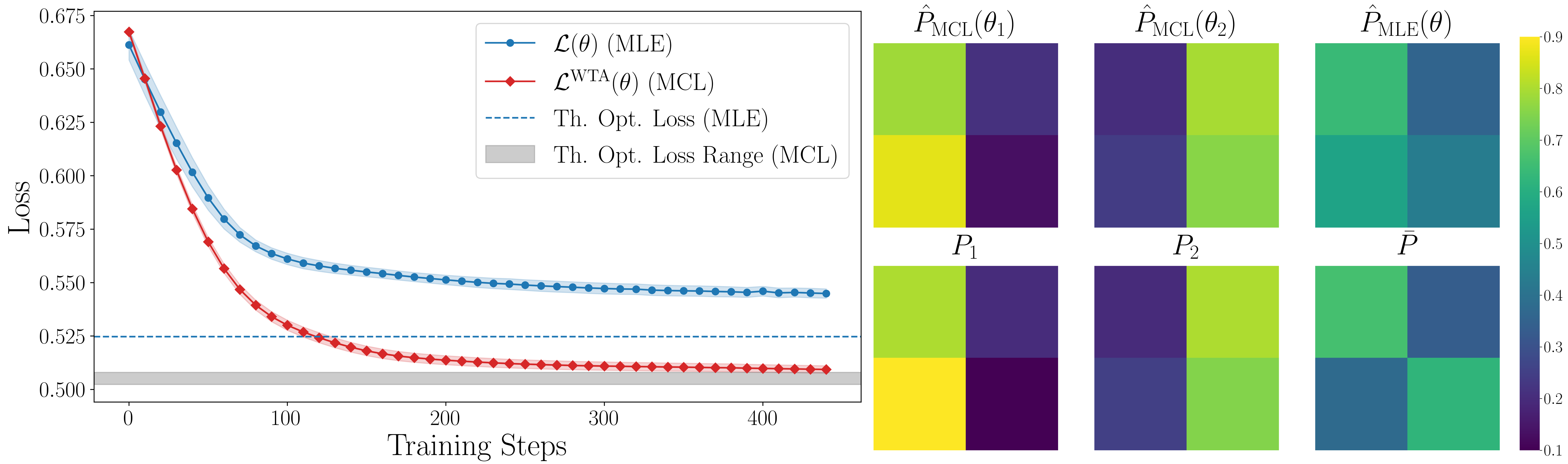}
    \caption{\textbf{Comparison of MCL with MLE.} 
    \textit{(Left)} Validation loss over training steps (averaged across three seeds) for \texttt{LoRA-MLE} (blue) and \texttt{LoRA-MCL} (red). The theoretical optimal MLE loss is the entropy $\mathcal{H}(x)$. The gray shaded region represents the bounds of the theoretical optimal MCL loss, as given by $(a)$ and $(b)$ in \eqref{eq:inequality1}. \textit{(Right)} Learned transition matrices \textit{(top)} versus references \textit{(bottom)}. MLE converges approximately toward the weighted average $\bar{P}$ (right-hand side of \eqref{eq:stationary}), whereas \texttt{LoRA-MCL} recovers the two modes.}
    \label{fig:illustration_markov}
\end{figure*}

To illustrate our approach, let us evaluate our algorithm on a synthetic dataset, for which results are given in Fig.~\ref{fig:illustration_markov}. We used $\mathcal{V}=\{1,2\}$, and $(P_{1},P_{2}) = (P(p_1,q_1),P(p_{2},q_{2}))$, with $P(p,q) \triangleq \left[\begin{array}{cc}
1-p & p \\
q & 1-q
\end{array}\right]\;$ and $p, q \in [0,1]$. We sampled data from a mixture of two Markov chains following \eqref{eq:data_generating_process}: we generated a sequence by sampling first $P_k$ with $k \sim \mathcal{U}\{1,2\}$. Once $P_k$ was set, we sampled the initial state uniformly, then we sampled the Markov chain according to the transition matrix until reaching the maximum sequence length ($T = 32$ here).

We considered a GPT-2-like architecture \citep{radford2019language, black2021gpt} using local-attention suggested by \citet{makkuva2024attention} to improve convergence on Markov chain data. We then trained the model with one and two hypotheses, using LoRA adapters as described above, one hypothesis corresponding to vanilla MLE. Training details are provided in Apx.~\ref{apx:synt_data}. Figures \ref{fig:illustration_markov} and \ref{fig2:illustration_markov} show both the evolution of the losses along training and the predicted transition matrix from the trained models. We see that the optimum is close to being global in this setup. While \texttt{LoRA-MCL} can capture the two modes of the mixture, we see that MLE tends to predict the weighted average of the transition matrices given by \eqref{eq:stationary}, which is consistent with Prop.\,\ref{corr:mc}.

\vspace{-0.5em}

\section{Empirical evaluation}

We evaluate \texttt{LoRA-MCL} on realistic datasets and large-scale models for audio and image captioning tasks, as well as machine translation. Predicting a textual description for images or audio signals is an ill-posed problem: from an input image or audio clip, multiple descriptions may be plausible; this is a real-world case where the conditional output distribution is inherently multi-modal. Similarly, Machine Translation is a one-to-many problem \cite{ott2018analyzing}. We demonstrate that \texttt{LoRA-MCL} provides a competitive approach for capturing these distributions when fine-tuning either audio, vision-language, or language-only models. We describe the experimental setup. See Apx.~\ref{apx:exp_details_audio} for details.

\subsection{Experimental setup in captioning}

\begin{figure*}[!ht]
    \centering
    \includegraphics[width=0.9\linewidth]{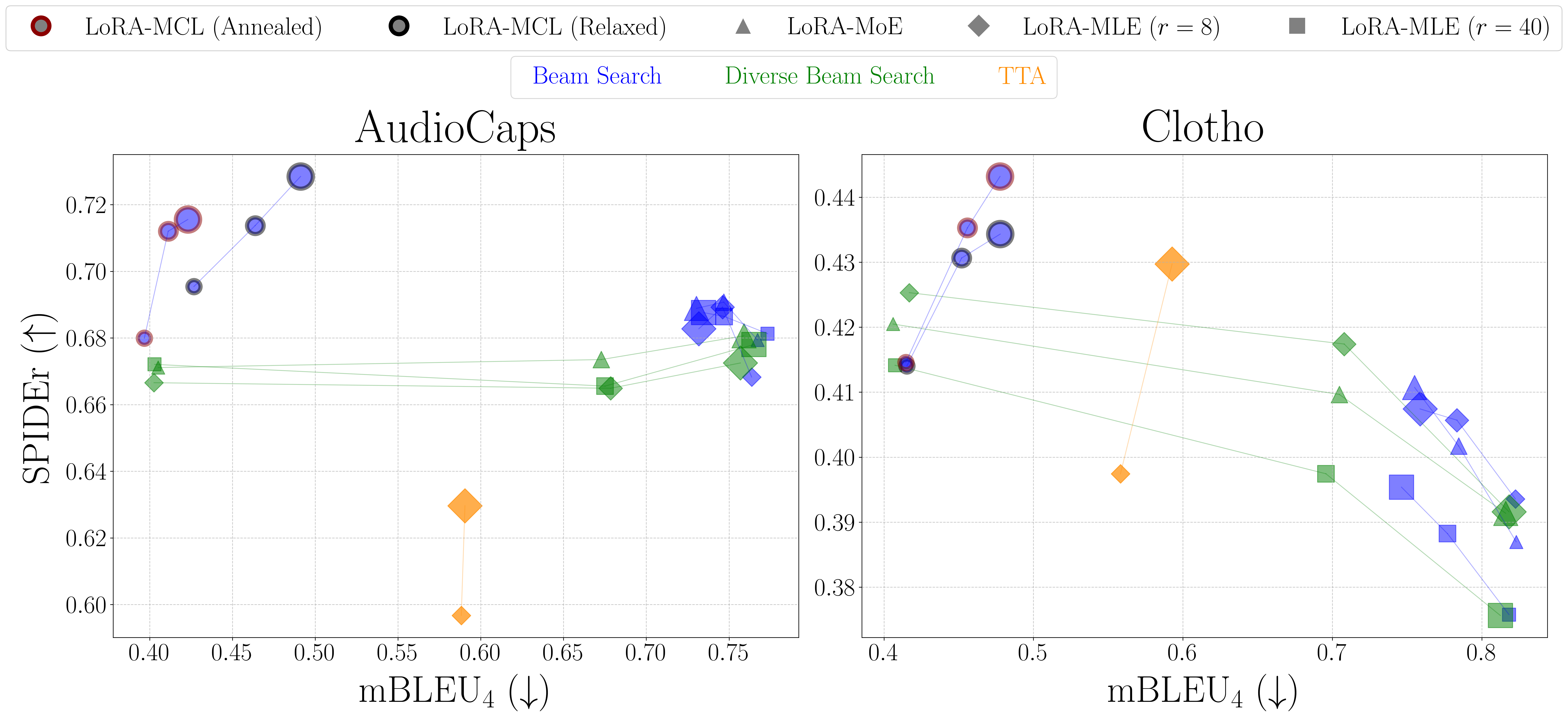}
    \caption{
    \textbf{Quality–diversity trade-off on audio captioning (5 candidates).} SPIDEr~$(\uparrow)$ for quality, mBLEU-4~$(\downarrow)$ for diversity. Marker shape stands for the method, color for the decoding method, and size is proportional to forward passes per example at inference. \texttt{LoRA-MLE} uses $r\in\{8,8K\}$ for parameter parity. \texttt{LoRA-MCL} uses circle markers: Relaxed (black edge) and Annealed (red edge).
    }
    \label{fig:quality-vs-diversity}
\end{figure*}

\textbf{Datasets.} We experimented on both Clotho-V2 \citep{font2013freesound, drossos2020clotho}, and AudioCaps \citep{gemmeke2017audio, kim2019audiocaps} datasets for the audio captioning task, while we make use of the TextCaps dataset \citep{sidorov2020textcaps} for the task of image captioning with reading comprehension. Table \ref{tab:audio_data_stat} describes the dataset sizes.

\textbf{Experimental details in audio.} We used the instructed Qwen2-Audio \citep{chu2024qwen2} as the base model, which has ${\sim} 8.4$ billion parameters and a vocabulary size $|\mathcal{V}| = 156{,}032$. We used LoRA adapters applied to the $Q, K, V$ linear projections of the attention modules, and the upside and downside projections of the feedforward blocks, across all layers. We used a rank $r$ and scaling factor $\alpha$, with $r = \alpha = 8$ unless otherwise stated. We trained for 1 and 10 epochs on AudioCaps and Clotho, respectively.

\textbf{Experimental details with visual data.} We used LLaVA 1.6 \citep{liu2023visual}, as the base model which features ${\sim} 7.1$ billion parameters and a vocabulary size $|\mathcal{V}| = 32{,}000$. We applied LoRA adapters only for the LLM decoder following \citet{zhou2024empirical}. The adapters were applied to $Q, K, V$, upside and downside projections as in Qwen2-Audio, and we used $r = \frac{\alpha}{4} = 8$ unless otherwise stated. Training was done over 1 epoch (without validation data), and the validation set of TextCaps was used for evaluation.

\textbf{Metrics.} We evaluate both quality and diversity. For quality, we report test-loss and standard Natural Language Generation metrics (BLEU, ROUGE, METEOR) \citep{papineni2002bleu, lin2004rouge, banerjee2005meteor}, and captioning-specific scores (CIDEr, SPICE, and SPIDEr) \citep{vedantam2015cider, anderson2016spice, liu2017improved}, which better correlate with human judgments in captioning. We also use Sentence-BERT \citep{reimers2019sentence}. We consider sentence-based oracle evaluation for these metrics (see \citet{lee2016stochastic, labbe2022my}). For diversity, we used mBLEU-4 \citep{mei2022diverse}, measuring similarity across generated captions \citep{zhu2018texygen}.

\textbf{Baselines.} We compare \texttt{LoRA-MCL} against the MLE baseline (\texttt{LoRA-MLE}) trained using \eqref{eq:teacher_forcing}, under the same conditions as the ones considered for our multi-hypothesis model. Specifically, both models use the same LoRA configuration, the same number of trainable parameters (the LoRA rank for the baseline is $K\times$ larger to this end), and the same number of iterations. We also considered a Mixture of Low Rank Experts \cite{muqeethsoft, wumixture, li2024mixlora} (\texttt{LoRA-MoE}) as a baseline. See Apx.~\ref{secapp:training_methods} for more details on the training methods. At inference time, for each decoding method applied to the baseline that returns $K$ sentences, we decode the same number of candidates with \texttt{LoRA-MCL}. When evaluating MAP methods such as greedy, beam search (BS) \citep{beamSearch}, and diverse beam search (DBS) \citep{vijayakumar2018diverse}, we ensure a consistent computational budget by aligning the number of forward passes. In this case, if \texttt{LoRA-MLE} or \texttt{LoRA-MoE} uses a beam size of $B$, our model uses a beam size of $\frac{B}{K}$ per hypothesis. Finally, we experimented with Test-Time Augmentation (TTA) \cite{wang2019aleatoric} with \texttt{LoRA-MLE} in Audio Captioning, applying SpecAugment \cite{park2019specaugment} $K$ times to the input Spectrogram to expect diverse outputs. We refer to Apx.~\ref{apx:decoding} for more details on all decoding methods.

\subsection{Audio captioning}

\label{sec:audio_captionin}

\textbf{Quality vs. diversity trade-off.} 
Quality–diversity performance is shown in Fig.~\ref{fig:quality-vs-diversity}. For readability, only the TTA runs (orange diamonds) with optimal augmentation strength on the quality–diversity are displayed in Fig.~\ref{fig:quality-vs-diversity}. It is strong on Clotho, but performs poorly on AudioCaps. We notice that a well-chosen value of $\lambda$ ($\lambda = 1.0$ in Fig.~\ref{fig:quality-vs-diversity}) allows DBS applied to \texttt{LoRA-MLE} and \texttt{LoRA-MoE} to be competitive. \texttt{LoRA-MCL} (circles) achieves the best trade-off between quality and diversity, appearing in the top-left corner of the plot, where the best relaxation technique (annealed or relaxed) depends on the setup. Although increasing the beam size generally improves performance for standard beam search, we observe that increasing the beam size within each group in DBS can negatively impact DBS, as observed in Clotho. Additional results and comparisons are provided in Apx.~\ref{sec:additional_results}.

\begin{table}[t]
    \centering
    \caption{\textbf{Test Loss $(\downarrow)$ as a function of $K$.} \texttt{LoRA-MCL} is trained for $K \in \{3, 5, 7\}$ using $\varepsilon=0.05$ and $r = 8$.}
    \scalebox{0.75}{
    \begin{tabular}{llcc}
\toprule
Training & $K$ & AudioCaps & Clotho \\
\midrule
\texttt{LoRA-MLE} ($r = 8$) & 1 & 2.203 & 2.812 \\
\texttt{LoRA-MLE} ($r = 8 \times 3$) & 1 & 2.195 & 2.868 \\
\texttt{LoRA-MLE} ($r = 8 \times 5$) & 1 & 2.181 & 2.910 \\
\texttt{LoRA-MLE} ($r = 8 \times 7$) & 1 & 2.182 & 2.935 \\
\midrule\midrule
\texttt{LoRA-MCL} & 3 & 2.063 & 2.663 \\
\texttt{LoRA-MCL} & 5 & 1.999 & 2.643 \\
\texttt{LoRA-MCL} & 7 & \textbf{1.932} & \textbf{2.612}\\
\bottomrule
    \end{tabular}}
    \label{tab:nll_vs_k}
\end{table}

\textbf{Effect of the number of hypotheses.} Table \ref{tab:nll_vs_k} reports the test negative log-likelihood as a function of $K$. The monotonically decreasing trend provides further evidence for Prop.~\ref{prop:em}, indicating that \texttt{LoRA-MCL} achieves better coverage of the data distribution modes as $K$ increases.

\subsection{Image description with reading comprehension}
\label{sec:image_captioning}

\begin{figure*}[ht]
    \begin{minipage}[t]{0.4\textwidth}
        \captionof{table}{\textbf{SPIDEr $(\uparrow)$ \& mBLEU-4 $(\downarrow)$ on different parts of synthetic test set.}}
        \centering
        \resizebox{\linewidth}{!}{
            \begin{tabular}{lllll}
            \toprule
            Test subset & Training & SPIDEr & mBLEU-4  \\
            \midrule
            French & \texttt{LoRA-MLE} & 0.411 & 0.138 \\
            & \texttt{LoRA-MCL} & \textbf{0.464} & \textbf{0.027}  \\
            \midrule
            English & \texttt{LoRA-MLE} & \textbf{0.756} & 0.126 \\
            & \texttt{LoRA-MCL} & 0.722 & \textbf{0.029}\\
            \bottomrule
        \end{tabular}
        }
        \label{tab:translationTable}
    \end{minipage}
    \hfill
    \centering
    \begin{minipage}[t]{0.29\textwidth}
    \vspace{0em}
        \centering
        \begin{minipage}[t]{0.46\textwidth}
            \includegraphics[width=\linewidth]{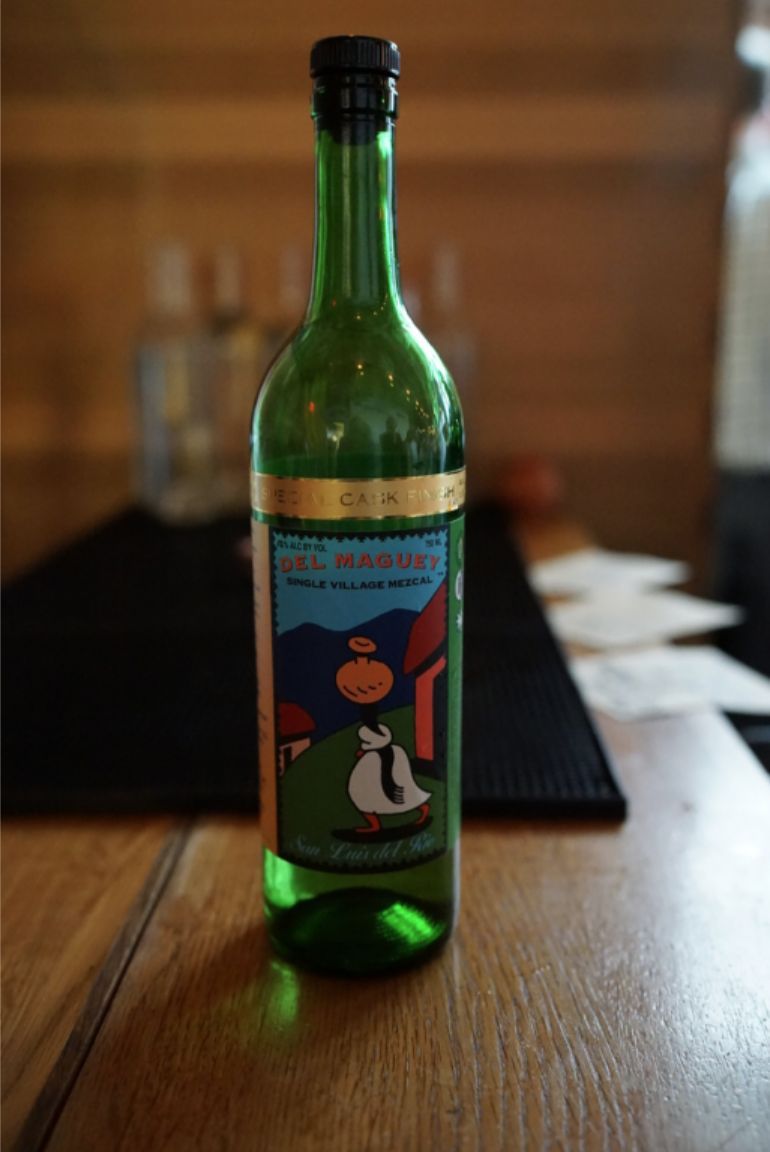}
        \end{minipage}%
        \hfill
        \begin{minipage}[t]{0.46\textwidth}
        \vspace{-8.75em}
            \raggedright
             {\fontsize{6pt}{7pt}\selectfont \textbf{\texttt{LoRA-MLE.}}\\      
            \{A bottle of Cerveza is on a table.\}\\
            
            \{Une bouteille de vin de cidre de cidre de cidre [...]\}\\
            \textbf{\texttt{LoRA-MCL.}}\\
            \{A bottle of beer with a label that says "Sel Maguet"\}\\
            \{Une bouteille de vin est étiquetée avec le mot « Maguay ».\}\\}
        \end{minipage}
    \end{minipage}
    \begin{minipage}[t]{0.29\textwidth}
    \vspace{1.5em}
        \centering
        \begin{minipage}[t]{0.46\textwidth}            \includegraphics[width=\linewidth]{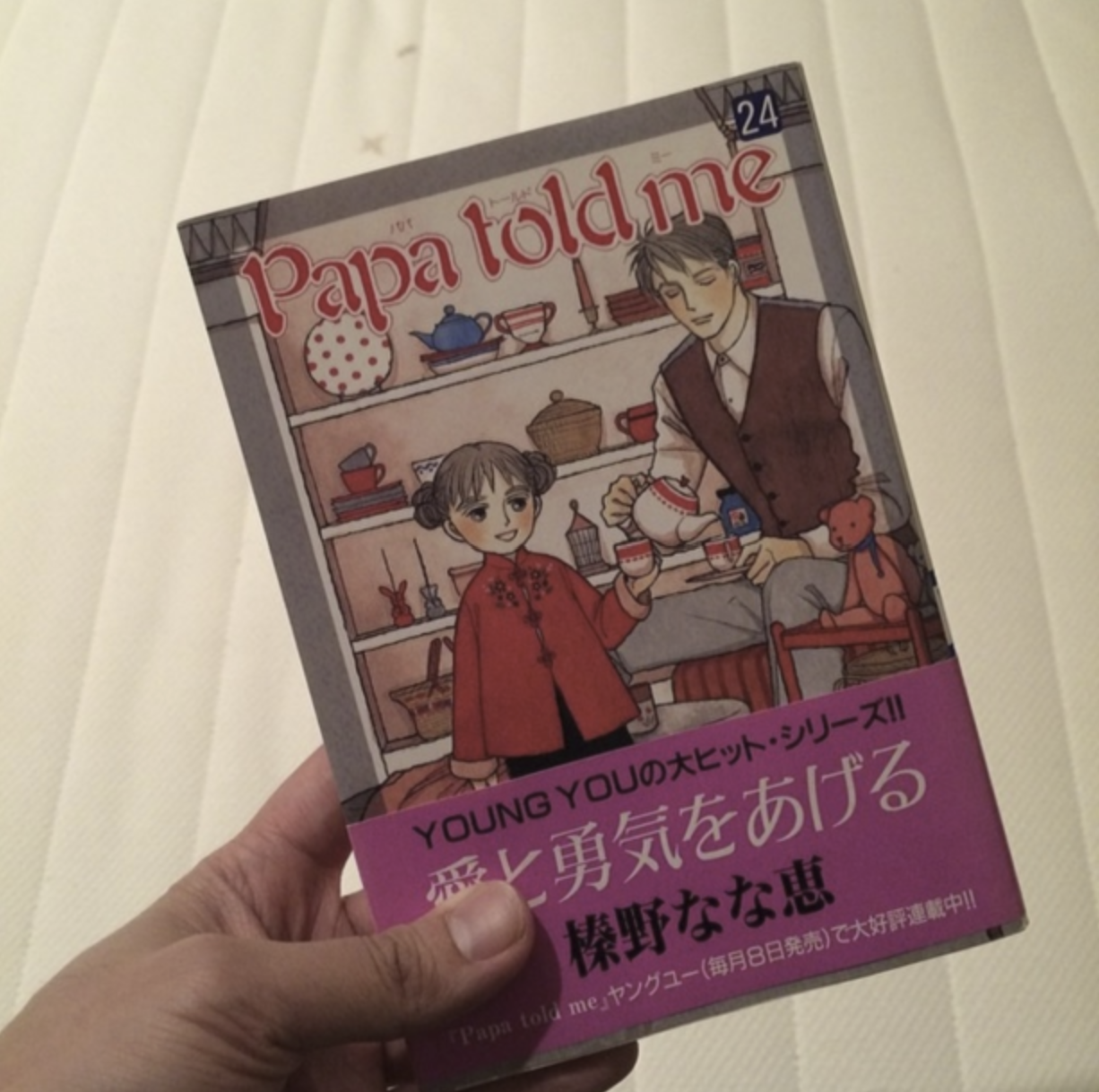}
        \end{minipage}%
        \hfill
        \begin{minipage}[t]{0.46\textwidth}
        \vspace{-7em}
            \raggedright      
            {\fontsize{6pt}{7pt}\selectfont \textbf{\texttt{LoRA-MLE.}}\\
            \{A book titled Papa Told Me is being held by a person.\}\\     
            \{A book called Papa told me is being held by a person.\}\\
            \textbf{\texttt{LoRA-MCL.}}\\
            \{A book titled Papa Told Me is being held by a person\}\\
            \{Un livre papier intitulé Papa Told Me.\}\\}
        \end{minipage}
    \end{minipage}
    
    \caption{\textbf{Observing specialization in bilingual image description.} Quantitative (\textit{Left}) and Qualitative \textit{(Right)} analysis for \texttt{LoRA-MLE} and \texttt{LoRA-MCL} in the setup of Sec.~\ref{sec:specialization}.}
    \label{fig:qual_predictions}
    \end{figure*}

\normalsize

\begin{figure*}[ht!]
\centering

\begin{minipage}[t]{0.62\textwidth}
\vspace{0pt}
\captionsetup{position=top}
\captionof{table}{\textbf{Quality and Diversity Evaluation on TextCaps ($K = 3$).} \texttt{LoRA-MCL}: $\varepsilon=0.1$, $r=8$, $\alpha=32$. \texttt{LoRA-MLE}: $r=24$, $\alpha=96$; rows marked $^\dagger$ use $r=8$, $\alpha=32$. Best in \textbf{bold}; second-best \underline{underlined}. Higher is better except mBLEU-4.}
\label{tab:perf_textcaps}

\scriptsize
\resizebox{\linewidth}{!}{
\begin{tabular}{lllllll}
\toprule
Training & Decoding & Beam & $\mathrm{mBLEU}_{4}$ & $\mathrm{CIDEr}_{D}$ & $\mathrm{SPICE}$ & $\mathrm{SPIDEr}$ \\
\midrule
\texttt{LoRA-MLE} & BS & 3 & 0.688 & 1.517 & 0.244 & 0.873 \\
\texttt{LoRA-MLE} & BS & 6 & 0.786 & 1.557 & 0.246 & 0.895 \\
\texttt{LoRA-MLE} & DBS ($\lambda=0.8$) & 3 & 0.437 & 1.590 & 0.251 & 0.909 \\
\texttt{LoRA-MLE} & DBS ($\lambda=1.0$) & 3 & \textbf{0.416} & 1.586 & 0.250 & 0.906 \\
\texttt{LoRA-MLE} & DBS ($\lambda=0.8$) & 6 & 0.671 & 1.573 & 0.251 & 0.903 \\
\texttt{LoRA-MLE} & DBS ($\lambda=0.8$)$^\dag$ & 3 & 0.531 & 1.589 & 0.255 & 0.912 \\
\texttt{LoRA-MLE} & DBS ($\lambda=1.0$)$^\dag$ & 3 & 0.425 & \underline{1.601} & 0.252 & 0.915 \\
\texttt{LoRA-MoE} & DBS ($\lambda=0.8$) & 3 & 0.441 & 1.616 & 0.254 & 0.924 \\
\texttt{LoRA-MoE} & DBS ($\lambda=1.0$) & 3 & \underline{0.421} & 1.622 & 0.253 & 0.926 \\
\texttt{LoRA-MoE} & DBS ($\lambda=0.8$) & 6 & 0.678 & 1.608 & 0.255 & 0.922 \\
\midrule\midrule
\texttt{LoRA-MCL} & BS & 1 & 0.520 & \textbf{1.674} & \underline{0.255} & \textbf{0.955} \\
\texttt{LoRA-MCL} & BS & 2 & 0.490 & \underline{1.627} & \textbf{0.258} & \underline{0.932} \\
\bottomrule
\end{tabular}
}
\end{minipage}
\hfill
\begin{minipage}[t]{0.35\textwidth}
\vspace{0pt}
\captionsetup{position=top}
\captionof{figure}{\textbf{Quality vs. Diversity in Machine Translation.}}
\label{fig:results_mt}

\centering
\includegraphics[width=\linewidth]{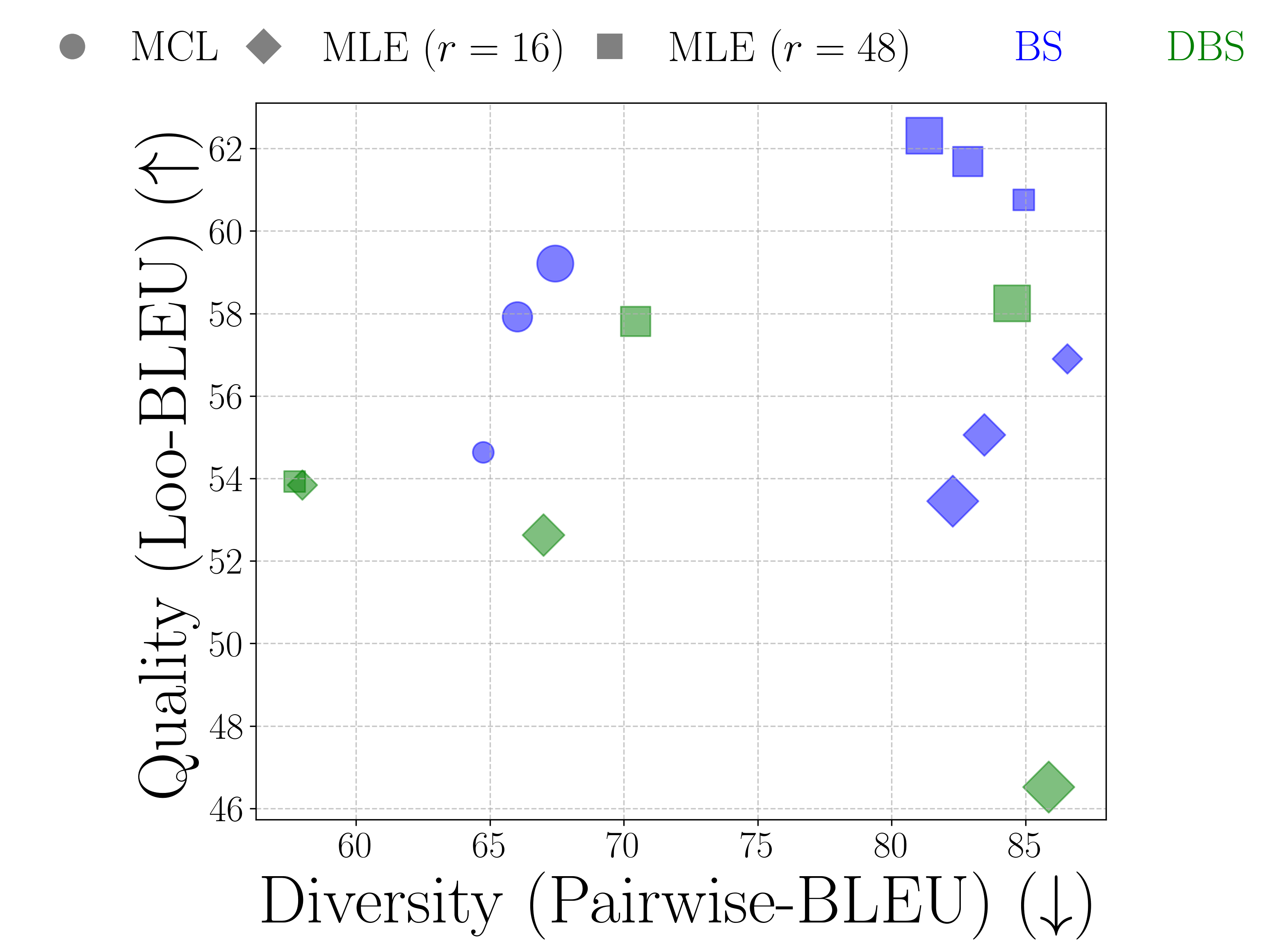}
\end{minipage}

\end{figure*}

\subsubsection{Quality and diversity evaluation}

Evaluation in image captioning Table~\ref{tab:perf_textcaps} confirms the trends observed in the audio captioning task. At an equal number of forward passes, \texttt{LoRA-MCL} outperforms \texttt{LoRA-MLE} and \texttt{LoRA-MoE}, even using DBS with a $\lambda$ diversity parameter specifically optimized for the task (SPIDEr of $0.955$ vs. $0.926$). Consistently with audio captioning, increasing the number of beams in each group can decrease diversity and does not improve the performance of \texttt{LoRA-MLE} and \texttt{LoRA-MoE}. However, we noticed that the DBS with \texttt{LoRA-MLE} tends to generate more diverse outputs than the greedy decoding of \texttt{LoRA-MCL} (mBLEU of $0.416$ vs. $0.520$). Combining DBS with LoRA-MCL, inspired by \citet{guzman2014efficiently}, could help address this issue. Additional comparisons are provided in Apx.~\ref{apx:add_results}.

\subsubsection{Observing hypothesis specialization in bilingual image description}

\begin{figure}[h!]
\centering
\includegraphics[width=0.65\linewidth]{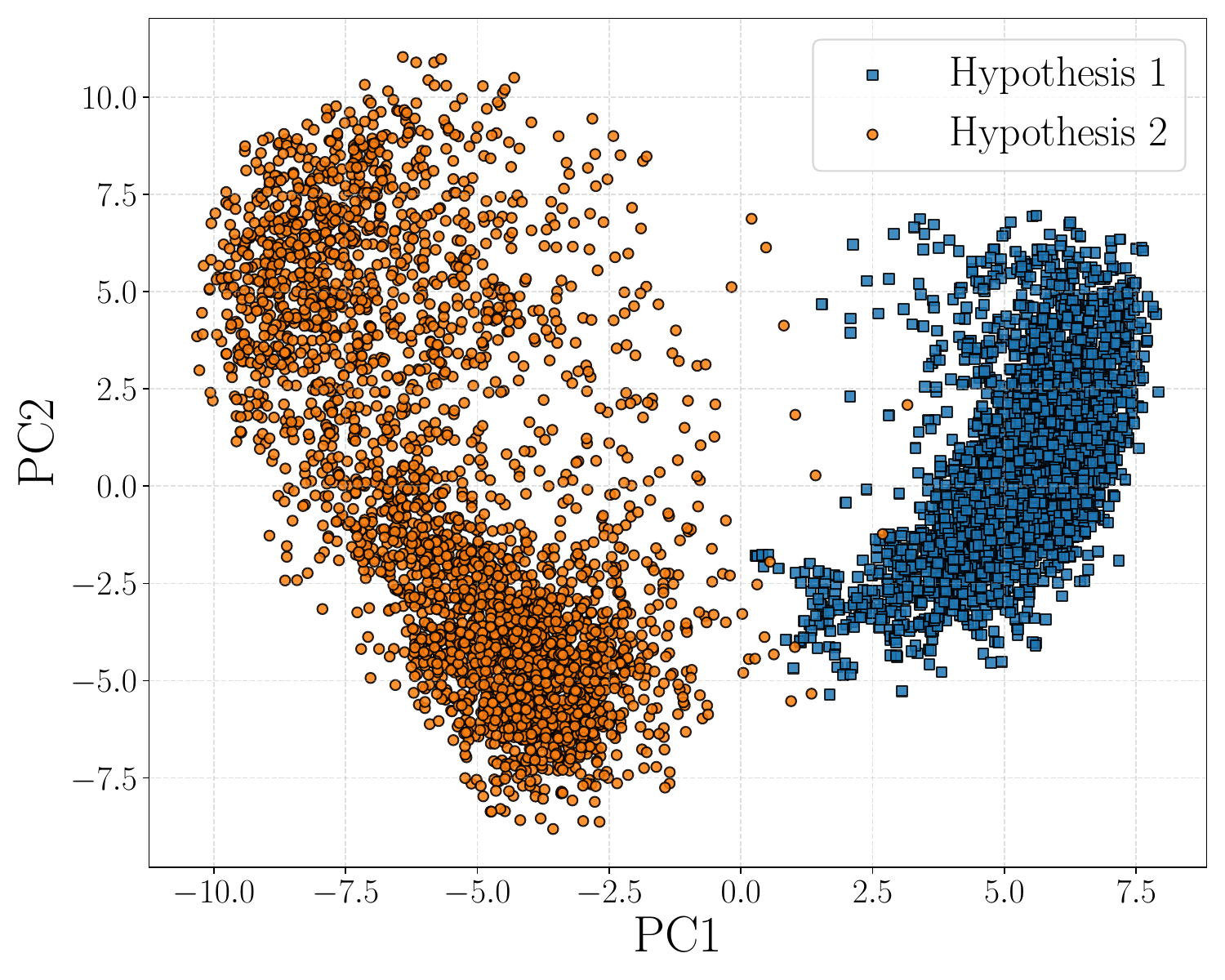}
\caption{\textbf{PCA of hypothesis candidate embeddings.} PC1 and PC2 correspond to the first two principal components, explaining $31.81\%$ and $15.22\%$ of the variance respectively.}
\label{fig:pca_toy}
\end{figure}

\label{sec:specialization}
To highlight the behavior of \texttt{LoRA-MCL} in a realistic case where one can control the modes of the data-generating process, we simulated an artificial bi-modal distribution of the dataset, similar to the setup of the toy experiment of Sec.~\ref{sec:synt_data}. We did so by translating half of the captions of the data from English to French (using T5-small~\citep{T5}), while keeping the prompts in English.

We trained a two-hypothesis \texttt{LoRA-MCL} model and \texttt{LoRA-MLE} baseline. We observed a specialization of each hypothesis towards a given language (one hypothesis learned French and the other English): at test-time, the winning head is the first one in $\sim 89\%$ of the French captions and the second one in $\sim 97\%$ of the English captions. Table~\ref{tab:translationTable} reports quality/diversity on the synthetic test set. \texttt{LoRA-MCL} uses greedy decoding, \texttt{LoRA-MLE} DBS with $\lambda = 0.8$ (maximizing its performance). Overall performance is similar, but \texttt{LoRA-MCL} is notably more diverse (mBLEU-4: FR $0.027$ vs $0.138$, EN $0.029$ vs $0.126$) and outperforms on French (SPIDEr $0.464$ vs $0.411$), with a slight reduction in English performance. Consistently with Sec.~\ref{sec:markov}, \texttt{LoRA-MLE} learns an average of the two modes (likely biased to English from LLaVA pretraining), whereas \texttt{LoRA-MCL} separates them. Additional experimental evidence is provided in Apx.~\ref{apx:specialization}.

Fig.\,\ref{fig:qual_predictions} illustrates the behavior of the models: \texttt{LoRA-MLE} learns a weighted average of the two modes, shifted towards English, and sometimes fails to output French captions within the two candidates, as in the book example. We found \texttt{LoRA-MLE} to be prone to errors when outputting French sentences: in the generation examples with an image of a bottle, \texttt{LoRA-MLE} enters a repetition loop. On the other hand, \texttt{LoRA-MCL} is less affected by those artifacts on French sentences, and captures the two modes of the distribution, benefiting from hypothesis specialization. Fig.~\ref{fig:pca_toy} presents a PCA of the captions generated by each head, embedded using a Sentence-BERT model (StyleDistance from \citet{patel2025styledistance}). The resulting frontiers are clearly visible, demonstrating a strong specialization of each head. See Apx.~\ref{apx:specialization} for further analysis.

\subsection{Diverse machine translation}
\label{sec:translation}

We evaluate \texttt{LoRA-MCL} for zero-shot machine translation with LLMs. Following \citet{xuparadigm}, we use a two-stage paradigm: (1) full-parameter fine-tuning on a monolingual corpus, and (2) LoRA fine-tuning on parallel data. We build on \href{https://huggingface.co/haoranxu/ALMA-7B-Pretrain}{ALMA-7B}, stage-1 fine-tuned, which we fine-tune on parallel data from \citet{xuparadigm} (see Apx.~\ref{apx:diverse_nmt}), comprising WMT’17–20 test sets and Flores-200 dev/test sets, restricted to English–German ($\sim$14k pairs). Evaluation uses the \textit{newstest2014} subset from \citet{ott2018analyzing}, containing 500 English sentences with ten German references each.

Fig.~\ref{fig:results_mt} displays the results, with a legend mirroring the one in Fig.~\ref{fig:quality-vs-diversity}. \texttt{LoRA-MCL} (circles) is trained with $K=3$ and $\varepsilon=0.05$, and each model generates $3$ sequences per input. We evaluate \texttt{LoRA-MLE} with ranks $r=16$ (diamond) and $r=48$ (square), using BS (blue) with widths $3$, $6$, and $9$, and DBS (green) with $\lambda=0.8$. We follow the quality–diversity evaluation protocol of \citet{shen2019mixture}. Scores are reported using both Leave-One-Out BLEU (Loo-BLEU) and Pairwise-BLEU \citep{shen2019mixture}. The results show that \texttt{LoRA-MCL} achieves a balance between quality and diversity, confirming the effectiveness of the method.

\section{Related Work}
\textbf{MCL to predict diverse and plausible outputs.} MCL \citep{guzman2012multiple, lee2016stochastic} is a training paradigm that minimizes the WTA loss across a set of models, encouraging specialization \citep{rupprecht2017learning, letzelter2024winner}. While the collapse issue needs to be addressed \citep{rupprecht2017learning, amcl}, MCL has demonstrated broad applicability across tasks \citep{lee2017confident, seo2020trajectory, garcia2021distillation}, typically using a shared backbone and multiple heads. However, using multiple heads may be impractical in LLMs. Mixture-of-Experts (MoE) \citep{jacobs1991adaptive, shazeer2017outrageously} offers an alternative for managing computational costs, since only a subset of experts is active at each forward pass. In LLMs, however, MoE has primarily been used to improve scalability rather than to encourage diversity, and suffers from redundancy among experts \citep{jiang2024mixtral}. While MoE can be adapted to the LoRA setting \cite{wumixture, li2024mixlora}, there is no clear consensus on the degree of specialization achieved. To our knowledge, this work is the first to adapt MCL to next-token language modeling using multiple LoRA modules.

\textbf{Generating Multiple Outputs with LMs.} Language models are commonly trained via next-token prediction, framed as Maximum Likelihood Estimation. This is arguably the most popular method for training large-scale language models \citep{shlegerislanguage, radford2019language, touvron2023llama}, including those that take audio or images as input \citep{chu2024qwen2, liu2023visual}, with much of its success attributed to tokenization \citep{makkuva2024attention, rajaraman2024analysis}. Generating diverse and plausible sequences at inference remains challenging: $(i)$ Sampling methods \citep{holtzmancurious, fan2018hierarchical, meister2023locally} may be unreliable depending on the chosen parameters; $(ii)$ Exact MAP decoding is intractable due to the exponential search space \citep{eikema2020map}; $(iii)$ Strategies like Beam Search often yield repetitive or overly coherent outputs \citep{fan2018hierarchical, holtzmancurious, keskar2019ctrl}. Diverse Beam Search \citep{mei2022diverse} and Test-time Augmentation \cite{wang2019aleatoric, kim2022exploring, kaya2025efficient} inject diversity through test-time parameters (e.g., penalty $\lambda$), but in contrast, \texttt{LoRA-MCL} infers diversity from input’s ambiguity.

\textbf{Diversity in Audio and Visual Captioning.} Audio \citep{drossos2017automated, mei2021audio, mei2022automated} and image captioning \citep{aneja2018convolutional, herdade2019image, hossain2019comprehensive} have traditionally relied on MLE-trained, task-specific models. A key challenge is the limited diversity of generated captions, leading to training objectives tailored to this issue \citep{mei2022diverse, xu2022diversity, xu2024towards, zhang2024generating, wang2019describing, wang2020diversity, mahajan2020diverse}. These approaches often require architectural changes and models trained from scratch. With the rise of general-purpose multimodal LLMs \citep{chu2024qwen2, liu2023visual}, addressing the diversity–quality trade-off remains critical. We show that \texttt{LoRA-MCL} effectively tackles this at the fine-tuning stage.

\section{Conclusion}

\texttt{LoRA-MCL} combines MCL with LoRA to train language models for diverse, plausible predictions. We show that when the target sequence is a mixture, \texttt{LoRA-MCL} can capture the modes of the data distribution. We validate this on Markov chains as well as the tasks of audio, image captioning, and machine translation. Future work will focus on further interpreting the concepts learned by each hypothesis, and investigating recent LoRA variants such as \citet{hayou2024lora+, zhang2025lora}. Orthogonally, extending MCL beyond fine-tuning to a pretraining setting represents a promising direction for further research.

\textbf{Limitations.} In the relaxed variant, setting $\varepsilon$ too high guarantees that all hypotheses receive gradients, but it may also overly homogenize the models and reduce diversity. A similar effect arises in the annealed variant, where the temperature scheduler parameters influence performance. Exploring dynamic adjustment of these parameter values depending on the data distribution is left for future work.

\section*{Impact Statement}

This paper presents work whose goal is to advance the field
of Machine Learning. There are many potential societal
consequences of our work, none which we feel must be
specifically highlighted here.

\section*{Acknowledgments}

We thank Mickael Chen for his valuable advice, Gilles Puy for his help in the implementation of the parallelization over the hypotheses, Lounès Meddahi and Monika Wysoczanska for their help with data management. This work was funded by the French Association for Technological Research (ANRT CIFRE contract 2022-1854), and granted access to the HPC resources of IDRIS (allocation 2024-AD011014345) by GENCI. This work was also supported by the European Union’s Horizon Europe research and innovation program under grant agreement No 101214398 (ELLIOT).

\bibliography{references}

@inproceedings{fan2018hierarchical,
  title={Hierarchical Neural Story Generation},
  author={Fan, Angela and Lewis, Mike and Dauphin, Yann},
  booktitle={ACL},
  year={2018}
}

@inproceedings{holtzmancurious,
  title={The Curious Case of Neural Text Degeneration},
  author={Holtzman, Ari and Buys, Jan and Du, Li and Forbes, Maxwell and Choi, Yejin},
  booktitle={ICLR},
    year={2019}
}

@inproceedings{vijayakumar2018diverse,
  title={Diverse beam search for improved description of complex scenes},
  author={Vijayakumar, Ashwin and Cogswell, Michael and Selvaraju, Ramprasaath and Sun, Qing and Lee, Stefan and Crandall, David and Batra, Dhruv},
  booktitle={AAAI},
  year={2018}
}

@article{williams1989learning,
  title={A learning algorithm for continually running fully recurrent neural networks},
  author={Williams, Ronald J and Zipser, David},
  journal={Neural computation},
  year={1989},
}

@inproceedings{lee2016stochastic,
  title={Stochastic multiple choice learning for training diverse deep ensembles},
  author={Lee, Stefan and Purushwalkam Shiva Prakash, Senthil and Cogswell, Michael and Ranjan, Viresh and Crandall, David and Batra, Dhruv},
  booktitle={NeurIPS},
  year={2016}
}

@article{guzman2012multiple,
  title={Multiple choice learning: Learning to produce multiple structured outputs},
  author={Guzman-Rivera, Abner and Batra, Dhruv and Kohli, Pushmeet},
  journal={NeurIPS},
  year={2012}
}

@inproceedings{drossos2017automated,
  title={Automated audio captioning with recurrent neural networks},
  author={Drossos, Konstantinos and Adavanne, Sharath and Virtanen, Tuomas},
  booktitle={WASPAA},
  year={2017}
}

@inproceedings{kim2019audiocaps,
  title={Audiocaps: Generating captions for audios in the wild},
  author={Kim, Chris Dongjoo and Kim, Byeongchang and Lee, Hyunmin and Kim, Gunhee},
  booktitle={NAACL},
  year={2019}
}

@inproceedings{drossos2020clotho,
  title={Clotho: An audio captioning dataset},
  author={Drossos, Konstantinos and Lipping, Samuel and Virtanen, Tuomas},
  booktitle={ICASSP},
  year={2020}
  }

@inproceedings{gemmeke2017audio,
  title={Audio set: An ontology and human-labeled dataset for audio events},
  author={Gemmeke, Jort F and Ellis, Daniel PW and Freedman, Dylan and Jansen, Aren and Lawrence, Wade and Moore, R Channing and Plakal, Manoj and Ritter, Marvin},
  booktitle={ICASSP},
  year={2017}
 }

@inproceedings{font2013freesound,
  title={Freesound technical demo},
  author={Font, Frederic and Roma, Gerard and Serra, Xavier},
  booktitle={ACMMM},
  year={2013}
}

@inproceedings{papineni2002bleu,
  title={Bleu: a method for automatic evaluation of machine translation},
  author={Papineni, Kishore and Roukos, Salim and Ward, Todd and Zhu, Wei-Jing},
  booktitle={ACL},
  year={2002}
}

@inproceedings{lin2004rouge,
  title={Rouge: A package for automatic evaluation of summaries},
  author={Lin, Chin-Yew},
  booktitle={Text summarization branches out},
  year={2004}
}

@inproceedings{banerjee2005meteor,
  title={METEOR: An automatic metric for MT evaluation with improved correlation with human judgments},
  author={Banerjee, Satanjeev and Lavie, Alon},
  booktitle={ACL Workshops},
  year={2005}
}

@inproceedings{vedantam2015cider,
  title={Cider: Consensus-based image description evaluation},
  author={Vedantam, Ramakrishna and Lawrence Zitnick, C and Parikh, Devi},
  booktitle={CVPR},
  year={2015}
}

@inproceedings{anderson2016spice,
  title={Spice: Semantic propositional image caption evaluation},
  author={Anderson, Peter and Fernando, Basura and Johnson, Mark and Gould, Stephen},
  booktitle={ECCV},
  year={2016},
}

@inproceedings{liu2017improved,
  title={Improved image captioning via policy gradient optimization of spider},
  author={Liu, Siqi and Zhu, Zhenhai and Ye, Ning and Guadarrama, Sergio and Murphy, Kevin},
  booktitle={ICCV},
  year={2017}
}

@inproceedings{zhangbertscore,
  title={BERTScore: Evaluating Text Generation with BERT},
  author={Zhang, Tianyi and Kishore, Varsha and Wu, Felix and Weinberger, Kilian Q and Artzi, Yoav},
  booktitle={ICLR},
  year={2020}
}

@inproceedings{zhou2022can,
  title={Can audio captions be evaluated with image caption metrics?},
  author={Zhou, Zelin and Zhang, Zhiling and Xu, Xuenan and Xie, Zeyu and Wu, Mengyue and Zhu, Kenny Q},
  booktitle={ICASSP},
  year={2022},
}

@article{reimers2019sentence,
  title={Sentence-BERT: Sentence Embeddings using Siamese BERT-Networks},
  author={Reimers, Nils},
  journal={EMNLP-IJCNLP},
  year={2019}
}

@inproceedings{zhu2018texygen,
  title={Texygen: A benchmarking platform for text generation models},
  author={Zhu, Yaoming and Lu, Sidi and Zheng, Lei and Guo, Jiaxian and Zhang, Weinan and Wang, Jun and Yu, Yong},
  booktitle={SIGIR},
  year={2018}
}

@inproceedings{kausik2023learning,
  title={Learning mixtures of markov chains and mdps},
  author={Kausik, Chinmaya and Tan, Kevin and Tewari, Ambuj},
  booktitle={ICML},
  year={2023},
}

@article{malinin2020uncertainty,
  title={Uncertainty estimation in autoregressive structured prediction},
  author={Malinin, Andrey and Gales, Mark},
  journal={ICLR},
  year={2020}
}

@inproceedings{rupprecht2017learning,
  title={Learning in an uncertain world: Representing ambiguity through multiple hypotheses},
  author={Rupprecht, Christian and Laina, Iro and DiPietro, Robert and Baust, Maximilian and Tombari, Federico and Navab, Nassir and Hager, Gregory D},
  booktitle={ICCV},
  year={2017}
}

@article{devlin2018bert,
  title={Bert: Pre-training of deep bidirectional transformers for language understanding},
  author={Devlin, Jacob},
  journal={NAACL},
  year={2019}
}

@article{jelinek1977perplexity,
  title={Perplexity—a measure of the difficulty of speech recognition tasks},
  author={Jelinek, Fred and Mercer, Robert L and Bahl, Lalit R and Baker, James K},
  journal={The Journal of the Acoustical Society of America},
  year={1977},
  publisher={Acoustical Society of America}
}

@article{meister2023locally,
  title={Locally typical sampling},
  author={Meister, Clara and Pimentel, Tiago and Wiher, Gian and Cotterell, Ryan},
  journal={TACL},
  year={2023},
}

@article{keskar2019ctrl,
  title={Ctrl: A conditional transformer language model for controllable generation},
  author={Keskar, Nitish Shirish and McCann, Bryan and Varshney, Lav R and Xiong, Caiming and Socher, Richard},
  journal={arXiv preprint arXiv:1909.05858},
  year={2019}
}

@inproceedings{hu2022lora,
  title={Lora: Low-rank adaptation of large language models.},
  author={Hu, Edward J and Shen, Yelong and Wallis, Phillip and Allen-Zhu, Zeyuan and Li, Yuanzhi and Wang, Shean and Wang, Lu and Chen, Weizhu and others},
  booktitle={ICLR},
  year={2022}
}

@inproceedings{letzelter2024winner,
  title={Winner-takes-all learners are geometry-aware conditional density estimators},
  author={Letzelter, Victor and Perera, David and Rommel, C{\'e}dric and Fontaine, Mathieu and Essid, Slim and Richard, Ga{\"e}l and Perez, Patrick},
  booktitle={ICML},
  year={2024},
}

@inproceedings{zhang2021trading,
  title={Trading Off Diversity and Quality in Natural Language Generation},
  author={Zhang, Hugh and Duckworth, Daniel and Ippolito, Daphne and Neelakantan, Arvind},
  booktitle={HumEval},
  year={2021}
}

@inproceedings{hewitt2022truncation,
  title={Truncation Sampling as Language Model Desmoothing},
  author={Hewitt, John and Manning, Christopher D and Liang, Percy},
  booktitle={EMNLP},
  year={2022}
}

@article{sucontrastive,
  title={Contrastive Search Is What You Need For Neural Text Generation},
  author={Su, Yixuan and Collier, Nigel},
  journal={TMLR},
  year={2023}
}

@article{welleckdecoding,
  title={From Decoding to Meta-Generation: Inference-time Algorithms for Large Language Models},
  author={Welleck, Sean and Bertsch, Amanda and Finlayson, Matthew and Schoelkopf, Hailey and Xie, Alex and Neubig, Graham and Kulikov, Ilia and Harchaoui, Zaid},
  journal={TMLR},
  year={2024}
}

@inproceedings{sutskever2014sequence,
  title={Sequence to sequence learning with neural networks},
  author={Sutskever, Ilya and Vinyals, Oriol and Le, Quoc V},
  booktitle={NeurIPS},
  year={2014}
}

@article{gupta2016mixtures,
  title={On mixtures of Markov chains},
  author={Gupta, Rishi and Kumar, Ravi and Vassilvitskii, Sergei},
  journal={NeurIPS},
  year={2016}
}

@inproceedings{vaswani2017attention,
  title={Attention is all you need},
  author={Vaswani, Ashish and Shazeer, Noam and Parmar, Niki and Uszkoreit, Jakob and Jones, Llion and Gomez, Aidan N and Kaiser, {\L}ukasz and Polosukhin, Illia},
  booktitle={NeurIPS},
  year={2017}
}

@article{blei2003latent,
  title={Latent dirichlet allocation},
  author={Blei, David M and Ng, Andrew Y and Jordan, Michael I},
  journal={JMLR},
  year={2003}
}

@inproceedings{yang2018breaking,
  title={Breaking the Softmax Bottleneck: A High-Rank RNN Language Model},
  author={Yang, Zhilin and Dai, Zihang and Salakhutdinov, Ruslan and Cohen, William W},
  booktitle={ICLR},
  year={2018}
}

@article{nigam2000text,
  title={Text classification from labeled and unlabeled documents using EM},
  author={Nigam, Kamal and McCallum, Andrew Kachites and Thrun, Sebastian and Mitchell, Tom},
  journal={ML},
  year={2000}
}

@article{dieng2020topic,
  title={Topic modeling in embedding spaces},
  author={Dieng, Adji B and Ruiz, Francisco JR and Blei, David M},
  journal={TACL},
  year={2020}
}

@inproceedings{
wen2023an,
title={An Equal-Size Hard {EM} Algorithm for Diverse Dialogue Generation},
author={Yuqiao Wen and Yongchang Hao and Yanshuai Cao and Lili Mou},
booktitle={ICLR},
year={2023}
}

@inproceedings{min2019discrete,
  title={A Discrete Hard EM Approach for Weakly Supervised Question Answering},
  author={Min, Sewon and Chen, Danqi and Hajishirzi, Hannaneh and Zettlemoyer, Luke},
  booktitle={EMNLP-IJCNLP},
  year={2019}
}

@article{chu2024qwen2,
  title={Qwen2-audio technical report},
  author={Chu, Yunfei and Xu, Jin and Yang, Qian and Wei, Haojie and Wei, Xipin and Guo, Zhifang and Leng, Yichong and Lv, Yuanjun and He, Jinzheng and Lin, Junyang and others},
  journal={arXiv preprint arXiv:2407.10759},
  year={2024}
}

@inproceedings{liu2023visual,
  title={Visual instruction tuning},
  author={Liu, Haotian and Li, Chunyuan and Wu, Qingyang and Lee, Yong Jae},
  booktitle={NeurIPS},
  year={2023}
}

@article{touvron2023llama,
  title={Llama: Open and efficient foundation language models},
  author={Touvron, Hugo and Lavril, Thibaut and Izacard, Gautier and Martinet, Xavier and Lachaux, Marie-Anne and Lacroix, Timoth{\'e}e and Rozi{\`e}re, Baptiste and Goyal, Naman and Hambro, Eric and Azhar, Faisal and others},
  journal={arXiv preprint arXiv:2302.13971},
  year={2023}
}

@article{makkuva2024attention,
  title={Attention with Markov: A Curious Case of Single-Layer Transformers},
  author={Makkuva, Ashok Vardhan and Bondaschi, Marco and Girish, Adway and Nagle, Alliot and Jaggi, Martin and Kim, Hyeji and Gastpar, Michael},
  journal={ICLR},
  year={2025}
}

@article{radford2019language,
  title={Language models are unsupervised multitask learners},
  author={Radford, Alec and Wu, Jeffrey and Child, Rewon and Luan, David and Amodei, Dario and Sutskever, Ilya and others},
  journal={OpenAI blog},
  year={2019}
}

@article{black2021gpt,
  title={Gpt-neo: Large scale autoregressive language modeling with mesh-tensorflow},
  author={Black, Sid and Leo, Gao and Wang, Phil and Leahy, Connor and Biderman, Stella},
  journal={Zenodo},
  year={2021}
}

@article{mclachlan2019finite,
  title={Finite mixture models},
  author={McLachlan, Geoffrey J and Lee, Sharon X and Rathnayake, Suren I},
  journal={Annual review of statistics and its application},
  year={2019},
}

@article{redner1984mixture,
  title={Mixture densities, maximum likelihood and the EM algorithm},
  author={Redner, Richard A and Walker, Homer F},
  journal={SIAM review},
  year={1984},
}

@inproceedings{letzelter2023resilient,
  title={Resilient Multiple Choice Learning: A learned scoring scheme with application to audio scene analysis},
  author={Letzelter, Victor and Fontaine, Mathieu and Chen, Micka{\"e}l and P{\'e}rez, Patrick and Essid, Slim and Richard, Ga{\"e}l},
  booktitle={NeurIPS},
  year={2023}
}

@article{zekri2024large,
  title={Large language models as markov chains},
  author={Zekri, Oussama and Odonnat, Ambroise and Benechehab, Abdelhakim and Bleistein, Linus and Boull{\'e}, Nicolas and Redko, Ievgen},
  journal={arXiv preprint arXiv:2410.02724},
  year={2024}
}

@inproceedings{rajaraman2024analysis,
  title={An Analysis of Tokenization: Transformers under Markov Data},
  author={Rajaraman, Nived and Jiao, Jiantao and Ramchandran, Kannan},
  booktitle={NeurIPS},
  year={2024}
}

@article{edelman2024evolution,
  title={The evolution of statistical induction heads: In-context learning markov chains},
  author={Edelman, Ezra and Tsilivis, Nikolaos and Edelman, Benjamin and Malach, Eran and Goel, Surbhi},
  journal={NeurIPS},
  year={2024}
}

@article{wolf2019huggingface,
  title={Huggingface's transformers: State-of-the-art natural language processing},
  author={Wolf, Thomas and Debut, Lysandre and Sanh, Victor and Chaumond, Julien and Delangue, Clement and Moi, Anthony and Cistac, Pierric and Rault, Tim and Louf, R{\'e}mi and Funtowicz, Morgan and others},
  journal={EMNLP},
  year={2020}
}

@Misc{peft,
  title =        {PEFT: State-of-the-art Parameter-Efficient Fine-Tuning methods},
  author =       {Sourab Mangrulkar and Sylvain Gugger and Lysandre Debut and Younes Belkada and Sayak Paul and Benjamin Bossan},
  year =         {2022}
}

@inproceedings{sidorov2020textcaps,
  title={Textcaps: a dataset for image captioning with reading comprehension},
  author={Sidorov, Oleksii and Hu, Ronghang and Rohrbach, Marcus and Singh, Amanpreet},
  booktitle={ECCV},
  year={2020}
}

@inproceedings{loshchilovdecoupled,
  title={Decoupled Weight Decay Regularization},
  author={Loshchilov, Ilya and Hutter, Frank},
  booktitle={ICLR},
  year={2017}
}

@inproceedings{labbe2022my,
  title={Is my automatic audio captioning system so bad? spider-max: a metric to consider several caption candidates},
  author={Labb{\'e}, Etienne and Pellegrini, Thomas and Pinquier, Julien},
  booktitle={DCASE},
  year={2022}
}

@book{mackay2003information,
  title={Information theory, inference and learning algorithms},
  author={MacKay, David JC},
  year={2003},
  publisher={Cambridge university press}
}

@inproceedings{ippolito2019comparison,
  title={Comparison of Diverse Decoding Methods from Conditional Language Models},
  author={Ippolito, Daphne and Kriz, Reno and Sedoc, Jo{\~a}o and Kustikova, Maria and Callison-Burch, Chris},
  booktitle={ACL},
  year={2019}
}

@inproceedings{lee2017confident,
  title={Confident multiple choice learning},
  author={Lee, Kimin and Hwang, Changho and Park, KyoungSoo and Shin, Jinwoo},
  booktitle={ICML},
  year={2017},
}

@inproceedings{seo2020trajectory,
  title={Trajectory-wise multiple choice learning for dynamics generalization in reinforcement learning},
  author={Seo, Younggyo and Lee, Kimin and Clavera Gilaberte, Ignasi and Kurutach, Thanard and Shin, Jinwoo and Abbeel, Pieter},
  booktitle={NeurIPS},
  year={2020}
}

@inproceedings{garcia2021distillation,
  title={Distillation multiple choice learning for multimodal action recognition},
  author={Garcia, Nuno Cruz and Bargal, Sarah Adel and Ablavsky, Vitaly and Morerio, Pietro and Murino, Vittorio and Sclaroff, Stan},
  booktitle={WACV},
  year={2021}
}

@article{t5,
  title={Exploring the limits of transfer learning with a unified text-to-text transformer},
  author={Raffel, Colin and Shazeer, Noam and Roberts, Adam and Lee, Katherine and Narang, Sharan and Matena, Michael and Zhou, Yanqi and Li, Wei and Liu, Peter J},
  journal={MLR},
  year={2020}
}

@inproceedings{eikema2020map,
  title={Is MAP Decoding All You Need? The Inadequacy of the Mode in Neural Machine Translation},
  author={Eikema, Bryan and Aziz, Wilker},
  booktitle={COLING},
  year={2020}
}

@inproceedings{amcl,
  title={Annealed Multiple Choice Learning: Overcoming limitations of Winner-takes-all with annealing},
  author={Perera, David and Letzelter, Victor and Mariotte, Th{\'e}o and Cort{\'e}s, Adrien and Chen, Mickael and Essid, Slim and Richard, Ga{\"e}l},
  booktitle={NeurIPS},
  year={2024}
}

@article{shlegerislanguage,
  title={Language Models Are Better Than Humans at Next-token Prediction},
  author={Shlegeris, Buck and Roger, Fabien and Chan, Lawrence and McLean, Euan},
  journal={TMLR},
  year={2022}
}

@book{beamSearch,
  title={The harpy speech recognition system.},
  author={Lowerre, Bruce T},
  year={1976},
  publisher={Carnegie Mellon University}
}

@inproceedings{zhou2024empirical,
  title={An empirical study on parameter-efficient fine-tuning for multimodal large language models},
  author={Zhou, Xiongtao and He, Jie and Ke, Yuhua and Zhu, Guangyao and Guti{\'e}rrez-Basulto, V{\'\i}ctor and Pan, Jeff Z},
  booktitle={ACL},
  year={2024}
}

@article{labb2024conette,
  title={CoNeTTE: An efficient Audio Captioning system leveraging multiple datasets with Task Embedding},
  author={Labb, Etienne and Pellegrini, Thomas and Pinquier, Julien and others},
  journal={TASLPRO},
  year={2024}
}

@inproceedings{shazeer2017outrageously,
  title={Outrageously large neural networks: The sparsely-gated mixture-of-experts layer},
  author={Shazeer, Noam and Mirhoseini, Azalia and Maziarz, Krzysztof and Davis, Andy and Le, Quoc and Hinton, Geoffrey and Dean, Jeff},
  booktitle={ICLR},
  year={2017}
}

@article{jacobs1991adaptive,
  title={Adaptive mixtures of local experts},
  author={Jacobs, Robert A and Jordan, Michael I and Nowlan, Steven J and Hinton, Geoffrey E},
  journal={Neural computation},
  year={1991},
}

@article{zhao2025sample,
  title={Sample, Scrutinize and Scale: Effective Inference-Time Search by Scaling Verification},
  author={Zhao, Eric and Awasthi, Pranjal and Gollapudi, Sreenivas},
  journal={ICML},
  year={2025}
}

@article{jiang2024mixtral,
  title={Mixtral of experts},
  author={Jiang, Albert Q and Sablayrolles, Alexandre and Roux, Antoine and Mensch, Arthur and Savary, Blanche and Bamford, Chris and Chaplot, Devendra Singh and Casas, Diego de las and Hanna, Emma Bou and Bressand, Florian and others},
  journal={arXiv preprint arXiv:2401.04088},
  year={2024}
}

@inproceedings{zheng2023judging,
  title={Judging llm-as-a-judge with mt-bench and chatbot arena},
  author={Zheng, Lianmin and Chiang, Wei-Lin and Sheng, Ying and Zhuang, Siyuan and Wu, Zhanghao and Zhuang, Yonghao and Lin, Zi and Li, Zhuohan and Li, Dacheng and Xing, Eric and others},
  booktitle={NeurIPS},
year={2023}
}

@article{hayou2024lora+,
  title={LoRA+: Efficient Low Rank Adaptation of Large Models},
  author={Hayou, Soufiane and Ghosh, Nikhil and Yu, Bin},
  journal={ICML},
  year={2024}
}

@inproceedings{mei2021audio,
  title={Audio captioning transformer},
  author={Mei, Xinhao and Liu, Xubo and Huang, Qiushi and Plumbley, Mark D and Wang, Wenwu},
  booktitle={DCASE},
  year={2021}
}

@article{mei2022automated,
  title={Automated audio captioning: An overview of recent progress and new challenges},
  author={Mei, Xinhao and Liu, Xubo and Plumbley, Mark D and Wang, Wenwu},
  journal={EURASIP},
  year={2022},
}

@article{hossain2019comprehensive,
  title={A comprehensive survey of deep learning for image captioning},
  author={Hossain, MD Zakir and Sohel, Ferdous and Shiratuddin, Mohd Fairuz and Laga, Hamid},
  journal={ACM Computing Surveys},
  year={2019},
}

@article{herdade2019image,
  title={Image captioning: Transforming objects into words},
  author={Herdade, Simao and Kappeler, Armin and Boakye, Kofi and Soares, Joao},
  journal={NeurIPS},
  year={2019}
}

@inproceedings{aneja2018convolutional,
  title={Convolutional image captioning},
  author={Aneja, Jyoti and Deshpande, Aditya and Schwing, Alexander G},
  booktitle={CVPR},
  year={2018}
}

@inproceedings{mei2022diverse,
  title={Diverse audio captioning via adversarial training},
  author={Mei, Xinhao and Liu, Xubo and Sun, Jianyuan and Plumbley, Mark D and Wang, Wenwu},
  booktitle={ICASSP},
  year={2022},
}

@article{xu2024towards,
  title={Towards Diverse and Efficient Audio Captioning via Diffusion Models},
  author={Xu, Manjie and Li, Chenxing and Tu, Xinyi and Ren, Yong and Fu, Ruibo and Liang, Wei and Yu, Dong},
  journal={Interspeech},
  year={2025}
}

@inproceedings{xu2022diversity,
  title={Diversity-controllable and accurate audio captioning based on neural condition},
  author={Xu, Xuenan and Wu, Mengyue and Yu, Kai},
  booktitle={ICASSP},
  year={2022},
}

@article{zhang2024generating,
  title={Generating Accurate and Diverse Audio Captions through Variational Autoencoder Framework},
  author={Zhang, Yiming and Du, Ruoyi and Tan, Zheng-Hua and Wang, Wenwu and Ma, Zhanyu},
  journal={IEEE Signal Processing Letters},
  year={2024},
}

@inproceedings{wang2019describing,
  title={Describing like humans: on diversity in image captioning},
  author={Wang, Qingzhong and Chan, Antoni B},
  booktitle={CVPR},
  year={2019}
}

@article{wang2020diversity,
  title={On diversity in image captioning: Metrics and methods},
  author={Wang, Qingzhong and Wan, Jia and Chan, Antoni B},
  journal={IEEE Transactions on Pattern Analysis and Machine Intelligence},
  year={2020},
}

@article{mahajan2020diverse,
  title={Diverse image captioning with context-object split latent spaces},
  author={Mahajan, Shweta and Roth, Stefan},
  journal={NeurIPS},
  year={2020}
}

@book{cover1999elements,
  title={Elements of information theory},
  author={Cover, Thomas M},
  year={1999},
  publisher={John Wiley \& Sons}
}

@article{radford2018improving,
  title={Improving language understanding by generative pre-training},
  author={Radford, Alec and Narasimhan, Karthik and Salimans, Tim and Sutskever, Ilya and others},
  journal={OpenAI blog},
  year={2018}
}

@inproceedings{xiong2024efficient,
  title={Efficient and effective uncertainty quantification for LLMs},
  author={Xiong, Miao and Santilli, Andrea and Kirchhof, Michael and Golinski, Adam and Williamson, Sinead},
  booktitle={NeurIPS Safe Generative AI Workshop},
  year={2024}
}

@inproceedings{he2015delving,
  title={Delving deep into rectifiers: Surpassing human-level performance on imagenet classification},
  author={He, Kaiming and Zhang, Xiangyu and Ren, Shaoqing and Sun, Jian},
  booktitle={ICCV},
  year={2015}
}

@article{muqeethsoft,
  title={Soft Merging of Experts with Adaptive Routing},
  author={Muqeeth, Mohammed and Liu, Haokun and Raffel, Colin},
  journal={TMLR},
  year={2024}
}

@inproceedings{wumixture,
  title={Mixture of LoRA Experts},
  author={Wu, Xun and Huang, Shaohan and Wei, Furu},
  booktitle={ICLR},
  year={2024}
}

@article{li2024mixlora,
  title={Mixlora: Enhancing large language models fine-tuning with lora-based mixture of experts},
  author={Li, Dengchun and Ma, Yingzi and Wang, Naizheng and Ye, Zhengmao and Cheng, Zhiyuan and Tang, Yinghao and Zhang, Yan and Duan, Lei and Zuo, Jie and Yang, Cal and others},
  journal={arXiv preprint arXiv:2404.15159},
  year={2024}
}

@inproceedings{makansi2019overcoming,
  title={Overcoming limitations of mixture density networks: A sampling and fitting framework for multimodal future prediction},
  author={Makansi, Osama and Ilg, Eddy and Cicek, Ozgun and Brox, Thomas},
  booktitle={CVPR},
  year={2019}
}

@inproceedings{narayanan2021divide,
  title={Divide-and-conquer for lane-aware diverse trajectory prediction},
  author={Narayanan, Sriram and Moslemi, Ramin and Pittaluga, Francesco and Liu, Buyu and Chandraker, Manmohan},
  booktitle={CVPR},
  year={2021}
}

@article{nehme2024hierarchical,
  title={Hierarchical uncertainty exploration via feedforward posterior trees},
  author={Nehme, Elias and Mulayoff, Rotem and Michaeli, Tomer},
  journal={NeurIPS},
  year={2024}
}

@inproceedings{zuo2021taming,
 title={Taming sparsely activated transformer with stochastic experts},
  author={Zuo, Simiao and Liu, Xiaodong and Jiao, Jian and Kim, Young Jin and Hassan, Hany and Zhang, Ruofei and Zhao, Tuo and Gao, Jianfeng},
  booktitle={ICLR},
  year={2022}
}

@inproceedings{xuparadigm,
  title={A Paradigm Shift in Machine Translation: Boosting Translation Performance of Large Language Models},
  author={Xu, Haoran and Kim, Young Jin and Sharaf, Amr and Awadalla, Hany Hassan},
   booktitle={ICLR},
  year={2024}
}

@inproceedings{shen2019mixture,
  title={Mixture models for diverse machine translation: Tricks of the trade},
  author={Shen, Tianxiao and Ott, Myle and Auli, Michael and Ranzato, Marc’Aurelio},
  booktitle={ICML},
  year={2019},
}

@inproceedings{ott2018analyzing,
  title={Analyzing uncertainty in neural machine translation},
  author={Ott, Myle and Auli, Michael and Grangier, David and Ranzato, Marc’Aurelio},
  booktitle={ICML},
  year={2018},
}

@article{costa2022no,
  title={No language left behind: Scaling human-centered machine translation},
  author={Costa-Juss{\`a}, Marta R and Cross, James and {\c{C}}elebi, Onur and Elbayad, Maha and Heafield, Kenneth and Heffernan, Kevin and Kalbassi, Elahe and Lam, Janice and Licht, Daniel and Maillard, Jean and others},
  journal={arXiv preprint arXiv:2207.04672},
  year={2022}
}

@article{park2019specaugment,
  title={SpecAugment: A Simple Data Augmentation Method for Automatic Speech Recognition},
  author={Park, Daniel S and Chan, William and Zhang, Yu and Chiu, Chung-Cheng and Zoph, Barret and Cubuk, Ekin D and Le, Quoc V},
  journal={Interspeech},
  year={2019},
  publisher={ISCA}
}

@article{zhang2025lora,
  title={LoRA-One: One-Step Full Gradient Could Suffice for Fine-Tuning Large Language Models, Provably and Efficiently},
  author={Zhang, Yuanhe and Liu, Fanghui and Chen, Yudong},
  journal={ICML},
  year={2025}
}

@inproceedings{diederik2014adam,
  title={Adam: A method for stochastic optimization},
  author={Diederik, P Kingma and Jimmy, Ba},
  journal={arXiv preprint arXiv:1412.6980},
  booktitle={ICLR},
  year={2014}
}

@article{post2018call,
  title={A call for clarity in reporting BLEU scores},
  author={Post, Matt},
  journal={WMT},
  year={2018}
}

@article{wang2019aleatoric,
  title={Aleatoric uncertainty estimation with test-time augmentation for medical image segmentation with convolutional neural networks},
  author={Wang, Guotai and Li, Wenqi and Aertsen, Michael and Deprest, Jan and Ourselin, S{\'e}bastien and Vercauteren, Tom},
  journal={Neurocomputing},
  year={2019},
  publisher={Elsevier}
}

@article{kim2022exploring,
  title={Exploring train and test-time augmentations for audio-language learning},
  author={Kim, Eungbeom and Kim, Jinhee and Oh, Yoori and Kim, Kyungsu and Park, Minju and Sim, Jaeheon and Lee, Jinwoo and Lee, Kyogu},
  journal={arXiv preprint arXiv:2210.17143},
  year={2022}
}

@article{kaya2025efficient,
  title={Efficient Test-Time Scaling for Small Vision-Language Models},
  author={Kaya, Mehmet Onurcan and Elliott, Desmond and Papadopoulos, Dim P},
  journal={ICLR},
  year={2026}
}

@inproceedings{guzman2014efficiently,
  title={Efficiently enforcing diversity in multi-output structured prediction},
  author={Guzman-Rivera, Abner and Kohli, Pushmeet and Batra, Dhruv and Rutenbar, Rob},
  booktitle={AISTATS},
  year={2014},
}

@article{bertsekas1997nonlinear,
  title={Nonlinear programming},
  author={Bertsekas, Dimitri P},
  journal={Journal of the Operational Research Society},
  year={1997},
  publisher={Taylor \& Francis}
}

@article{rose2002vector,
  title={Vector quantization by deterministic annealing},
  author={Rose, Kenneth and Gurewitz, Eitan and Fox, Geoffrey C},
  journal={IEEE Transactions on Information theory},
  year={2002},
  publisher={IEEE}
}

@article{cortes1995support,
  title={Support-vector networks},
  author={Cortes, Corinna and Vapnik, Vladimir},
  journal={Machine learning},
  year={1995},
  publisher={Springer}
}

@inproceedings{patel2025styledistance,
  title={Styledistance: Stronger content-independent style embeddings with synthetic parallel examples},
  author={Patel, Ajay and Zhu, Jiacheng and Qiu, Justin and Horvitz, Zachary and Apidianaki, Marianna and McKeown, Kathleen and Callison-Burch, Chris},
  booktitle={NAACL},
  year={2025}
}

@article{beyer2024paligemma,
  title={Paligemma: A versatile 3b vlm for transfer},
  author={Beyer, Lucas and Steiner, Andreas and Pinto, Andr{\'e} Susano and Kolesnikov, Alexander and Wang, Xiao and Salz, Daniel and Neumann, Maxim and Alabdulmohsin, Ibrahim and Tschannen, Michael and Bugliarello, Emanuele and others},
  journal={arXiv preprint arXiv:2407.07726},
  year={2024}
}

@inproceedings{zhang2022magic,
  title={Magic: Multimodal relational graph adversarial inference for diverse and unpaired text-based image captioning},
  author={Zhang, Wenqiao and Shi, Haochen and Guo, Jiannan and Zhang, Shengyu and Cai, Qingpeng and Li, Juncheng and Luo, Sihui and Zhuang, Yueting},
  booktitle={AAAI},
  volume={36},
  year={2022}
}

@inproceedings{xu2021towards,
  title={Towards accurate text-based image captioning with content diversity exploration},
  author={Xu, Guanghui and Niu, Shuaicheng and Tan, Mingkui and Luo, Yucheng and Du, Qing and Wu, Qi},
  booktitle={CVPR},
  year={2021}
}

@article{zhu2025diffusion,
  title={Diffusion-based diverse audio captioning with retrieval-guided Langevin dynamics},
  author={Zhu, Yonggang and Men, Aidong and Xiao, Li},
  journal={Information Fusion},
  year={2025},
  publisher={Elsevier}
}

@inproceedings{tromble2008lattice,
  title={Lattice Minimum Bayes-Risk decoding for statistical machine translation},
  author={Tromble, Roy and Kumar, Shankar and Och, Franz Josef and Macherey, Wolfgang},
  booktitle={EMNLP},
  year={2008}
}

@article{dempster1977maximum,
  title={Maximum likelihood from incomplete data via the EM algorithm},
  author={Dempster, Arthur P and Laird, Nan M and Rubin, Donald B},
  journal={Journal of the royal statistical society},
  year={1977},
  publisher={Wiley Online Library}
}
\bibliographystyle{icml2026}

\newpage
\appendix
\onecolumn
\newpage
\appendix
\onecolumn

\startcontents[app]
{\hypersetup{linkcolor=black}
  \section*{Table of Contents}
  \printcontents[app]{l}{1}{\setcounter{tocdepth}{3}}
}

\newpage

\section{Notations and setup}
\label{app:notation}

In the following, let $x \triangleq (x_{t})_{t=1}^{T} \in \mathcal{V}^{T}$ be a sequence of $T$ tokens belonging to a finite vocabulary $\mathcal{V} = \{1,\dots,|\mathcal{V}|\}$, and $c \triangleq (c_{t})_{t=1}^{\tau} \in \cC$ be a sequence of $\tau$ context embeddings of dimension $d$. In the following $\cX = \mathcal{V}^{T}$ and $\cC = (\mathbb{R}^{d})^{\tau}$.

Language modeling aims at learning the law $p(x \mid c) = \prod_{t=1}^{T} p(x_t \mid x_{<t}, c)$ using a model $p_{\theta}$ with parameters $\theta \in \Theta$, by maximum likelihood estimation, minimizing the negative log-likelihood loss
 \begin{equation}
 \mathcal{L}(\theta) = - \mathbb{E}_{c,x}[\log p_{\theta}(x \mid c)] = \mathbb{E}_{c,x}\left[- \sum_{t=1}^{T} \log p_{\theta}(x_t \mid x_{<t},c) \right]\;,
 \label{eqapx:teacher_forcing}
 \end{equation} where $x_{< t}$ denotes the sequence of tokens prior to time $t$. In practice, we assume that $p_{\theta} = s_{\eta} \circ f_{\theta}$, where $f_{\theta}(x_{<t},c) \in \mathbb{R}^{\mathcal{V}}$ are the predicted logits and $s_\eta: z \in \mathbb{R}^{\mathcal{V}} \mapsto \left( \frac{\mathrm{exp}(z_j/\eta)}{\sum_{q = 1}^{|\mathcal{V}|} \mathrm{exp}(z_q/\eta)} \right) \in [0,1]^{\mathcal{V}}$ is the $\mathrm{softmax}$ operator with temperature $\eta > 0$.

In the following, we make the following assumption.

\begin{asm}[Expressiveness]
    In the following, we assume that the model $p_{\theta}$ is perfectly expressive. Formally, let $\mathcal{F}_{\Theta} \triangleq \{ p_{\theta}: c \in \mathcal{C} \rightarrow p(\cdot \mid c) \in \mathcal{P}(\mathcal{X}) \mid \theta \in \Theta \}$ be the family of conditional distributions realized by the model, where $\mathcal{P}(\mathcal{X})$ is the set of probability distributions on $\mathcal{X}$. We assume the family $ \mathcal{F}_{\Theta}$ is perfectly expressive, that is $\mathcal{F}_{\Theta} = \mathcal{P}(\mathcal{X})^{\mathcal{C}}$.
    \label{asm:expressiveness}
\end{asm}

First, note that we have the Proposition \ref{prop:gibbs}, a well-known result that is due to the Gibbs inequality (e.g., \citet{mackay2003information}), for which we provide a proof for completeness.

\begin{prop}[\citeauthor{mackay2003information}, \citeyear{mackay2003information}]
    Under Assumption \ref{asm:expressiveness}, for the next-token prediction loss \eqref{eqapx:teacher_forcing}, one can show that \begin{equation} 
    \label{eqapx:min_mle}
    \mathrm{min}_{\theta} \; \mathcal{L}(\theta) = \mathcal{H}(x \mid c)\,,\end{equation}where $\mathcal{H}(x \mid c) \triangleq - \mathbb{E}_{c,x} [ \mathrm{log} \; p(x \mid c)]$.
    \label{prop:gibbs}
\end{prop}
\begin{proof}
    Let us denote $\mathscr{S}(p) = \{ (x,c) \mid p(x,c) > 0\}$ the support of $p$. We use the convention $p(x,c) \, \mathrm{log} \, p(x,c) = 0$ for $(x,c) \in \cX \times \cC - \mathscr{S}(p)$.
    Because $\mathrm{log}$ is a concave function, we have under Jensen's inequality:
    \begin{equation}- \int_{\mathscr{S}(p)} \mathrm{log} \left[ \frac{p_{\theta}(x \mid c)}{p(x,c)} \right] p(x,c) \; \mathrm{d}x \mathrm{d}c \geq - \mathrm{log} \left( \int_{\mathscr{S}(p)} \frac{p_{\theta}(x \mid c)}{p(x,c)} p(x,c) \; \mathrm{d}x \mathrm{d}c \right) \geq 0 \;.
    \label{eq:jensen}
    \end{equation}
    However, because of the convention, the left-hand side of \eqref{eq:jensen} is also equal to the integral over $\cX \times \cC$. This shows: 
    \begin{equation*}
        - \int_{\cX \times \cC} p(x,c) \; \mathrm{log} \; p_{\theta}(x \mid c) \; \mathrm{d}x \mathrm{d}c \geq - \int_{\cX \times \cC} p(x,c) \; \mathrm{log} \; p(x \mid c)  \; \mathrm{d}x \mathrm{d}c = \mathcal{H}(x \mid c) \;,
\end{equation*}
where the equality is reached for parameter $\theta$ such that $p_{\theta} = p$, whose existence is guaranteed by Assumption \ref{asm:expressiveness}.
\end{proof}

In the following, we denote the Kullback–Leibler divergence between two distributions $\alpha$ and $\beta$ as $\mathrm{KL}(\alpha \,\|\, \beta) \triangleq \int_{\mathscr{S}(\beta)} \alpha(x) \mathrm{log} \frac{\alpha(x)}{\beta(x)} \mathrm{d}x$, where $\mathscr{S}(\beta) \triangleq \{x \in \mathcal{X} \mid \beta(x) > 0\}$. Note that we have the equality: 
\begin{equation}
   - \mathbb{E}_{\alpha}[\mathrm{log} \, \beta(x)] = \mathrm{KL}(\alpha \,\|\, \beta) + \mathcal{H}(\alpha)\;, 
   \label{eq:kl_sum}
\end{equation}
where the left-hand side is usually referred to as the Cross-Entropy, and $\mathcal{H}(\alpha) \triangleq - \mathbb{E}_{\alpha}[\mathrm{log}\, \alpha(x)]$ is the entropy of $\alpha$. 
When the context is clear, we will also write the entropy of a distribution $\alpha$ as $\mathcal{H}(x)$ where $x \sim \alpha$.

\section{Proof of Proposition \ref{prop:em}}
\label{sec:proof_prop:em}

Let us now consider the following assumptions.

\begin{asm}[Mixture of latent processes]
\label{asm:mixture}
    The data-generating process writes in the form $p(x \mid c) = \sum_{k = 1}^{K} p(z_k \mid c) \; p(x \mid z_k, c)$. The Mixture is said to be uniform if $\forall k, p(z_k \mid c) = \frac{1}{K}$.
\end{asm}

\begin{asm}[Minimization of the true risk]
\label{asm:true_risk}
    The batch size is large enough so that the minimization of the empirical risk comes down to minimizing the true risk \eqref{eqapx:teacher_forcing}.
\end{asm}

\begin{rmk}
Under Assumptions \ref{asm:mixture} and \ref{asm:true_risk}, the optimal reachable loss by maximum likelihood estimation is $\min_{\theta} \mathcal{L}(\theta) = \mathbb{E}_{c} \left[ \mathcal{H}(x \mid c) \right]$, where $x \sim \frac{1}{K} \sum_{k=1}^{K} p(x \mid z_k, c)$.
\end{rmk}

\begin{asm}[Disjoint components]
\label{asm:disjoint}
    This assumption states that $p(x \mid c, z_s) = 0$ when $ p(x \mid c, z_k) > 0$, for $s \neq k$.
\end{asm}
\begin{prop}

Under Assumptions \ref{asm:expressiveness}, \ref{asm:mixture}, and \ref{asm:true_risk}, we have that:
     \begin{itemize}
    \item[(i)] The winner-takes-all two-step optimization in \texttt{LoRA-MCL} acts as a conditional form of the hard-EM algorithm.
    \item[(ii)] Under Assumption \ref{asm:disjoint}, and assuming (with one permutation) that $p(x \mid z_k, c) = p(x \mid c ; \theta_k)$ for each $k$, $\mathcal{L}^{\mathrm{WTA}}(\theta) = - \mathbb{E}_{x, c} \left[ \underset{k = 1,\dots,K}{\mathrm{max}} \mathrm{log} \, p(x \mid c, z_k)\right]$. In this case, we also have: 
    \begin{equation}
    \label{apxeq:opt}
    \mathcal{L}^{\mathrm{WTA}}(\theta) = \mathcal{H}(x \mid c,z) \triangleq \mathbb{E}_{c} \left[ \sum_{k=1}^{K} p(z_k \mid c) \mathcal{H}\big(x \mid c, z_k\big) \right]\,,\end{equation}
    where $\mathcal{H}(x \mid c,z)$ is the conditional entropy given the random variable $z$.
    \item[(iii)] We have the following inequalities:
    \begin{equation}
    \label{apxeq:inequality1}
    \mathrm{min}_{\theta} \, \mathcal{L}(\theta) - \mathrm{log}\,K \overset{(a)}{\leq} \mathrm{min}_{\theta} \, \mathcal{L}^{\mathrm{WTA}}(\theta) \overset{(b)}{\leq} \mathcal{H}(x \mid c,z) \overset{(c)}{\leq} \mathrm{min}_{\theta} \, \mathcal{L}(\theta)\;,
    \end{equation}
    where $\mathrm{min}_{\theta} \; \mathcal{L}(\theta) = \mathcal{H}(x \mid c)$.
    \end{itemize}
\end{prop}

\textit{Proof of (i)}
    First, let us remind that the hard-EM consists of fitting a distribution $p_{\theta}(x,z)$ to observed data $x \sim p(x)$ where $z$ are (unknown) hidden variables.
    The fitting starts from randomly initialized parameters $\theta$ and latent variables $z$. It consists of repeating the following operations at each iteration $t$ until convergence:
    \begin{enumerate}
        \item (Expectation) $z_k^{\star} = \mathrm{argmax}_{k} \; p(x,z_k ; \theta^{(t)})$
        \item (Maximization) $\theta^{(t+1)} = \mathrm{argmax}_{\theta}  \; p(x,z_{k}^{\star} ; \theta^{(t})$
    \end{enumerate}
    Let us define: 
    \begin{equation}
        D(\theta,q) \triangleq \int_{\cX} \sum_{k=1}^{K} q(k \mid x) \mathrm{log}\, p(x,z_k;\theta)\;\mathrm{d}p(x),~D(\theta) \triangleq \int_{\cX} \max_{k = 1,\dots,K} \;  \mathrm{log}\,p(x,z_k;\theta)\;\mathrm{d}p(x)\;,
    \end{equation}
    where $q$ is a discrete distribution over $\{1,\dots,K\}$ with exactly one non-zero component that controls the assignment of each $x$ to a fixed $k^{\star}$. Let us define $q(\theta)$ as the discrete distribution defined so that $q(k \mid x; \theta) \triangleq \mathbf{1}[k = \mathrm{argmax}_s\; p(x,z_s;\theta)]$. Note that then $D(\theta) = D\left(\theta,q(\theta)\right)$. 
    
    For the vanilla (or soft) EM algorithm, the complete data log-likelihood $\int_{\cX} \mathrm{log} \, p(x; \theta) p(x) \mathrm{d}x$ is expected to increase at each iteration $t$. Similarly, for the hard-EM, we have that the $D(\theta)$ increases at each iteration.
    
    Indeed, the expectation step comes down to computing $q(\theta^{(t)})$. For the Maximization step, we have: $D\left(\theta^{(t+1)},q(\theta^{(t)})\right) \geq D\left(\theta^{(t)},q(\theta^{(t)})\right)$, by definition. At the next expectation step, we have: $D\left(\theta^{(t+1)},q(\theta^{(t+1)})\right) \geq D\left(\theta^{(t+1)},q(\theta^{(t)})\right)$, because $q(\theta^{(t+1)})$ computes the best assignment given the parameters $\theta^{(t+1)}$. This shows that $D(\theta^{(t+1}) \geq D(\theta^{(t)})$.

    The main difference compared to the vanilla form of the (hard) EM algorithm is that $(i)$ the goal here is to fit a \textit{conditional} distribution $p(x \mid c)$ given pairs $(c, x) \sim p(c, x)$, and $(ii)$ step 2 performs a gradient update (of the neural network weights) instead of a full maximization.
    
    Note that under Assumption \ref{asm:mixture} the complete data log-likelihood writes as $\mathrm{log} \, p(x \mid c ; \theta) = \mathrm{log} \left[ \; \sum_{k = 1}^{K} p(x, z_k \mid c  ; \theta) \right]$.
    However, the variables $z_k$ are not known in practice. \texttt{LoRA-MCL} works by analogy with the Hard-EM algorithm, which consists, in the Expectation step, of picking for each pair $(x;c)$, $k^{\star}(x,c) = \mathrm{argmax}_{k} \; p(x \mid c  ; \theta_k)$.
    Indeed, under Assumption \ref{asm:true_risk}, each training step of \texttt{LoRA-MCL} writes as the optimization of
    \begin{equation}
    \label{eq:wta_risk}
    \mathcal{L}^{\mathrm{WTA}}(\theta) = - \int_{\cX \times \cC} \max_{k = 1,\dots,K} \mathrm{log} \; p(x \mid c; \theta_k) \mathrm{d} p (c,x)\;.
    \end{equation}
    The loss is generally expected to \textit{decrease} across training iterations, although strict monotonicity is not guaranteed without additional assumptions on the learning rate and the smoothness of the gradient of the loss \cite{bertsekas1997nonlinear}. Since the loss is bounded below (by $0$), the sequence of loss values $\{\mathcal{L}^{\mathrm{WTA}}(\theta^{(t)})\}_{t \ge 0}$ is therefore expected to converge in practice.
    
    To conclude, we can view $(\theta_1, \dots, \theta_K)$ as the parameters involved in the estimation of the modes of the conditional distribution with $p(x \mid c) = \sum_{k=1}^{K} p(x \mid c ; \theta_k) p(\theta_k \mid c)$. Note that the current form of the algorithm does not estimate the weight of each mode
    $p(\theta_k \mid c)$, further work could include incorporating \textit{scoring} heads to estimate $p(\theta_k \mid c)$ each $k$ as in \citet{letzelter2023resilient}. \qed

    We then expect to be able to recover the distributions (with one permutation) $\{p(x \mid c, z_k)\}$ from estimated $\{p(x \mid c ; \theta_k)\}$, assuming identifiability of the data generating mixture, which we expect to be made easier if the components are enough separated (e.g., \citet{redner1984mixture} Par.\,2.5 or \citet{mclachlan2019finite} Sec.\,2.2).

\textit{Proof of (ii)} Let us assume that (with one permutation) $p(x \mid z_k, c) = p(x \mid c ; \theta_k)$ for $k \in \{1,\dots,K\}$. This is possible thanks to Assumption \ref{asm:expressiveness}. Let us show that \eqref{apxeq:opt}.  
Let us define: 
\begin{equation}
    \cX_{k}(c, \theta) \triangleq \Big\{ x \in \cX \mid \mathrm{log} \; p(x \mid c, \theta_k) \geq \mathrm{log} \; p(x \mid c, \theta_s) \; \forall s \in \{1,\dots,K\} \Big\}\;.
\end{equation}
In this case, the WTA loss \eqref{eq:wta_loss} writes
as 
\begin{align*}
    \mathcal{L}^{\mathrm{WTA}}(\theta) &= - \int_{\cC} \sum_{k=1}^{K} \int_{\cX_{k}(c, \theta)} \mathrm{log} \; p(x \mid c ; \theta_k) \; p(x \mid c) \; \mathrm{d}x \; p(c)  \mathrm{d}c \\
    &= - \int_{\cC} \sum_{k=1}^{K} \sum_{s=1}^{K} \int_{\cX_{k}(c, \theta)} \mathrm{log} \; p(x \mid c ; z_k) \; p(x \mid c ; z_s) \;  p(z_s \mid c) \; \mathrm{d}x \; p(c) \; \mathrm{d}c\\
    &= - \int_{\cC} \sum_{k=1}^{K} \int_{\cX_{k}(c, \theta)} \mathrm{log} \; p(x \mid c ; z_k) \; p(x \mid c ; z_k) \; p(z_k \mid c) \; \mathrm{d}x \; p(c) \; \mathrm{d}c \text{\quad by Asm. \ref{asm:disjoint}} \\
    &= - \int_{\cC} \sum_{k=1}^{K} \mathcal{H}( x \mid c ; z_k ) \; p(z_k \mid c) \; p(c) \mathrm{d}x \; \mathrm{d}c \text{\quad as $\int_{\cX_{k}(c, \theta)} p(x \mid c ; z_k) \mathrm{d}x = 1$.} \\
    &= \mathbb{E}_{c} \left[ \sum_{k=1}^{K} p(z_k \mid c) \mathcal{H}\big(x \mid c; z_k\big) \right]\;.
\end{align*}
\qed

\textit{Proof of (iii)} Let us show that:
\begin{equation*}
    \mathrm{min}_{\theta} \, \mathcal{L}(\theta) - \mathrm{log}\,K \overset{(a)}{\leq} \mathrm{min}_{\theta} \, \mathcal{L}^{\mathrm{WTA}}(\theta) \overset{(b)}{\leq} \mathcal{H}(x \mid c,z) \overset{(c)}{\leq} \mathrm{min}_{\theta} \, \mathcal{L}(\theta)\;.
\end{equation*}

$(a)$: First, we have $\underset{k = 1,\dots,K}{\mathrm{max}} \, p(x \mid c, z_k) \leq \sum_{k=1}^{K} p(x \mid c, \theta_k)$. Therefore, 
\begin{align*}
    \mathcal{L}^{\mathrm{WTA}}(\theta) &\geq - \mathbb{E}_{x,c} \left[\mathrm{log} \frac{1}{K} \sum_{k=1}^{K} p(x \mid c, \theta_k)\right] - \mathrm{log}\,K\\
    &= \underbrace{\mathrm{KL}\left[ p(x \mid c) \,\|\, \frac{1}{K} \sum_{k=1}^{K} p(x \mid c, \theta_k) \right]}_{\geq 0} + \mathcal{H} \left( x \mid c \right) - \mathrm{log}\,K \text{\quad by \eqref{eq:kl_sum}}\\
    &\geq \mathcal{H} \left( x \mid c \right) - \mathrm{log}\,K\,.
\end{align*}
Because $\mathrm{min}_{\theta} \, \mathcal{L}(\theta) = \mathcal{H} \left( x \mid c \right)$, we have shown that $(a)$ occurs when the KL term vanishes, which is when $p(x \mid c) = \frac{1}{K} \sum_{k=1}^{K} p(x \mid c, \theta_k)$.

$(b)$: We have \begin{align*}
\mathcal{L}^{\mathrm{WTA}}(\theta) &= - \mathbb{E}_{x} [ \underset{k = 1,\dots,K}{\mathrm{max}} \mathrm{log} \, p(x \mid c, \theta_k) ]\\  
&= - \mathbb{E}_{z} \mathbb{E}_{x \mid z} \left[\mathrm{log} \frac{\underset{k = 1,\dots,K}{\mathrm{max}} \, p(x \mid c, \theta_k)}{p(x \mid c, z)} \right] \underbrace{- \mathbb{E}_{z} \mathbb{E}_{x \mid z} \left[ \mathrm{log} \, p(x \mid c, z)\right]}_{\mathcal{H}(x \mid c, z)}\,.
\end{align*}

Now let us leverage Assumption \ref{asm:expressiveness} to choose $\tilde{\theta}_k$ such that $p(x \mid c, \tilde{\theta}_k) = p(x \mid c, z_k)$ for each $k \in \{1, \dots, K\}$. In this case, $\mathrm{max}_k \, p(x \mid c, \tilde{\theta}_k) \geq p(x \mid c, z)$ for each $z \in \{z_{1},\dots,z_{K}\}$, and $- \mathbb{E}_{z} \mathbb{E}_{x \mid z} \left[\mathrm{log} \frac{\mathrm{max}_k \, p(x \mid c, \tilde{\theta}_k)}{p(x \mid c, z)} \right] \leq 0$. Then, $$\mathrm{min}_{\theta}\, \mathcal{L}^{\mathrm{WTA}}(\theta) \leq \mathcal{L}(\tilde{\theta}_1,\dots,\tilde{\theta}_K) \leq \mathcal{H}(x \mid c,z)\;,$$ which proves $(b)$.

Finally $(c)$ can be directly deduced from the inequality $\mathcal{H}(x \mid c,z) \leq \mathcal{H}(x \mid c)$.\qed

\section{Proof of Corollary \ref{corr:mc}}
\label{apx:proof_mc}

Let us consider the following assumptions.

\begin{asm}[Markov Chain]
    We assume that the data-generating process can be written as a uniform mixture of Markov chains of order $n \in \mathbb{N} \setminus \{0\}$, that is, for each $t$ and each $k$, $p(x_{t} \mid x_{<t}, c, z_{k}) = p(x_{t} \mid x_{t-1}, \dots, x_{t-n}, c, z_{k})$.
    \label{asm:mc}
\end{asm}

\begin{corr}
    As per Assumption \ref{asm:mc}, let us assume that the data-generating process writes as a uniform mixture of Markov chains of order $n = 1$. Let $\hat{P}(\theta) \triangleq (p(x_{t+1} = j \mid x_{t} = i))_{i,j}$ be the predicted transition matrix when using a language model with parameters $\theta$. Under the same assumptions that in Proposition \ref{prop:em}, we have:
     \begin{itemize}
     \item[(i)] Whenever the maximum likelihood estimator trained with next-token-prediction \eqref{eqapx:teacher_forcing} reaches its optimal loss $\mathcal{L}(\theta)$, we have $$\hat{P}(\theta)_{i,j} = \sum_{k=1}^{K} p(z = z_{k} \mid x_{t} = i) (P_k)_{i,j} = \frac{1}{\sum_{s=1}^{K} (\pi_s)_i} \sum_{k=1}^{K} (\pi_k)_i (P_k)_{i,j}\;,$$
     where $\statk \in [0,1]^{\mathcal{V}}$ is the stationary distribution of $P_k$.
    \item[(ii)] The inequality \eqref{eq:inequality1} holds in this context, where the conditional entropy $\mathcal{H}(x \mid z)$ can be computed by a weighted sum of the \textit{entropy rate} of each of the $K$ Markov Chains: 
    \begin{equation}
    \label{eq:mixture_entropy_rates}
    \mathcal{H}(x \mid z) = - \textcolor{black}{T} \sum_{k=1}^{K}  \sum_{i = 1}^{|\mathcal{V}|} (\statk)_{i} \Big[ \sum_{j=1}^{|\mathcal{V}|} (P_k)_{i,j} \; \mathrm{log} (P_k)_{i,j} \Big] \,.
    \end{equation}
    The entropy $\mathcal{H}(x)$, which is the lower bound of the MLE baseline, can be computed either exactly for short sequences or approximated, e.g., through Monte-Carlo integration.
    \end{itemize}
\end{corr}

\textit{Proof of (i)} The Cross entropy of the maximum-likelihood model is optimal whenever $p_{\theta} = p$.

In this case, for each $i,j \in \{1,\dots,|\mathcal{V}|\}$, we have $p_{\theta}(x_{t+1} = j \mid x_{t} = i) = p(x_{t+1} = j \mid x_{t} = i)$ and 
\begin{equation}
p_{\theta}(x_{t+1}= j \mid x_{t} = i) = \sum_{k=1}^{K} p(z = z_k \mid x_{t}=i) \, p(x_{t+1} = j \mid x_{t} = i, z_k)\,.
\end{equation} 
Furthermore, by Bayes' rule, we have:
$$p(z = z_k \mid x_{t}=i) = \frac{p(x_{t}=i \mid z = z_k) p(z = z_k)}{\sum_{s=1}^{K} p(x_{t}=i \mid z = z_s) p(z = z_s)} = \frac{(\pi_k)_i}{\sum_{s=1}^{K} (\pi_s)_i}\;,$$
because we have assumed a uniform prior over the mixture components, i.e., $p(z = z_k) = \frac{1}{K}$, and we assumed a stationary regime so that $p(x_{t}=i \mid z = z_k) = (\pi_k)_i$.
\qed

\textit{Proof of (ii)} Because we assumed first-order Markov Chains, we have $$\mathcal{H}(x) = \sum_{t=1}^{T} \mathcal{H}(x_t \mid x_{<t})\,.$$ Because we assumed stationary Markov chains, we have that $\mathcal{H}(x_t \mid x_{t-1})$ doesn't depend on $t$ and $\mathcal{H}(x_t \mid x_{t-1}) = \mathcal{H}(x_2 \mid x_{1}) = - \sum_{i = 1}^{|\mathcal{V}|} (\statk)_{i} \Big[ \sum_{j=1}^{|\mathcal{V}|} (P_k)_{i,j} \; \mathrm{log} (P_k)_{i,j} \Big] \,$ (see \citet{cover1999elements} page 66, Theorem 4.2.4).
\qed

Note that we considered first-order Markov chains in our analysis. However we expect the properties to generalize to higher orders ($n > 1$) by using transition matrices in the form $P \in [0,1]^{\mathcal{V}^{n+1}}$ where $$P_{i_{1}, \dots, i_{n+1}} = p(x_{t+1} = i_{n+1} \mid x_{t} = i_{n}, \dots, x_{t-n+1} = i_{1})\;.$$

\section{Experimental details of Section \ref{sec:synt_data}}
\label{apx:synt_data}

\begin{figure}
    \centering
    \includegraphics[width=0.9\linewidth]{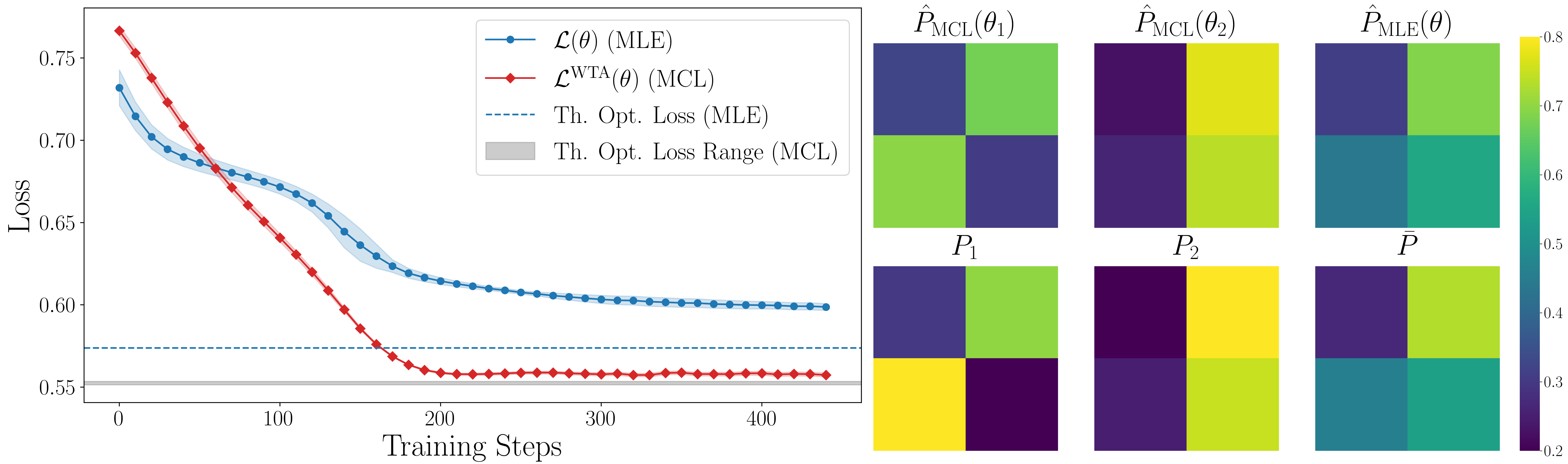}
    \caption{\textbf{Comparison of \texttt{LoRA-MCL} with standard maximum likelihood estimation (MLE).} The setup mirrors that of Figure \ref{fig:illustration_markov}, but uses different transition matrices. While the overall behavior remains consistent with Figure \ref{fig:illustration_markov}, we observe a distinct transition in the MLE loss. We interpret this as evidence that MLE increasingly incorporates contextual information as training progresses (see also \citep{makkuva2024attention}).}
    \label{fig2:illustration_markov}
\end{figure}

The results in Figure \ref{fig:illustration_markov} were obtained using the setup described in this section. The same illustration with different transition matrices is shown in Figure \ref{fig2:illustration_markov}. Note that in both figures, the loss and theoretical quantities are normalized by $T-1$, since the first token is excluded from the computation.

\textbf{Dataset.} We used a uniform mixture of two first-order homogeneous Markov chains with transition matrices $(P_{1}, P_{2}) = \big(P(p_1,q_1), P(p_{2},q_{2})\big)$, with $P(p,q) \triangleq \left[\begin{array}{cc}
1-p & p \\
q & 1-q
\end{array}\right]\;$ and $p, q \in [0,1]$. For the experiment in Figure \ref{fig:illustration_markov}, we used $(p_1,q_1)=(0.2,0.9)$ and $(p_2,q_2)=(0.8,0.25)$. Figure \ref{fig2:illustration_markov} uses $(p_1,q_1)=(0.7,0.8)$ and $(p_2,q_2)=(0.8,0.25)$. The first state of the sequences sampled according to the stationary distribution of the Markov chain given by $\pi_k = \frac{1}{p_k+q_k} (q_k,p_k)$ (see e.g., \citet{makkuva2024attention}). The sequences have a fixed length of $T = 32$.

\textbf{Architecture.} We considered a GPT-2-like architecture \citep{radford2019language} using the GPT-Neo implementation \citep{black2021gpt} using local-attention suggested by \citet{makkuva2024attention} to improve convergence on Markov chain data (with window size of $5$). The model has a hidden size of $64$, $2$ layers of transformer blocks with $2$-heads attention. LoRA adapters for the MLE baseline have rank $r = 64$, $\alpha = 64$, and dropout disabled. To align the numbers of parameters when $K=2$, we used $r = 32$ (and $\alpha = 32$) for \texttt{LoRA-MCL}. The models have a total of $65{,}536$ trainable parameters over a total of $230{,}912$. Note that aligning the ranks of \texttt{LoRA-MLE} and \texttt{LoRA-MCL} leads to the same conclusions. Weights of the base model were kept frozen (both for $\texttt{LoRA-MLE}$ and $\texttt{LoRA-MCL}$) to mimic the dynamics on larger language models.

\textbf{Training details.} We used a cosine scheduler with learning rate of $10^{-4}$, weight decay of $10^{-3}$ with AdamW optimizer, with $(\beta_1,\beta_2)=(0.9,0.95)$ as in \citet{makkuva2024attention}. We used a batch size of $128$, and trained for $500$ iterations, with validation loss computed every $10$ steps. Only the first $450$ iterations are plotted in the Figures \ref{fig:illustration_markov} and \ref{fig2:illustration_markov}. For the $\texttt{LoRA-MCL}$ runs, we trained with the vanilla winner-takes-all update. Figure~\ref{fig:illustration_markov} (left) shows the mean and standard deviation of the winner-takes-all validation loss across three training seeds, for $K = 1$ (\texttt{LoRA-MLE}) and $K = 2$ (\texttt{LoRA-MCL}).

\textbf{Theoretical quantities computations} To verify the theoretical results, we computed the following quantities:
\begin{itemize}
    \item \textbf{Theoretical Optimal Loss of the MLE model.} It is expressed as $\mathcal{H}(x)$ where $x \sim \frac{1}{K} \sum_{k=1}^{K} p(x \mid z_k)$. It can be approximated by Monte-Carlo sampling with samples $(z_s,x_s)$ by first sampling modes $z_s \sim \mathcal{U}\{1,\dots,K\}$, then $x_s \mid z_s \sim p(x \mid z_s)$, and by computing
    \begin{equation}- \frac{1}{N} \sum_{s=1}^{N} \sum_{t=2}^{T} \mathrm{log} \, (P_s)_{x_{s,t},x_{s,t+1}} \,.
    \label{eq:practical}
    \end{equation} 
    Here, $(P_s)_{x_{s,t},x_{s,t+1}}$ denotes the entry of $P_s$ at row $x_{s,t}$ and column $x_{s,t+1}$, and $N$ is the total number of samples. We used $N = 50{,}000$ here. Note that the term corresponding to $t = 1$ in \eqref{eq:practical} is discarded, as the first token is typically excluded from the loss computation. For the illustration, \eqref{eq:practical} was normalized by $T-1$, since the first token is discarded.
    
    \item \textbf{Theoretical Optimal Loss of the MCL model.} It is computed by subtracting $\mathrm{log} K$ to the Theoretical optimal loss \eqref{eq:practical} of the MLE model.
    To verify $(iii)$ in Proposition \ref{prop:em}, we also computed the mixture of entropy rates given by \eqref{eq:mixture_entropy_rates}, using a normalization factor of $T - 1$.
\end{itemize}

\section{Experimental details on captioning and translation tasks}
\label{apx:captioning}
\subsection{Setup}

Dataset statistics are provided in Table \ref{tab:audio_data_stat}.

\textbf{Audio Datasets.} We conducted experiments on two audio captioning datasets: Clotho-V2 \citep{font2013freesound, drossos2020clotho} and AudioCaps \citep{gemmeke2017audio, kim2019audiocaps}. For both datasets, we used the official training, validation, and test splits. Clotho-V2 provides five reference captions per audio clip across all splits, whereas AudioCaps includes a single caption per clip in the training set and five captions per clip in the validation and test sets. During training on Clotho-V2, which is performed on $10$ epochs, at each epoch, one of the five reference captions is sampled uniformly at random for each audio clip.

\textbf{Vision datasets.} We conducted experiments on the \href{https://textvqa.org/textcaps/download/}{TextCaps} dataset. Specifically, we used the official training split for training and the official validation split as our test set. Since each image in the dataset is annotated with five different captions, we duplicated each image five times—associating each duplicate with a distinct caption—to ensure that all reference captions are seen during a single training epoch.

\textbf{Preprocessing of audio data.} Following the implementation of \citet{labb2024conette}, for Clotho and AudioCaps raw audio files are resampled from $44.1\, \mathrm{kHz}$ and $32\, \mathrm{kHz}$ respectively to $16 \, \mathrm{kHz}$. They are then cropped to a maximum length of 30 seconds for Clotho and 10 seconds for AudioCaps. The data is then fed to the Qwen-2-Audio pipeline, which includes conversion of the raw waveform into a $128$-channel mel-spectrogram, with a window size of $25$ ms and a hop size of $10$ ms. During training, the language model processes sequences formatted as:

\textit{``<Audio bos token><Audio tokens><Audio eos token>Generate the caption in English:<reference text><text eos>''},

where \textit{<Audio bos token>} and \textit{<Audio eos token>} denote the beginning and end of the audio sequence, \textit{<Audio tokens>} correspond to the audio features, \textit{<reference text>} to the target caption (uniformly sampled from the available reference captions during training), and \textit{<text eos>} to the end-of-sentence token. The loss is computed from the index of the first \textit{<reference text>} token and includes the \textit{<text eos>} token. During inference, the same formatting is used (except that the part \textit{<reference text><text eos>} is discarded). This procedure follows the official \href{https://huggingface.co/docs/transformers/main/en/model_doc/qwen2_audio}{Qwen-2-Audio documentation} in the Transformers library. 

\textbf{Preprocessing of image data.}
Our image pre-processing pipeline follows the recipe of LLaVA (i.e., resizing to $(336,336)$ and normalization using CLIP mean and standard deviation).

For both modalities, we used the HuggingFace transformers \citep{wolf2019huggingface} and PEFT \citep{peft} Python libraries as part of the implementation. Note that for each of the experiments, we set the repetition penalty \citep{keskar2019ctrl} to $1.1$ for decoding.

\begin{table}[ht]
    \centering
    \caption{\textbf{Statistics of the audio and image captioning datasets.} Num. samples includes \{train, validation, test\} sets, except for AudioCaps, where we used only \{train, validation\} sets.}
    \vspace{0.5em}
    \resizebox{0.8\columnwidth}{!}{
    \begin{tabular}{ccccc}
\toprule Dataset & Num. samples & Duration (h) & Num. Captions & Modality\\
\midrule AudioCaps \citep{kim2019audiocaps} & 
48,286
& 
134.1
& 
54 K
& Audio\\
Clotho \citep{drossos2020clotho} & 5,929 & 37.0 & 30 K & Audio\\
TextCaps \citep{sidorov2020textcaps} & 
25,119
& N/A & 
126 K
& Image
\\
\bottomrule
\end{tabular}}
    \label{tab:audio_data_stat}
\end{table}

\subsection{Metrics}

In the following, we describe how to assess a language model in generating sequences $\hat{x}^{1}, \dots, \hat{x}^{K}$ conditioned on a context $c$ (for instance, an audio recording paired with a captioning prompt in the case of Audio Captioning) using a given decoding method. In the case of \texttt{LoRA-MCL}, we denote by $\theta_1, \dots, \theta_K$ the parameters of the language models corresponding to each hypothesis. We assume access to a set of $R \geq 1$ references $x^{1}, \dots, x^{R}$ for each context, which can be regarded as samples from the \textit{ground-truth} conditional distribution $p(x \mid c)$. The evaluation is performed on a dataset of $N$ pairs $(c_i, \{x_{i}^1, \dots, x_{i}^{R}\})_{i=1}^{N}$.

\subsubsection{Negative log-likelihood}

\textbf{Test NLL and Perplexity ($\downarrow$).} A standard way to evaluate a language model is through its test loss (e.g., \citep{xiong2024efficient}), which measures the average likelihood of the reference sentences under the trained model. When considering multiple language models, the oracle NLL is defined by averaging, for each reference, the best NLL across the $K$ hypotheses:
\begin{equation}
\mathrm{NLL} \triangleq - \frac{1}{N R} \sum_{i=1}^{N} \sum_{x \in \{ x_{i}^{1}, \dots, x_{i}^{R}\}} \underset{k = 1,\dots,K}{\mathrm{max}} \;\frac{1}{T(x)} \sum_{t=1}^{T(x)} \log p(x_{t} \mid x_{<t}, c_i, \theta_k)\;,
\label{eqapx:metric_nll}
\end{equation}
where $T(x)$ denotes the length of the reference sequence $x$ (in number of tokens), and $\mathrm{log}$ refers to the natural logarithm. In the context of LLMs, perplexity \citep{jelinek1977perplexity} is usually defined as $\mathrm{PPL} = \exp(\mathrm{NLL})$. It can be interpreted as the effective number of equally likely tokens among which the model is uncertain when predicting the next token. In particular, if the model always predicts a uniform distribution over the vocabulary $\mathcal{V}$ then $\mathrm{PPL}=|\mathcal{V}|$.

\subsubsection{Natural language generation quality metrics}

Originally developed within the human translation community, BLEU (Bilingual Evaluation Understudy) \citep{papineni2002bleu}, ROUGE (Recall-Oriented Understudy for Gisting Evaluation) \citep{lin2004rouge}, and METEOR (Metric for Evaluation of Translation with Explicit Ordering) \citep{banerjee2005meteor} were introduced to measure the closeness between a machine translation and a professional human translation. For consistency, all sequences are tokenized using the Penn Treebank Tokenizer (\href{https://nlp.stanford.edu/nlp/javadoc/javanlp/edu/stanford/nlp/process/PTBTokenizer.html}{PTB
 Tokenizer}). Implementations rely on the \href{https://github.com/open-aac/aac-metrics}{AAC
 Metrics} and \href{https://github.com/tylin/coco-caption}{COCO
 Caption} libraries. An $n$-gram refers to a group of $n$ consecutive tokens in a tokenized sequence. When comparing candidate and reference sentences, we use the $F_{\beta}$ score, which balances precision and recall: 
 \begin{equation}F_{\beta}(\mathrm{precision},\mathrm{recall}) \triangleq \frac{(1+\beta^{2}) \mathrm{precision} \cdot \mathrm{recall}}{\beta^{2} \mathrm{precision} + \mathrm{recall}}\;.\end{equation}

Each metric $\mathcal{M}$ is defined as a function of a candidate $\hat{x}^k$ and a set of references $X_i = \{x_i^1, \dots, x_i^R\}$. However, these metrics do not natively handle multiple candidates (see e.g., \citet{lee2016stochastic}[Sec.,4.3]; \citet{labbe2022my}). We therefore adopt a sentence-based oracle evaluation, where the final score is computed as
\begin{equation}
\mathcal{M}_{\mathrm{Oracle}} = \frac{1}{N} \sum_{i=1}^{N} \underset{k = 1, \dots, K}{\mathrm{max}} \mathcal{M}(\hat{x}^{k}_{i},X_i)\;.
\label{eq:oracle}
\end{equation}

\textbf{BLEU ($\uparrow$).} For a given $n$, the modified $n$-gram precision $p_n$ (see Section 2.1 in \citet{papineni2002bleu}), measures the fraction of the unigrams in a predicted caption that also appear in the reference captions, while clipping the numerator by the maximum number of times a word occurs in any single reference translation. $\mathrm{BLEU}_{n}$ combines the n-grams precision up to length $n$ computing a geometric mean of  $p_s$ for $s=1,\dots,n$. It also applies a multiplicative brevity penalty $\mathrm{BP}$ to penalize too short predicted captions compared to the reference sequences lengths. In this case, we have $\mathcal{M} = \mathrm{BLEU}_{n}$ and:
\begin{equation}\mathrm{BLEU}_{n}(\hat{x}, X) \triangleq \mathrm{BP}(\hat{x}, X) \times \left(\prod_{s=1}^{n} p_s(\hat{x}, X)\right)^{\frac{1}{n}}\;,\end{equation}
where  $\mathrm{BP}(\cdot,\cdot)$ and $p_{s}(\cdot,\cdot)$ are functions of the candidates and the set of references (please refer to Section 2.3 of \citet{papineni2002bleu}, and the \href{https://aac-metrics.readthedocs.io/en/stable/aac_metrics.classes.bleu.html}{documentation} for details).

\textbf{ROUGE ($\uparrow$).} The ROUGE family of metrics \citep{lin2004rouge} was originally introduced for the automatic evaluation of summaries, following BLEU, and is based on measuring $n$-gram co-occurrence between candidate and reference sequences. In this work, we use $\mathrm{ROUGE}_{L}$, which relies on the Longest Common Subsequence (LCS) between a candidate and a reference. Formally, we set $\mathcal{M} = \mathrm{ROUGE}_{L}$ with
\begin{equation}\mathrm{ROUGE}_{L}(\hat{x}, X) \triangleq \underset{x^{r} \in X}{\mathrm{max}} \; F_{\beta}(P_{\mathrm{LCS}}(\hat{x},x^{r}),R_{\mathrm{LCS}}(\hat{x},x^{r}))\;,\end{equation}
where the precision and recall are defined as:
$$P_{\mathrm{LCS}}(\hat{x},x^{r}) = \frac{\mathrm{LCS}(\hat{x},x^{r})}{\mathrm{length}(\hat{x})},\; R_{\mathrm{LCS}}(\hat{x},x^{r}) = \frac{\mathrm{LCS}(\hat{x},x^{r})}{\mathrm{length}(x^r)}\;,$$ and $\beta = 1.2$ by default.

\textbf{METEOR ($\uparrow$).} Unlike BLEU, METEOR \citep{banerjee2005meteor} explicitly incorporates recall, measures word-level matches between a candidate and the reference, and accounts for grammaticality through word order. Here we set $\mathcal{M} = \mathrm{METEOR}$, defined as an $F_{\beta}$ score between precision and recall, with an additional fragmentation penalty that lowers the score when matching words are not in the correct order:
\begin{equation}
\mathrm{METEOR}(\hat{x}, X) \triangleq \underset{x^{r} \in X}{\mathrm{max}} \left(1-\mathrm{Penalty}(\hat{x},x^{r})\right) 
F_{\beta}(P(\hat{x},x^{r}),R(\hat{x},x^{r}))\end{equation} where the precision are recall of the unigram matches between a candidate and a reference are given by $$P(\hat{x},x^{r}) = \frac{\mathrm{matches}(\hat{x},x^{r})}{\mathrm{length}(\hat{x})},\; R(\hat{x},x^{r}) = \frac{\mathrm{matches}(\hat{x},x^{r})}{\mathrm{length}(x^r)}\;,$$ with $\beta = \frac{1}{3}$ by default. The penalty depends on the number of chunks, i.e., groups of consecutive matches in the correct order, and penalizes disordered matches (See Section 2.2 of \citet{banerjee2005meteor}).

\subsubsection{Captioning evaluation}

Introduced specifically for image captioning, the more recent metrics CIDEr (Consensus-based Image Description Evaluation) \citep{vedantam2015cider}, SPICE (Semantic Propositional Image Caption Evaluation) \citep{anderson2016spice}, and SPIDEr \citep{liu2017improved} have demonstrated stronger correlation with human judgment. As in the previous section, we employ sentence-based oracle evaluation as defined in \eqref{eq:oracle}.

\textbf{CIDEr ($\uparrow$).} CIDEr \citep{vedantam2015cider} assigns weights to $n$-grams in the candidate and reference captions using TF–IDF. An $n$-gram receives higher weight if $(i)$ its term frequency (TF) is high, i.e., it appears often in the sequence, and $(ii)$ it is informative, i.e., it occurs infrequently across the set of reference captions in the corpus. For each sequence $x$, we construct a vector $g^{n}(x)$ of dimension equal to the number of $n$-grams of length $n$, where the $k$-th component $[g^{n}(x)]_k$ is the TF–IDF weight of the $k$-th $n$-gram in $x$. CIDEr then computes the average cosine similarity between the candidate and each reference:
\begin{equation}
\operatorname{CIDEr}_n(\hat{x}, X)\triangleq\frac{1}{|X|} \sum_{x_{r} \in X} \frac{g^{n}(\hat{x}) \cdot g^{n}(x^{r})}{\left\|g^{n}(\hat{x})\right\|\left\|g^{n}(x^{r})\right\|}\;, \quad \operatorname{CIDEr} \triangleq \frac{1}{4} \sum_{n=1}^{4} \operatorname{CIDEr}_n \;,\end{equation}
where $\cdot$ denotes the Euclidean dot product.

\textbf{SPICE ($\uparrow$).} SPICE \citep{anderson2016spice} was designed to capture semantic adequacy by focusing on objects, attributes, and relations rather than surface $n$-gram overlap. Given a set of object classes $C$, relations $R$, and attributes $A$, each caption $x$ is parsed into a scene graph $T(x)=O(x) \cup E(x) \cup K(x)$ where $O(x) \subseteq C$ is the set of objects, $E(x) \subseteq O(x) \times R \times O(x)$ encodes relations between objects, and $K(x) \subseteq O(x) \times A$ represents attributes associated with objects.
SPICE is defined as an $F_{1}$ score over the tuples in the semantic graphs:
\begin{equation}
\operatorname{SPICE}(\hat{x}, X)\triangleq F_{1}(P(\hat{x}, X),R(\hat{x}, X))\;,\end{equation}
with precision and recall given by
$$ P(\hat{x}, X) = \frac{\mathrm{matches}(T(\hat{x}),T(X))}{|T(\hat{x})|} \quad R(\hat{x}, X) = \frac{\mathrm{matches}(T(\hat{x}),T(X))}{|T(X)|}\;,$$
Here, $\mathrm{matches}(T(\hat{x}), T(X))$ counts the number of matching tuples between the candidate and the reference semantic graphs.

\textbf{SPIDEr ($\uparrow$).} SPIDEr \citep{liu2017improved} combines the strengths of CIDEr (capturing consensus through $n$-gram overlap) and SPICE (capturing semantic adequacy through scene graphs). It is defined as the simple average of the two metrics:
\begin{equation}\mathrm{SPIDEr}(\hat{x}, X) \triangleq \frac{\mathrm{CIDEr}(\hat{x}, X) + \mathrm{SPICE}(\hat{x}, X)}{2}\;.\end{equation}

\textbf{sBERT Similarity ($\uparrow$).} BERTScore \citep{zhangbertscore} leverages contextual embeddings from a pretrained BERT model \citep{devlin2018bert} to compute token-level similarity. This allows it to $(i)$ better match paraphrases and $(ii)$ capture long-range dependencies while penalizing semantic changes. sBERT Similarity \citep{zhou2022can} considers Sentence BERT \citep{reimers2019sentence} (by default paraphrase-TinyBERT-L6-v2) as the pretrained model due to its capability to compare semantics with a single embedding per sequence. Denote their contextual embeddings by $e(x), e(\hat{x}) \in \mathbb{R}^d$. For multiple references, it is defined as an average cosine similarity:
\begin{equation}
\mathrm{sBERT}(\hat{x},X) \triangleq \frac{1}{R} \sum_{x^{r} \in X} \frac{e(\hat{x})^{\top} e(x^{r})}{\left\|e(\hat{x})\right\|\left\|e(x^r)\right\|}\;.
\end{equation}

\subsubsection{Diversity evaluation}

Quality evaluation measures how well candidate captions match the references. By contrast, diversity evaluation considers only the set of generated candidates, irrespective of the references. Below, we present the diversity metrics used in our setup. Note that the oracle formulation in \eqref{eq:oracle} does not apply here.

\textbf{Div-$n$ ($\uparrow$).} Div-$n$ is defined as the ratio between the number of distinct $n$-grams in the $K$ generated captions $\hat{x}^{1}, \dots, \hat{x}^{K}$ and the total number of $n$-grams across those captions. Higher values indicate greater lexical diversity.

\textbf{mBLEU-$n$ ($\downarrow$).} Mutual BLEU (mBLEU-$n$) is computed by treating each generated caption $\hat{x}^k$ as a candidate and evaluating its BLEU score against the remaining captions $\{x^s \;|\; s \neq k\}$. The final score is the average across all $K$ captions:
\begin{equation}\mathrm{mBLEU}_{n}(\hat{x}^{1},\dots,\hat{x}^{K}) = \frac{1}{K} \sum_{k=1}^{K} \mathrm{BLEU}_{n}(\hat{x}^{k}, \{\hat{x}^s \; | \; s \neq k\})\;.\end{equation}
Lower $\mathrm{mBLEU}_{n}$ values indicate greater diversity among the generated captions.

\subsection{Training methods}
\label{secapp:training_methods}
We describe below the specificity of each of the training methods used.

\textbf{\texttt{LoRA-MLE}.} Through the article, we refer to \texttt{LoRA-MLE} as the training method that optimizes \eqref{eq:teacher_forcing}, where the LoRA adapters are trained, and the rest of the model is frozen. This corresponds exactly to the case where $K=1$ is \texttt{LoRA-MCL} (as described hereafter). We used the default initialization of Low-Rank adapters in \href{https://huggingface.co/docs/peft/index}{PEFT} library, where $A$ is initialized with Kaiming Uniform \citep{he2015delving} initialization $A \sim \mathcal{U}[-\frac{1}{\sqrt{d}},\frac{1}{\sqrt{d}}]$ where $d$ is the number of input features, and $B$ is initialized with zeros.

\textbf{\texttt{LoRA-MoE.}} We denote by \texttt{LoRA-MoE} the use of multiple adapters within LoRA modules at each layer where LoRA is applied, trained in the style of a Mixture of Experts (MoE) \citep{jacobs1991adaptive, shazeer2017outrageously}.
Let the hidden state be $h \in \mathbb{R}^{\mathcal{B} \times T \times d}$, where $\mathcal{B}$ is the batch size, $T$ the sequence length, and $d$ the feature dimension.
Under the \textit{soft} MoE formulation of \citet{muqeethsoft}, the computation at layer $\ell$ is given by
$$
h \leftarrow h W_{\ell} +\sum_k [\gamma(h)]_k h A_{\ell}^{k} B_{\ell}^{k}\;,
$$ 
where $W_{\ell}$ is the frozen base model at layer $\ell$, and the \textit{router} $\gamma: \mathbb{R}^{d} \rightarrow \Delta^{K-1}$ is applied independently to each token embedding $h_{b,t} \in \mathbb{R}^d$, producing a $K$-dimensional vector of mixing weights. In practice, we implement $\gamma$ as a linear projection followed by a softmax. In our experiments, we initialized the linear layer weights to zero so that each expert contributed equally at the start of training. In large language models, MoE typically employs hard (discrete) routing, e.g., top-$k$ selection from the router to control computational cost. In the context of LoRA, however, this tradeoff is less critical since LoRA computations are lightweight relative to the base model. Here, we adopt \textit{soft} routing, which keeps the expert block fully differentiable and avoids reliance on gradient approximation or additional load-balancing losses \citep{wumixture, li2024mixlora}.

\textbf{\texttt{LoRA-MCL.}} This is the method introduced in this paper. At each LoRA-enabled layer $\ell$, a family of $K$ LoRA adapters $(A_k^{\ell}, B_k^{\ell})_{k=1}^{K}$ is trained, while the rest of the model remains frozen. The training objective is $$
    \mathcal{L}^{\mathrm{WTA}}(\theta) = - \mathbb{E}_{c,x}\Big[  \sum_{k=1}^{K} q_k(x,c) \log p(x \mid c ; \theta_k) \Big]\,,$$
where the coefficients $q_k$ depend on the chosen WTA mode. Let $k^{\star}(x,c) = \mathrm{argmax}_{k} \; p(x \mid c ; \theta_k)$ be the index of the winning hypothesis for input $c$ and target $x$. In the \emph{vanilla WTA} mode, we set $q_k(x,c) = \mathbf{1}[k = k^{\star}(x,c)]$, which directly optimizes the Oracle NLL Loss \eqref{eqapx:metric_nll}. However, this formulation risks \emph{collapse}, where some hypotheses are rarely selected and thus under-trained. To mitigate this, \citet{rupprecht2017learning} proposed the \emph{relaxed WTA} mode:
$$q_{k}(x,c) = (1-\varepsilon)\mathbf{1}\left[k = k^{\star}(x,c)\right] + \frac{\varepsilon}{K-1}\mathbf{1}\left[k \neq k^{\star}(x,c)\right]\;.$$
which gives higher weight to the winning hypothesis while still providing a small gradient to the others, controlled by $\varepsilon > 0$. Subsequent methods extend this idea by making $q_k$ a function of the training step $t$, thereby adjusting the contribution of non-winning hypotheses during learning \citep{makansi2019overcoming, narayanan2021divide, nehme2024hierarchical, amcl}. For example, in the \emph{annealed MCL} method \citep{amcl}, a temperature parameter $\uptau$ is introduced and we have:
\begin{equation}
    q_k(x,c ; \uptau) = \frac{p(x \mid c ; \theta_k)^{\frac{1}{\uptau}}}{Z_{x,c}(\uptau)}\,, \quad \quad Z_{x,c} = \sum_{s=1}^{K} p(x \mid c ; \theta_s)^{\frac{1}{\uptau}}\;,
\end{equation} 
where the temperature $t \mapsto \uptau(t)$ follows a decreasing schedule, typically $\uptau(t) = \uptau(0)\rho^t$ with $\rho < 1$ and $\uptau(0) > 0$ where $t$ is the training step. At high temperatures, training is distributed more evenly across all hypotheses, which helps to prevent collapse. As $\uptau \to 0$, the method converges to the greedy WTA setup, which can maximize (oracle) performance provided that all hypotheses have been sufficiently trained. In the Audio Captioning experiments, Annealed MCL was trained with $\uptau(0)=1.0$, $\rho = 0.999$, and we switched back to vanilla WTA when the temperature reached $10^{-6}$.

\subsection{Decoding methods}
\label{apx:decoding}
Formally, Maximum-A-Posteriori decoding in the context of language modeling consists, given (fixed) parameters $\theta$ of finding sequences $\hat{x} = (\hat{x}_1, \dots, \hat{x}_T) \in \mathcal{V}^{T}$ that maximizes $p_{\theta}(\hat{x} \mid c) = p_{\theta}(\hat{x}_{1} \mid c) \prod_{t=2}^{T} p_{\theta}(\hat{x}_{t} \mid \hat{x}_{<t}, c)$ given a context $c \in \mathcal{C}$. Because an exhaustive search of the most likely sequence given $c$ and $\theta$ would be intractable, we used the following heuristics.

\textbf{Greedy \& Beam Search.} Given a beam size (or beam width) $B$, beam search~\citep{beamSearch} proceeds as follows. First, compute the $B$ most likely tokens $\hat{x}_{1}^1,\dots,\hat{x}_{1}^B$ from the distribution $p_{\theta}(\hat{x}_1 \mid c)$. Next, run a forward pass for each candidate $\hat{x}_{1}^i$ through the language model to obtain $B$ distributions $p_{\theta}(\cdot \mid \hat{x}_1^{k}, c)$ (for $k=1,\dots,B$), each over the vocabulary $\mathcal{V}$, yielding $B \times |\mathcal{V}|$ candidate probabilities. From these $B \times |\mathcal{V}|$ values, select the top-$B$ tokens to form the next candidates $\hat{x}_2^1,\dots,\hat{x}_2^B$, while keeping track of the preceding tokens in each beam. Repeat this procedure for $t=1,\dots,T$ to produce $B$ sequences $\hat{x}^1,\dots,\hat{x}^B \in \mathcal{V}^{T}$. Greedy search is the special case $B=1$, i.e., at each step only the top-1 token is chosen: $\hat{x}_{t+1}^{1} = \mathrm{argmax}_{x_{t+1}} p_{\theta}(x_{t+1} \mid \hat{x}_{<t})$. While this procedure is quite effective and reliable in practice, Beam Search is known to yield low diversity, returning candidates that differ only slightly near the ends of their decoding paths when asked for multiple outputs~\citep{vijayakumar2018diverse}.

\textbf{Diverse Beam Search.} Diverse Beam Search~\citep{vijayakumar2018diverse} is an alternative to Beam Search in which the beam set of size $B$ is partitioned into $G$ disjoint groups of equal size $B' = B/G$. The goal is to maximize the likelihood within each group while encouraging dissimilarity across groups. Let the set of hypotheses for group $g$ at time $t$ be $X_t^{g} \triangleq \{\, \hat{x}_{1:t}^{\,b + (g-1)B'} \mid b=1,\dots,B'\,\}$. Dissimilarity is measured through a dissimilarity function $\Delta(\cdot,\cdot)$ that measures dissimilarity between a sequence $\hat{x}_{1:t}$ and a group $X_t^{g}$. In the following, we denote by $\mathrm{Top}\text{-}B'(S, f)$ the operator returning the subset of $S$ containing the $B'$ highest-scoring elements under $f: x \in S \mapsto \mathbb{R}$. Let us also denote by $\mathrm{Expand}(X_{t-1}^{g},\mathcal{V}) = \{ [x, v] \mid v \in \mathcal{V}, x \in X_{t-1}^{g}\}$ denote the sequence obtained by appending token $v \in \mathcal{V}$ to a hypothesis in $X_{t-1}^{g}$ at decoding step $t$, with $X_0^g = \{\emptyset\}$. Then, for group $g=1$ a beam search with size $B'$ is performed, i.e., $X_t^{1} = \mathrm{Top}\text{-}{B'}\big( \mathrm{Expand}(X_{t-1}^{1}, \mathcal{V})\,,
x \in \mathcal{V}^{t} \mapsto \log p_{\theta}(x \mid c)\big),$
and for groups $g \in \{2,\dots,G\}$,
$$
X_t^{g}
= \mathrm{Top}\text{-}{B'}\big(\mathrm{Expand}(X_{t-1}^{g},\mathcal{V})\,, x \in \mathcal{V}^{t} \mapsto \log p_{\theta}(x \mid c)
+ \lambda \sum_{h=1}^{g-1} \Delta(x, X_t^{h}) \big)\;.
$$
This trades off sequence likelihood and dissimilarity to the previous groups, with diversity penalty $\lambda > 0$.
In practice, the dissimilarity decomposes as $\Delta(\hat{x}_{1:t}, X_t^{h})=\sum_{u_{1:t} \in X_t^{h}} \delta(\hat{x}_{1:t}, u_{1:t}),$
where $\delta$ is a pairwise dissimilarity.
In the \href{https://huggingface.co/docs/transformers/en/internal/generation_utils}{Transformers} implementation, $\delta$ uses a Hamming penalty at the current step, $\delta(\hat{x}_{1:t}, u_{1:t})=\mathds{1}\!\big[\hat{x}_t \neq u_t\big],$ so tokens already chosen at position $t$ by earlier groups are penalized.

\textbf{Test-time Augmentation.} TTA involves applying to the context $c$ a random perturbation $\mathcal{T}_{\phi}: \mathcal{C} \rightarrow \mathcal{C}$, parameterized by $\phi$ that controls the augmentation strength. We generate $K$ perturbed contexts $\tilde{c}_1, \dots, \tilde{c}_K \sim \mathcal{T}_{\phi}(c)$ and perform MAP generation via Beam Search with beam size $\tfrac{B}{K}$ for each perturbed context. In our audio captioning experiments, we implement TTA using SpecAugment \cite{park2019specaugment}, where $\phi$ consists of two parameters: the percentage of time and frequency bands masked in the input spectrogram.

\textbf{Remarks on the decoding setup for each training method.} 
Beam Search was used as the decoding strategy for all three training methods described in Section~\ref{secapp:training_methods}. To ensure a fair comparison under a fixed computational budget (i.e., a comparable number of forward passes), we adjust the beam size depending on the method:
\begin{itemize}
    \item \texttt{LoRA-MLE} with BS or DBS uses a beam of size $B$, where $B \geq K$ if $K$ sequences are to be returned. For comparability, \texttt{LoRA-MLE} with TTA uses a beam size of $B/K$.
    \item \texttt{LoRA-MCL} decodes each of the $K$ hypotheses with a beam of size $B/K$, and each hypothesis returns a single sequence.
    \item \texttt{LoRA-MoE} was evaluated with two decoding strategies:
    \begin{enumerate}
        \item \textit{MLE Decoding}, where \texttt{LoRA-MoE} is treated as a single-hypothesis model with \texttt{LoRA-MoE} layers integrated into the base architecture. In this case, MAP decoding with beam size $B$ yields up to $K$ sequences (as in \texttt{LoRA-MLE}).
        \item \textit{Stochastic Router Decoding}, following \citet{zuo2021taming}, where the expert index is sampled randomly from the router’s mixing-weight distribution at each layer. Each forward pass uses a beam size of $B/K$, returning one sequence.
    \end{enumerate}
\end{itemize}

Each of the above approaches is presented in the main submission results, except for Stochastic Router, which was omitted for conciseness due to its poor performance. For completeness, its results are provided in Tables \ref{tab:full_results_cl_5_hyp}, \ref{tab:full_results_ac_5_hyp}, and \ref{tab:full_results_textcaps}.

\subsection{Parallelization over the hypotheses in \texttt{LoRA-MCL}}
\label{apx:acceleration}

A naive implementation of MCL for winner selection, as in Section \ref{sec:mcl_language}, may require a loop over the $K$ hypotheses in the batch to determine the winner associated with each index of the batch, which would slow down training.

To alleviate this issue, we propose the following methodology. Let $x \in \mathbb{R}^{\mathcal{B} \times T \times d}$ denote the input of the transformer architecture, where $\mathcal{B}$ is the batch size, $T$ is the total sequence length, and $d$ is the number of features. We duplicate $x$, $K$ times along the batch size dimension to get $h = \begin{bmatrix} h^{(1)}, \dots, h^{(K)} \end{bmatrix} \in \mathbb{R}^{\mathcal{B}K \times T \times d}$. 

Computation at LoRA layer $\ell$ writes as:
\begin{equation}
\begin{bmatrix}
h^{(1)} & \cdots & h^{(K)}
\end{bmatrix}
\leftarrow
\begin{bmatrix}
h^{(1)} & \cdots & h^{(K)}
\end{bmatrix}
\begin{bmatrix}
A^{1}_{\ell} B^{1}_{\ell} & 0 & 0 & 0 \\
0 & A^{2}_{\ell} B^{2}_{\ell} & 0 & 0 \\
\vdots & \vdots & \ddots & \vdots \\
0 & 0 & 0 & A^{K}_{\ell} B^{K}_{\ell}
\end{bmatrix}
+ 
\begin{bmatrix}
h^{(1)} W_{\ell} & \cdots & h^{(K)} W_{\ell}
\end{bmatrix}
\,,
\label{eq:group_lora}
\end{equation}
where $W_{\ell} \in \mathbb{R}^{d \times d}$ is the base model (whose parameters remain frozen during training). In practice, this computation is conducted using a Group Convolution operation.\footnote{
This can be done by first reshaping $h$ to shape $(\mathcal{B} \times Kd \times T)$ and applying a \texttt{nn.Conv1d($Kd$, $Kd$, kernel\_size=1, groups=$K$)} following the \href{https://pytorch.org/docs/stable/generated/torch.nn.Conv1d.html}{PyTorch} layer implementation, reshaping back $h$ to shape $(K\mathcal{B} \times T \times d)$ before adding the base model output, and then repeating at the next LoRA unit.
} While this duplication virtually multiplies the batch size by $K$, the memory overhead remains manageable, assuming that $r \ll d$.

\textbf{Computation cost comparison.} We use the synthetic data experimental setup of Sec.~\ref{sec:synt_data}, comparing the training time and peak GPU memory of \texttt{LoRA-MLE} against \texttt{LoRA-MCL} (with both standard implementation and group convolution variant). Table \ref{runtime:toy} shows that \texttt{LoRA-MLE} is faster and less costly than \texttt{LoRA-MCL}, where the compute cost increases sublinearly with $K$. In this setup, we also see that the \texttt{LoRA-MCL} with group conv is significantly faster than \texttt{LoRA-MCL} standard (for a fixed $K$), with the gap increasing with $K$, although the group variant induces a slight memory overhead. We ran the same experiment on Qwen-2-Audio with AudioCaps (Table \ref{runtime:audiocaps}), where we observed similar trends. The experiments on Clotho, where sequence lengths are longer (30 seconds of audio, against 10 seconds for Audiocaps), show that group LoRA can also induce time overhead, due to non-contiguous memory layouts introduced by reshaping. Investigating how to handle this, e.g., with custom CUDA kernels, could be an interesting path for further work.

\begin{table}[h]
\centering
\caption{\textbf{Training cost comparing \texttt{LoRA-MLE} and \texttt{LoRA-MCL} on toy experiment setup on a single Tesla P100-PCIE-16GB.} Training time (seconds, $\downarrow$) and peak GPU memory (MiB, $\downarrow$). Results averaged over 10 seeds.}
\vspace{2pt}
\begin{tabular}{l c c c}
\midrule
Training & $K$ & Training time (s) $\downarrow$ & Peak GPU memory (MiB) $\downarrow$ \\
\midrule
\texttt{LoRA-MLE} & 1 & $43.29 \pm 0.94$ & 87.8 \\
\midrule
\multicolumn{4}{l}{\textbf{$K = 3$}} \\
\texttt{LoRA-MCL} standard & 3 & $54.95 \pm 0.51$ & 205.7 \\
\texttt{LoRA-MCL} group conv & 3 & $46.16 \pm 1.39$ & 237.5 \\
\midrule
\multicolumn{4}{l}{\textbf{$K = 5$}} \\
\texttt{LoRA-MCL} standard & 5 & $64.53 \pm 0.51$ & 319.6 \\
\texttt{LoRA-MCL} group conv & 5 & $49.92 \pm 0.47$ & 383.2 \\
\midrule
\end{tabular}
\label{runtime:toy}
\end{table}

\begin{table}[h]
\centering
\caption{\textbf{Training cost comparing \texttt{LoRA-MLE} and \texttt{LoRA-MCL} on the AudioCaps setup with Qwen-2-Audio a single NVIDIA H100 80GB.} Training time (seconds) measured over 100 iterations. Results averaged over 5 seeds.}
\vspace{2pt}
\begin{tabular}{l c c c}
\midrule
Training & $K$ & Training time (s) for 100 iterations & Peak GPU memory (MiB) \\
\midrule
\texttt{LoRA-MLE} & 1 & $19.52 \pm 2.22$ & 35364.3 \\
\midrule
\multicolumn{4}{l}{\textbf{$K = 2$}} \\
\texttt{LoRA-MCL} standard & 2 & $35.57 \pm 3.34$ & 38168.2 \\
\texttt{LoRA-MCL} group conv & 2 & $33.81 \pm 3.84$ & 38311.5 \\
\midrule
\multicolumn{4}{l}{\textbf{$K = 3$}} \\
\texttt{LoRA-MCL} standard & 3 & $49.51 \pm 2.86$ & 41039.4 \\
\texttt{LoRA-MCL} group conv & 3 & $36.28 \pm 0.10$ & 41394.1 \\
\midrule
\end{tabular}
\label{runtime:audiocaps}
\end{table}

\begin{table}[h]
\centering
\caption{\textbf{Training cost comparing \texttt{LoRA-MLE} and \texttt{LoRA-MCL} on the Clotho setup with Qwen-2-Audio a single NVIDIA H100 80GB.} Training time (seconds) measured over 100 iterations. Results averaged over 5 seeds.}
\vspace{2pt}
\begin{tabular}{l c c c}
\midrule
Training & $K$ & Training time (s) & Peak GPU memory (MiB) \\
\midrule
\texttt{LoRA-MLE} & 1 & $38.83 \pm 8.49$ & 40737.3 \\
\midrule
\multicolumn{4}{l}{\textbf{$K = 2$}} \\
\texttt{LoRA-MCL} standard   & 2 & $48.59 \pm 5.43$ & 48427.1 \\
\texttt{LoRA-MCL} group conv & 2 & $54.42 \pm 3.15$ & 48853.0 \\
\midrule
\multicolumn{4}{l}{\textbf{$K = 3$}} \\
\texttt{LoRA-MCL} standard   & 3 & $74.56 \pm 17.69$ & 56180.6 \\
\texttt{LoRA-MCL} group conv & 3 & $95.31 \pm 7.57$  & 57060.6 \\
\midrule
\end{tabular}
\end{table}

\subsection{Audio captioning experiments}
\label{apx:exp_details_audio}

\subsubsection{Experimental setup}

\textbf{Architecture and training.} We used the Instructed version of Qwen-2-Audio \citep{chu2024qwen2} as the base model, which features $\sim8.4$ billion parameters. We trained using \texttt{bfloat16} precision. We used LoRA adapters applied to the $Q, K, V$ linear projections of the attention modules, and the upside and downside projections of the feedforward blocks, for all the transformer blocks, which include both the audio encoder and language model decoder. We used a rank $r$, with $r = \alpha = 8$ unless otherwise stated, with dropout equal to $0.1$ (enabled during training, and disabled during inference). We trained with a batch size of $2$, with AdamW optimizer \citep{loshchilovdecoupled} (with $\beta_{1}=0.9$, and $\beta_{2}=0.98$), weight decay of $0.05$, using a cosine scheduler with minimum learning rate of $10^{-6}$ and maximum learning rate of $10^{-5}$, with a warmup ratio of $0.1$. Gradient clipping is used with a maximum gradient norm of $1.0$. The validation loss was computed once at the end of the epoch.

\subsubsection{Ablation on the relaxation parameters in \texttt{LoRA-MCL}}

\textbf{Effect of $\varepsilon$.} Let $\ell_k=-\log p\left(x \mid c, \theta_k\right)$ denote the NLL of hypothesis $k$ for a pair $(c, x)$. Recall that the relaxed MCL loss for such a pair is
$$
(1-\varepsilon) \ell_k^{\star}+\frac{\varepsilon}{K-1} \sum_{k=1, k \neq k^{\star}}^K \ell_k=\left(1-\frac{K \varepsilon}{K-1}\right) \ell_k^{\star}+\frac{\varepsilon}{K-1} \sum_{k=1}^K \ell_k\;,
$$
where $k^{\star}(x,c) = \mathrm{argmax}_k \; p(x \mid c; \theta_k)$ is the winner hypothesis. We see that the first term is a force that pushes the hypothesis $k$ toward the winner hypothesis, while the second term assigns equal weight to each hypothesis, pulling them toward the barycenter of the conditional distribution (See Figure 1 in \cite{amcl}). The first term vanishes when $\varepsilon=\frac{K-1}{K}$, which provides an upper bound on the value of $\varepsilon$. In practice, when choosing $\varepsilon$ for a new task, the smaller the better unless a collapse is observed; we recommend trying $\varepsilon \in (0,0.1]$. We run \texttt{LoRA-MCL} (relaxed) with $\varepsilon \in \{0.0005,0.05,0.1,0.3,0.5,0.8\}$ on AudioCaps (AC) and Clotho (CL) in Table \ref{tab:effect_relaxation} with $K=5$ (\texttt{LoRA-MCL} ``relaxed'' rows). We observe that outside the ``small'' $\varepsilon$ regime ($0.0005,0.05$), increasing $\varepsilon$ significantly degrades the diversity (e.g., mBLEU 4 from $0.410$ with $\varepsilon=0.0005$ on AudioCaps to $0.963$ with $\varepsilon=0.8$ ). For quality (as measured by Oracle SPIDEr), performance also tends to degrade outside the small $\varepsilon$ regime. This occurs because large $\varepsilon$ diminishes the benefits of MCL training, causing the model to behave increasingly like \texttt{LoRA-MLE}, where the heads eventually become uniformised.

\textbf{Effect of the temperature scheduler.} \texttt{LoRA-MCL} annealed is based on equations \eqref{eq:relaxation_wta_loss} and \eqref{eq:annealed_coeff}. The annealed method relies on the theoretical foundation of deterministic annealing (e.g., \citet{rose2002vector, amcl}), which predicts \textit{phase transition} phenomena in which the hypotheses ``split'' at specific temperature levels, namely critical temperatures, to explore hierarchically different modes of the conditional distribution. Given a fixed number of training steps, the temperature schedule needs to be neither too slow, otherwise the final training step may still occur above the critical temperature, nor too fast, in which case the model does not fully benefit from annealing and collapse may still occur. We provide an ablation on the decay rate in Table \ref{tab:effect_relaxation} (see the annealed rows) for $\rho \in\{0.9,0.995,0.999,0.9999\}$. Performance remains good across most of this range, except for $\rho = 0.9999$ where we observe a degradation of diversity and performance. We attribute this to the final temperature remaining above the critical temperature.

\begin{table}[ht]
    \centering
    \caption{\textbf{Ablation on the relaxation parameters in LoRA-MCL.} $\varepsilon$ is the relaxation parameter, and $\rho$ in the temperature scheduler $\uptau(t) = \rho^{t} \uptau_{0}$, with $\uptau_{0} = 1.0$. Evaluation on AudioCaps (AC) and Clotho (CL).}
    \begin{center}
    \resizebox{\columnwidth}{!}{
\begin{tabular}{lcccccc}
\toprule
Training & Decoding & Beam & $\mathrm{mBLEU_{4}}$ (AC) & $\mathrm{mBLEU_{4}}$ (CL) & $\mathrm{SPIDEr}$ (AC) & $\mathrm{SPIDEr}$ (CL) \\
\midrule
\texttt{LoRA-MLE} ($r = 8$) & DBS ($\lambda = 0.8$) & 5 & 0.448 & \underline{0.446} & 0.662 & 0.423 \\
\midrule \midrule
\texttt{LoRA-MCL} ($\varepsilon = 0.0005$) & BS & 5 & \textbf{0.410} & 0.488 & 0.706 & \underline{0.436} \\
\texttt{LoRA-MCL} ($\varepsilon = 0.05$) & BS & 5 & 0.491 & 0.478 & 0.728 & 0.434 \\
\texttt{LoRA-MCL} ($\varepsilon = 0.1$) & BS & 5 & 0.502 & 0.526 & \textbf{0.738} & 0.435 \\
\texttt{LoRA-MCL} ($\varepsilon = 0.3$) & BS & 5 & 0.624 & 0.643 & 0.700 & 0.421 \\
\texttt{LoRA-MCL} ($\varepsilon = 0.5$) & BS & 5 & 0.766 & 0.790 & 0.634 & 0.390 \\
\texttt{LoRA-MCL} ($\varepsilon = 0.8$) & BS & 5 & 0.963 & 0.952 & 0.508 & 0.329 \\
\midrule
\texttt{LoRA-MCL} ($\rho = 0.9$) & BS & 5 & 0.458 & \textbf{0.432} & 0.701 & 0.429 \\
\texttt{LoRA-MCL} ($\rho = 0.995$) & BS & 5 & \underline{0.422} & 0.490 & \underline{0.729} & \underline{0.438} \\
\texttt{LoRA-MCL} ($\rho = 0.999$) & BS & 5 & 0.423 & 0.478 & 0.716 & \textbf{0.443} \\
\texttt{LoRA-MCL} ($\rho = 0.9999$) & BS & 5 & 0.830 & 0.947 & 0.610 & 0.331 \\
\bottomrule
\end{tabular}}
\end{center}
\label{tab:effect_relaxation}
\end{table}

\subsubsection{Comparison of \texttt{LoRA-MCL} with multi-head fine-tuning}
\label{apx:multi-head}
In this section, we compare \texttt{LoRA-MCL} with multi-head MCL. Using Qwen-2-Audio for audio captioning, we duplicated the Language model head (``LMHead'', ~640M params) $K$ times, froze the rest of the model, and applied MCL training (without LoRA) with both relaxed and annealed variants. As an additional variant, showing that parameter count alone is not the only issue, we also duplicated an internal module (the multimodal projector ``MMProj'', $\sim$ 5.2M parameters) and trained it with the same scheme. As discussed in Section \ref{sec:method_lora_mcl}, multi-head initialization is not trivial. Copying the pretrained head yields no initialization diversity, while random reinitialization discards pretrained knowledge.

Results are displayed in Table \ref{tab:multi-head}. We find that $(i)$ Random initialization (Init ``Random'') degrades quality for both MMProj and LMHead, and in the LMHead case, produces largely unintelligible captions with artificially high diversity. $(ii)$ Copying the parameters of the pretrained model (Init ``Copy'') produces coherent outputs, but does not meaningfully benefit from MCL training due to amplified collapse risk, yielding significantly lower quality in practice. These findings show that multi-head approaches are not competitive with \texttt{LoRA-MCL} in a comparable training setup in the setting of Large language models.

\begin{table}
    \centering
    \begin{minipage}{.8\linewidth}
    \caption{\textbf{Comparison of \texttt{LoRA-MCL} with Multi-head fine-tuning}. Comparison is done with $K = 5$ hypotheses in AudioCaps, with the same experimental setup as in the paper. The ``Init'' column refer to the initialization technique for the trainable parameters. In the Multi-head versions, the trainable weights are either the Language model head (LMHead) or the Multimodal Projector (MMProj). The latter are trained with the relaxed variant with $\varepsilon = 0.05$.}
    \vspace{4pt}
    \centering
\begin{tabular}{lcccccc}
\toprule
Training & Init & $K$ & Decoding & Beam & $\mathrm{mBLEU_{4}}$ & $\mathrm{SPIDEr}$ \\
\midrule
Multi-head (LMHead) & Copy & 5 & BS & 5 & 0.489 & 0.561 \\
Multi-head (LMHead) & Random & 5 & BS & 5 & \underline{0.001} & 0.002 \\
Multi-head (MMProj) & Copy & 5 & BS & 5 & 0.530 & 0.394 \\
Multi-head (MMProj) & Random & 5 & BS & 5 & 0.256 & 0.140 \\
\midrule \midrule
\texttt{LoRA-MCL} ($\varepsilon = 0.05$) & LoRA & 5 & BS & 5 & 0.491 & \textbf{0.728} \\
\texttt{LoRA-MCL} (annealed) & LoRA & 5 & BS & 5 & 0.423 & \underline{0.716} \\
\bottomrule
\end{tabular}
\label{tab:multi-head}
    \end{minipage}
\end{table}

\subsubsection{Additional results}
\label{sec:additional_results}
\textbf{Evaluating with sampling-based decoding.} We also report results using sampling-based decoding methods in Tables \ref{tabapx:sampling_clotho} and \ref{tabapx:sampling_audiocaps} for Clotho and AudioCaps. Specifically, we use Top-$k$ sampling with $k = 50$, Top-$p$ sampling with $p = 0.95$, and Typical sampling with a threshold of $0.95$. For all sampling methods, we apply a repetition penalty of $1.1$, following \citep{keskar2019ctrl}. In these experiments, the temperature was set to $\eta = 1.0$. 

We observe that both \texttt{LoRA-MLE} and \texttt{LoRA-MCL} yield significantly higher diversity than MAP decoding, albeit at the cost of reduced output quality. This trade-off is evident in Tables~\ref{tab:full_results_cl_5_hyp} and \ref{tab:full_results_ac_5_hyp}. On AudioCaps, \texttt{LoRA-MCL} shows a slight improvement in both quality and diversity compared to \texttt{LoRA-MLE}. On Clotho, it provides a small gain in diversity, while quality slightly favors \texttt{LoRA-MLE}. These findings highlight the need for further evaluation with different annealing schedules to better characterize the quality–diversity trade-off in sampling-based decoding. Moreover, a deeper study of how the number of generated hypotheses influences sampling quality and its implications for test-time inference scaling with \texttt{LoRA-MCL}~\citep{zhao2025sample} is left for future work.

\textbf{Further comparison with prior work.} We compare against \citet{zhu2025diffusion} a diffusion-based method using retrieval-guided Langevin dynamics (DAC-RLD), which we identify as the strongest published method for diverse audio captioning on Clotho and AudioCaps with publicly available code and checkpoints. A VAE-based method from \citet{zhang2024generating} is also open-sourced (Clotho only) but underperforms DAC-RLD in both quality and diversity according to the reported scores. Adversarial training has also been explored \cite{mei2022diverse, xu2022diversity}, though these methods are outperformed by DAC-RLD, and do not release code. We evaluate the pretrained DAC-RLD checkpoints using our setup: 5 candidate captions generated per audio (instead of 50 followed by Minimum Bayes Risk decoding \cite{tromble2008lattice} in the original work), oracle sentence-level quality metrics, and mBLEU-4 for diversity. For DAC-RLD, we run both with Beam Search and Nucleus sampling, as in the original work. Results (Table \ref{tab:diffusion_comp}) show that DAC-RLD achieves diversity comparable to \texttt{LoRA-MCL} with nucleus sampling (slightly better or worse depending on the dataset) but at the cost of substantially lower SPIDEr scores than \texttt{LoRA-MCL} on both Clotho and AudioCaps. 

\begin{table}[!htbp]
    \centering
    \begin{minipage}{.9\columnwidth}
    \caption{\textbf{Comparison against DAC-RLD \cite{zhu2025diffusion} in Diverse Audio Captioning.} Evaluation on AudioCaps (AC) and Clotho (CL).}
    \vspace{4pt}
    \centering
    \resizebox{\linewidth}{!}{
\begin{tabular}{lcccccc}
\toprule
Training & Decoding & Beam & $\mathrm{mBLEU}_{4}$ (AC) & $\mathrm{mBLEU}_{4}$ (CL) & SPIDEr (AC) & SPIDEr (CL) \\
\midrule 
DAC-RLD & Nucleus 
($p = 0.95)$ & 1 & \textbf{0.157} & \underline{0.150}  & 0.435 & 0.244 \\
DAC-RLD & BS & 5 & 0.239 & 0.215 & 0.505 & 0.287 \\
\midrule
\texttt{LoRA-MCL} (annealed) & Nucleus ($p = 0.95)$ & 1 & \underline{0.163} & \textbf{0.098} & 0.569 & 0.325 \\
\texttt{LoRA-MCL} ($\varepsilon = 0.05$) & BS & 5 & 0.491 & 0.478 & \textbf{0.728} & \underline{0.434} \\
\texttt{LoRA-MCL} ($\rho = 0.999$) & BS & 5 & 0.423 & 0.478 & \underline{0.716} & \textbf{0.443} \\
\bottomrule
\end{tabular}}
\label{tab:diffusion_comp}
    \end{minipage}
\end{table}

\begin{table}
    \caption{\textbf{Results for Clotho with 5 hypotheses and MAP Decoding.} `BS', `DBS', `SR' and `TTA' stand for beam search, diverse beam search, stochastic router, and test-time augmentation respectively. We refer to the TTA parameters as $\phi_{i}$, where the strength that increases with $i$; $\phi_{1}=(0.2,0.3)$, $\phi_{2}=(0.4,0.6)$, and $\phi_{3}=(0.6,0.9)$ as the time and frequency proportion mask with SpecAugment (See Apx. \ref{apx:decoding}).
    }
    \begin{center}
    \resizebox{\columnwidth}{!}{
        \begin{tabular}{lcccccccccccc}
\toprule
Training & Decoding & Beam & $\mathrm{Div}_{2}$ & $\mathrm{mBLEU}_{4}$ & $\mathrm{BLEU}_{1}$ & $\mathrm{BLEU}_{4}$ & $\mathrm{METEOR}$ & $\mathrm{ROUGE}_{L}$ & $\mathrm{sBERT}$ & $\mathrm{CIDEr}_{D}$ & $\mathrm{SPICE}$ & $\mathrm{SPIDEr}$ \\
\midrule
\texttt{LoRA-MLE} ($r = 8$) & BS & 5 & 0.365 & 0.822 & 0.656 & 0.137 & 0.228 & 0.445 & 0.575 & 0.626 & 0.174 & 0.394 \\
\texttt{LoRA-MLE} ($r = 40$) & BS & 5 & 0.367 & 0.818 & 0.643 & 0.115 & 0.226 & 0.435 & 0.570 & 0.595 & 0.172 & 0.376 \\
\texttt{LoRA-MLE} ($r = 8$) & BS & 10 & 0.391 & 0.783 & 0.661 & 0.142 & 0.236 & 0.453 & 0.578 & 0.646 & 0.181 & 0.406 \\
\texttt{LoRA-MLE} ($r = 40$) & BS & 10 & 0.397 & 0.777 & 0.647 & 0.127 & 0.232 & 0.445 & 0.573 & 0.615 & 0.177 & 0.388 \\
\texttt{LoRA-MLE} ($r = 8$) & BS & 25 & 0.405 & 0.759 & 0.658 & 0.144 & 0.236 & 0.455 & 0.579 & 0.648 & 0.182 & 0.407 \\
\texttt{LoRA-MLE} ($r = 40$) & BS & 25 & 0.415 & 0.746 & 0.648 & 0.131 & 0.234 & 0.447 & 0.578 & 0.625 & 0.181 & 0.395 \\
\texttt{LoRA-MLE} ($r = 8$) & DBS ($\lambda=0.8$) & 5 & 0.605 & 0.446 & 0.686 & 0.147 & 0.241 & 0.471 & \underline{0.601} & 0.678 & 0.193 & 0.423 \\
\texttt{LoRA-MLE} ($r = 40$) & DBS ($\lambda=0.8$) & 5 & 0.613 & 0.440 & 0.677 & 0.140 & 0.239 & 0.463 & 0.599 & 0.659 & 0.194 & 0.414 \\
\texttt{LoRA-MLE} ($r = 8$) & DBS ($\lambda=1.0$) & 5 & 0.625 & 0.417 & 0.685 & 0.148 & 0.242 & 0.470 & \textbf{0.602} & 0.681 & 0.194 & 0.425 \\
\texttt{LoRA-MLE} ($r = 40$) & DBS ($\lambda=1.0$) & 5 & \underline{0.634} & \underline{0.407} & 0.678 & 0.142 & 0.239 & 0.463 & 0.600 & 0.660 & 0.193 & 0.414 \\
\texttt{LoRA-MLE} ($r = 8$) & DBS ($\lambda=0.8$) & 10 & 0.487 & 0.712 & 0.670 & 0.143 & 0.238 & 0.460 & 0.594 & 0.656 & 0.191 & 0.414 \\
\texttt{LoRA-MLE} ($r = 40$) & DBS ($\lambda=0.8$) & 10 & 0.499 & 0.694 & 0.661 & 0.132 & 0.235 & 0.448 & 0.590 & 0.634 & 0.187 & 0.401 \\
\texttt{LoRA-MLE} ($r = 8$) & DBS ($\lambda=1.0$) & 10 & 0.491 & 0.708 & 0.671 & 0.143 & 0.238 & 0.456 & 0.595 & 0.662 & 0.192 & 0.417 \\
\texttt{LoRA-MLE} ($r = 40$) & DBS ($\lambda=1.0$) & 10 & 0.501 & 0.696 & 0.659 & 0.131 & 0.234 & 0.444 & 0.591 & 0.629 & 0.188 & 0.397 \\
\texttt{LoRA-MLE} ($r = 8$) & DBS ($\lambda=0.8$) & 25 & 0.368 & 0.818 & 0.653 & 0.135 & 0.228 & 0.444 & 0.575 & 0.626 & 0.176 & 0.394 \\
\texttt{LoRA-MLE} ($r = 40$) & DBS ($\lambda=0.8$) & 25 & 0.374 & 0.811 & 0.644 & 0.115 & 0.226 & 0.435 & 0.571 & 0.594 & 0.174 & 0.377 \\
\texttt{LoRA-MLE} ($r = 8$) & DBS ($\lambda=1.0$) & 25 & 0.368 & 0.818 & 0.652 & 0.134 & 0.228 & 0.443 & 0.574 & 0.622 & 0.174 & 0.392 \\
\texttt{LoRA-MLE} ($r = 40$) & DBS ($\lambda=1.0$) & 25 & 0.372 & 0.812 & 0.643 & 0.115 & 0.226 & 0.435 & 0.571 & 0.592 & 0.174 & 0.376 \\
\texttt{LoRA-MLE} ($r = 8$) & TTA BS ($\phi_1$) & 1 & 0.443 & 0.699 & 0.652 & 0.114 & 0.225 & 0.440 & 0.581 & 0.608 & 0.174 & 0.383 \\
\texttt{LoRA-MLE} ($r = 8$) & TTA BS ($\phi_2$) & 1 & 0.540 & 0.558 & 0.663 & 0.130 & 0.230 & 0.453 & 0.588 & 0.634 & 0.179 & 0.397 \\
\texttt{LoRA-MLE} ($r = 8$) & TTA BS ($\phi_3$) & 1 & \textbf{0.642} & \textbf{0.404} & 0.653 & 0.118 & 0.225 & 0.441 & 0.581 & 0.596 & 0.174 & 0.376 \\
\texttt{LoRA-MLE} ($r = 8$) & TTA BS ($\phi_1$) & 5 & 0.412 & 0.745 & 0.655 & 0.137 & 0.232 & 0.452 & 0.580 & 0.637 & 0.181 & 0.402 \\
\texttt{LoRA-MLE} ($r = 8$) & TTA BS ($\phi_2$) & 5 & 0.516 & 0.593 & 0.681 & 0.156 & 0.243 & 0.470 & 0.591 & 0.685 & 0.194 & 0.430 \\
\texttt{LoRA-MLE} ($r = 8$) & TTA BS ($\phi_3$) & 5 & 0.619 & 0.445 & 0.675 & 0.148 & 0.236 & 0.463 & 0.586 & 0.648 & 0.186 & 0.407 \\
\midrule
\texttt{LoRA-MoE} & BS & 5 & 0.363 & 0.823 & 0.650 & 0.131 & 0.228 & 0.443 & 0.571 & 0.614 & 0.173 & 0.387 \\
\texttt{LoRA-MoE} & BS & 10 & 0.395 & 0.783 & 0.659 & 0.137 & 0.236 & 0.452 & 0.579 & 0.643 & 0.181 & 0.405 \\
\texttt{LoRA-MoE} & BS & 25 & 0.410 & 0.755 & 0.657 & 0.144 & 0.239 & 0.457 & 0.579 & 0.653 & 0.184 & 0.411 \\
\texttt{LoRA-MoE} & DBS ($\lambda=0.8$) & 5 & 0.611 & 0.441 & 0.682 & 0.143 & 0.241 & 0.468 & 0.599 & 0.668 & 0.194 & 0.418 \\
\texttt{LoRA-MoE} & DBS ($\lambda=1.0$) & 5 & 0.630 & 0.410 & 0.682 & 0.147 & 0.241 & 0.467 & 0.600 & 0.675 & 0.195 & 0.422 \\
\texttt{LoRA-MoE} & DBS ($\lambda=0.8$) & 10 & 0.490 & 0.706 & 0.670 & 0.142 & 0.238 & 0.459 & 0.593 & 0.662 & 0.192 & 0.416 \\
\texttt{LoRA-MoE} & DBS ($\lambda=1.0$) & 10 & 0.494 & 0.705 & 0.668 & 0.138 & 0.237 & 0.456 & 0.592 & 0.650 & 0.191 & 0.410 \\
\texttt{LoRA-MoE} & DBS ($\lambda=0.8$) & 25 & 0.370 & 0.816 & 0.653 & 0.133 & 0.230 & 0.447 & 0.573 & 0.623 & 0.176 & 0.392 \\
\texttt{LoRA-MoE} & DBS ($\lambda=1.0$) & 25 & 0.369 & 0.816 & 0.652 & 0.133 & 0.230 & 0.446 & 0.573 & 0.622 & 0.176 & 0.391 \\
\texttt{LoRA-MoE} & SR BS & 1 & 0.308 & 0.878 & 0.604 & 0.090 & 0.203 & 0.406 & 0.554 & 0.515 & 0.153 & 0.330 \\
\texttt{LoRA-MoE} & SR BS & 2 & 0.317 & 0.868 & 0.629 & 0.116 & 0.217 & 0.427 & 0.564 & 0.576 & 0.163 & 0.364 \\
\texttt{LoRA-MoE} & SR BS & 5 & 0.309 & 0.880 & 0.620 & 0.109 & 0.217 & 0.424 & 0.562 & 0.561 & 0.162 & 0.357 \\
\midrule\midrule
\texttt{LoRA-MCL} ($\varepsilon=0.0005$) & BS & 1 & 0.605 & 0.440 & 0.675 & 0.135 & 0.233 & 0.458 & 0.597 & 0.654 & 0.187 & 0.408 \\
\texttt{LoRA-MCL} ($\varepsilon=0.05$) & BS & 1 & 0.617 & 0.415 & 0.680 & 0.130 & 0.239 & 0.463 & 0.601 & 0.662 & 0.190 & 0.414 \\
\texttt{LoRA-MCL} (annealed) & BS & 1 & 0.621 & 0.415 & 0.673 & 0.131 & 0.237 & 0.458 & 0.599 & 0.665 & 0.189 & 0.415 \\
\texttt{LoRA-MCL} ($\varepsilon=0.0005$) & BS & 2 & 0.591 & 0.461 & 0.688 & 0.154 & 0.242 & 0.472 & 0.598 & 0.688 & 0.192 & 0.428 \\
\texttt{LoRA-MCL} ($\varepsilon=0.05$) & BS & 2 & 0.593 & 0.452 & \textbf{0.692} & 0.160 & \underline{0.247} & \textbf{0.481} & 0.599 & 0.694 & 0.193 & 0.431 \\
\texttt{LoRA-MCL} (annealed) & BS & 2 & 0.595 & 0.456 & 0.687 & 0.158 & 0.245 & 0.472 & 0.599 & 0.698 & 0.196 & 0.435 \\
\texttt{LoRA-MCL} ($\varepsilon=0.0005$) & BS & 5 & 0.581 & 0.488 & 0.687 & 0.156 & 0.244 & 0.476 & 0.599 & \underline{0.700} & \underline{0.196} & \underline{0.436} \\
\texttt{LoRA-MCL} ($\varepsilon=0.05$) & BS & 5 & 0.581 & 0.478 & 0.689 & \underline{0.162} & \textbf{0.249} & \underline{0.480} & 0.599 & 0.697 & 0.196 & 0.434 \\
\texttt{LoRA-MCL} (annealed) & BS & 5 & 0.584 & 0.478 & \underline{0.689} & \textbf{0.168} & 0.246 & 0.477 & 0.600 & \textbf{0.711} & \textbf{0.199} & \textbf{0.443} \\
\bottomrule
\end{tabular}
    }
    \end{center}
    \label{tab:full_results_cl_5_hyp}
    \end{table}

\begin{table}
    \caption{\textbf{Results for Clotho with 5 hypotheses and Sampling-based Decoding.} }
    \begin{center}
    \resizebox{\columnwidth}{!}{
        \begin{tabular}{lcccccccccccc}
\toprule
Training & Decoding & $\mathrm{Div}_{2}$ & $\mathrm{mBLEU}_{4}$ & $\mathrm{BLEU}_{4}$ & $\mathrm{METEOR}$ & $\mathrm{ROUGE}_{L}$ & $\mathrm{sBERT}$ & $\mathrm{CIDEr}_{D}$ & $\mathrm{SPICE}$ & $\mathrm{SPIDEr}$ \\
\midrule
\texttt{LoRA-MLE} ($r = 8$) & Top-$k$ sampling & 0.813 & 0.109 & 0.066 & \textbf{0.215} & 0.402 & \underline{0.593} & 0.497 & \textbf{0.176} & 0.323 \\
\texttt{LoRA-MLE} ($r = 8$) & Nucleus (Top-$p$) sampling & 0.812 & 0.112 & 0.073 & 0.212 & \underline{0.403} & 0.586 & \underline{0.516} & 0.173 & \underline{0.330} \\
\texttt{LoRA-MLE} ($r = 8$) & Typical $p$ sampling & 0.812 & 0.111 & 0.073 & \underline{0.212} & \textbf{0.403} & 0.586 & \textbf{0.516} & 0.173 & \textbf{0.331} \\
\midrule\midrule
\texttt{LoRA-MCL} (annealed) & Top-$k$ sampling & \underline{0.819} & \textbf{0.097} & \underline{0.074} & 0.212 & 0.403 & \textbf{0.597} & 0.507 & \underline{0.175} & 0.327 \\
\texttt{LoRA-MCL} (annealed) & Nucleus (Top-$p$) sampling & \textbf{0.820} & \underline{0.098} & \textbf{0.074} & 0.212 & 0.403 & 0.589 & 0.507 & 0.171 & 0.325 \\
\texttt{LoRA-MCL} (annealed) & Typical $p$ sampling & 0.819 & 0.103 & 0.072 & 0.211 & 0.402 & 0.590 & 0.509 & 0.172 & 0.327 \\

\bottomrule
\end{tabular}
    }
    \end{center}
     \label{tabapx:sampling_clotho}
    \end{table}

\begin{table}
    \caption{\textbf{Results for AudioCaps with 5 hypotheses and MAP Decoding.}}
    \begin{center}
    \resizebox{\columnwidth}{!}{
        \begin{tabular}{llllllllllll}
\toprule
Training & Decoding & Beam & $\mathrm{Div}_{2}$ & $\mathrm{mBLEU}_{4}$ & $\mathrm{BLEU}_{4}$ & $\mathrm{METEOR}$ & $\mathrm{ROUGE}_{L}$ & $\mathrm{sBERT}$ & $\mathrm{CIDEr}_{D}$ & $\mathrm{SPICE}$ & $\mathrm{SPIDEr}$ \\
\midrule
\texttt{LoRA-MLE} ($r = 8$) & BS & 5 & 0.395 & 0.764 & 0.267 & 0.377 & 0.606 & 0.700 & 1.121 & 0.250 & 0.668 \\
\texttt{LoRA-MLE} ($r = 40$) & BS & 5 & 0.392 & 0.773 & 0.280 & 0.382 & 0.606 & 0.701 & 1.144 & 0.251 & 0.681 \\
\texttt{LoRA-MLE} ($r = 8$) & BS & 10 & 0.410 & 0.746 & 0.286 & 0.385 & 0.610 & 0.704 & 1.157 & 0.256 & 0.689 \\
\texttt{LoRA-MLE} ($r = 40$) & BS & 10 & 0.407 & 0.747 & 0.298 & 0.382 & 0.610 & 0.704 & 1.151 & 0.255 & 0.686 \\
\texttt{LoRA-MLE} ($r = 8$) & BS & 25 & 0.417 & 0.732 & 0.281 & 0.383 & 0.609 & 0.706 & 1.144 & 0.258 & 0.683 \\
\texttt{LoRA-MLE} ($r = 40$) & BS & 25 & 0.415 & 0.735 & 0.289 & 0.384 & 0.611 & 0.706 & 1.152 & 0.260 & 0.688 \\
\texttt{LoRA-MLE} ($r = 8$) & DBS ($\lambda=0.8$) & 5 & 0.553 & 0.448 & 0.268 & 0.378 & 0.610 & 0.708 & 1.117 & 0.248 & 0.662 \\
\texttt{LoRA-MLE} ($r = 40$) & DBS ($\lambda=0.8$) & 5 & 0.557 & 0.444 & 0.263 & 0.380 & 0.612 & 0.707 & 1.130 & 0.248 & 0.669 \\
\texttt{LoRA-MLE} ($r = 8$) & DBS ($\lambda=1.0$) & 5 & 0.580 & 0.403 & 0.275 & 0.376 & 0.612 & 0.708 & 1.128 & 0.249 & 0.667 \\
\texttt{LoRA-MLE} ($r = 40$) & DBS ($\lambda=1.0$) & 5 & 0.580 & 0.403 & 0.268 & 0.378 & 0.614 & 0.709 & 1.138 & 0.250 & 0.672 \\
\texttt{LoRA-MLE} ($r = 8$) & DBS ($\lambda=0.8$) & 10 & 0.481 & 0.684 & 0.274 & 0.373 & 0.606 & 0.709 & 1.114 & 0.259 & 0.665 \\
\texttt{LoRA-MLE} ($r = 40$) & DBS ($\lambda=0.8$) & 10 & 0.481 & 0.683 & 0.278 & 0.380 & 0.610 & 0.710 & 1.124 & 0.257 & 0.670 \\
\texttt{LoRA-MLE} ($r = 8$) & DBS ($\lambda=1.0$) & 10 & 0.486 & 0.678 & 0.276 & 0.372 & 0.607 & 0.710 & 1.115 & 0.259 & 0.665 \\
\texttt{LoRA-MLE} ($r = 40$) & DBS ($\lambda=1.0$) & 10 & 0.492 & 0.675 & 0.276 & 0.376 & 0.609 & 0.709 & 1.118 & 0.256 & 0.666 \\
\texttt{LoRA-MLE} ($r = 8$) & DBS ($\lambda=0.8$) & 25 & 0.402 & 0.756 & 0.273 & 0.376 & 0.607 & 0.703 & 1.129 & 0.251 & 0.673 \\
\texttt{LoRA-MLE} ($r = 40$) & DBS ($\lambda=0.8$) & 25 & 0.399 & 0.763 & 0.282 & 0.381 & 0.605 & 0.702 & 1.139 & 0.253 & 0.680 \\
\texttt{LoRA-MLE} ($r = 8$) & DBS ($\lambda=1.0$) & 25 & 0.402 & 0.757 & 0.271 & 0.377 & 0.608 & 0.703 & 1.128 & 0.251 & 0.672 \\
\texttt{LoRA-MLE} ($r = 40$) & DBS ($\lambda=1.0$) & 25 & 0.398 & 0.765 & 0.282 & 0.381 & 0.606 & 0.701 & 1.136 & 0.252 & 0.678 \\
\texttt{LoRA-MLE} ($r = 8$) & TTA BS ($\phi_1$) & 1 & 0.385 & 0.731 & 0.198 & 0.343 & 0.574 & 0.693 & 0.969 & 0.226 & 0.584 \\
\texttt{LoRA-MLE} ($r = 8$) & TTA BS ($\phi_2$) & 1 & 0.479 & 0.588 & 0.207 & 0.352 & 0.586 & 0.695 & 0.992 & 0.231 & 0.597 \\
\texttt{LoRA-MLE} ($r = 8$) & TTA BS ($\phi_3$) & 1 & 0.576 & 0.438 & 0.185 & 0.332 & 0.567 & 0.683 & 0.916 & 0.218 & 0.550 \\
\texttt{LoRA-MLE} ($r = 8$) & TTA BS ($\phi_1$) & 5 & 0.395 & 0.733 & 0.241 & 0.358 & 0.592 & 0.700 & 1.057 & 0.242 & 0.636 \\
\texttt{LoRA-MLE} ($r = 8$) & TTA BS ($\phi_2$) & 5 & 0.491 & 0.590 & 0.260 & 0.360 & 0.597 & 0.703 & 1.047 & 0.246 & 0.630 \\
\texttt{LoRA-MLE} ($r = 8$) & TTA BS ($\phi_3$) & 5 & \textbf{0.596} & 0.435 & 0.251 & 0.347 & 0.586 & 0.693 & 1.005 & 0.235 & 0.605 \\
\midrule\midrule
\texttt{LoRA-MoE} & BS & 5 & 0.396 & 0.766 & 0.274 & 0.381 & 0.608 & 0.702 & 1.129 & 0.252 & 0.674 \\
\texttt{LoRA-MoE} & BS & 10 & 0.411 & 0.746 & 0.288 & 0.385 & 0.613 & 0.703 & 1.161 & 0.256 & 0.692 \\
\texttt{LoRA-MoE} & BS & 25 & 0.415 & 0.736 & 0.287 & 0.385 & 0.612 & 0.705 & 1.154 & 0.258 & 0.689 \\
\texttt{LoRA-MoE} & DBS ($\lambda=0.8$) & 5 & 0.555 & 0.443 & 0.275 & 0.379 & 0.611 & 0.707 & 1.136 & 0.248 & 0.670 \\
\texttt{LoRA-MoE} & DBS ($\lambda=1.0$) & 5 & 0.578 & 0.409 & 0.268 & 0.377 & 0.609 & 0.708 & 1.130 & 0.247 & 0.667 \\
\texttt{LoRA-MoE} & DBS ($\lambda=0.8$) & 10 & 0.484 & 0.677 & 0.279 & 0.374 & 0.608 & 0.709 & 1.122 & 0.256 & 0.667 \\
\texttt{LoRA-MoE} & DBS ($\lambda=1.0$) & 10 & 0.488 & 0.675 & 0.283 & 0.374 & 0.608 & 0.710 & 1.119 & 0.257 & 0.666 \\
\texttt{LoRA-MoE} & DBS ($\lambda=0.8$) & 25 & 0.402 & 0.757 & 0.275 & 0.382 & 0.609 & 0.701 & 1.135 & 0.253 & 0.676 \\
\texttt{LoRA-MoE} & DBS ($\lambda=1.0$) & 25 & 0.401 & 0.759 & 0.275 & 0.382 & 0.609 & 0.701 & 1.133 & 0.253 & 0.675 \\
\texttt{LoRA-MoE} & SR BS & 1 & 0.256 & 0.909 & 0.165 & 0.318 & 0.538 & 0.670 & 0.865 & 0.200 & 0.526 \\
\texttt{LoRA-MoE} & SR BS & 2 & 0.269 & 0.895 & 0.195 & 0.326 & 0.549 & 0.675 & 0.905 & 0.210 & 0.551 \\
\texttt{LoRA-MoE} & SR BS & 5 & 0.261 & 0.907 & 0.196 & 0.331 & 0.550 & 0.678 & 0.925 & 0.216 & 0.563 \\
\midrule\midrule
\texttt{LoRA-MCL} ($\varepsilon=0.0005$) & BS & 1 & \underline{0.580} & \textbf{0.374} & 0.259 & 0.377 & 0.613 & 0.711 & 1.119 & 0.248 & 0.662 \\
\texttt{LoRA-MCL} ($\varepsilon=0.05$) & BS & 1 & 0.551 & 0.427 & 0.273 & 0.382 & 0.622 & 0.712 & 1.181 & 0.253 & 0.695 \\
\texttt{LoRA-MCL} (annealed) & BS & 1 & 0.570 & \underline{0.397} & 0.275 & 0.379 & 0.615 & 0.713 & 1.149 & 0.255 & 0.680 \\
\texttt{LoRA-MCL} ($\varepsilon=0.0005$) & BS & 2 & 0.575 & 0.401 & 0.277 & 0.385 & 0.626 & 0.712 & 1.152 & 0.258 & 0.684 \\
\texttt{LoRA-MCL} ($\varepsilon=0.05$) & BS & 2 & 0.544 & 0.464 & 0.298 & 0.394 & 0.631 & 0.712 & 1.209 & 0.260 & 0.714 \\
\texttt{LoRA-MCL} (annealed) & BS & 2 & 0.574 & 0.411 & 0.309 & 0.392 & 0.632 & \underline{0.714} & 1.205 & 0.264 & 0.712 \\
\texttt{LoRA-MCL} ($\varepsilon=0.0005$) & BS & 5 & 0.579 & 0.410 & 0.306 & 0.392 & 0.631 & 0.712 & 1.190 & 0.263 & 0.706 \\
\texttt{LoRA-MCL} ($\varepsilon=0.05$) & BS & 5 & 0.542 & 0.491 & \underline{0.309} & \textbf{0.399} & \underline{0.636} & \textbf{0.715} & \textbf{1.237} & \underline{0.265} & \textbf{0.728} \\
\texttt{LoRA-MCL} (annealed) & BS & 5 & 0.575 & 0.423 & \textbf{0.315} & \underline{0.398} & \textbf{0.636} & 0.714 & \underline{1.211} & \textbf{0.268} & \underline{0.716} \\
\bottomrule
\end{tabular}

    }
    \end{center}
    \label{tab:full_results_ac_5_hyp}
    \end{table}

\begin{table}
    \caption{\textbf{Results for AudioCaps with 5 hypotheses and Sampling-based Decoding.}}
    \begin{center}
    \resizebox{\columnwidth}{!}{
        \begin{tabular}{lcccccccccccc}
\toprule
Training & Decoding & $\mathrm{Div}_{2}$ & $\mathrm{mBLEU}_{4}$ & $\mathrm{BLEU}_{4}$ & $\mathrm{METEOR}$ & $\mathrm{ROUGE}_{L}$ & $\mathrm{sBERT}$ & $\mathrm{CIDEr}_{D}$ & $\mathrm{SPICE}$ & $\mathrm{SPIDEr}$ \\
\midrule
\texttt{LoRA-MLE} ($r = 8$) & Top-$k$ sampling & 0.711 & 0.173 & 0.169 & 0.335 & 0.549 & 0.701 & 0.906 & 0.229 & 0.543 \\
\texttt{LoRA-MLE} ($r = 8$) & Nucleus (Top-$p$) sampling & 0.697 & 0.193 & 0.180 & \textbf{0.344} & 0.564 & \underline{0.704} & 0.940 & 0.234 & 0.563 \\
\texttt{LoRA-MLE} ($r = 8$) & Typical $p$ sampling & 0.699 & 0.189 & 0.182 & \underline{0.343} & 0.564 & 0.704 & 0.945 & 0.233 & 0.566 \\
\midrule\midrule
\texttt{LoRA-MCL} (annealed) & Top-$k$ sampling & \textbf{0.735} & \textbf{0.149} & 0.175 & 0.342 & \underline{0.568} & \textbf{0.705} & 0.938 & 0.233 & 0.563 \\
\texttt{LoRA-MCL} (annealed) & Nucleus (Top-$p$) sampling & 0.724 & 0.163 & \underline{0.182} & 0.340 & 0.566 & 0.702 & \underline{0.950} & \underline{0.235} & \underline{0.569} \\
\texttt{LoRA-MCL} (annealed) & Typical $p$ sampling & \underline{0.724} & \underline{0.160} & \textbf{0.186} & 0.341 & \textbf{0.568} & 0.703 & \textbf{0.953} & \textbf{0.235} & \textbf{0.570} \\
\bottomrule
\end{tabular}
    }
    \end{center}
     \label{tabapx:sampling_audiocaps}
    \end{table}

\newpage

\subsubsection{Qualitative Examples}

We provide some qualitative examples of the predictions on AudioCaps. Here, \texttt{LoRA-MCL} uses $\varepsilon=0.05$, Beam Search with $B = 5$ and $K=5$, \texttt{LoRA-MoE} uses Beam search with $B = 25$.

\textit{Example 1.} References:
\begin{itemize}
    \item A large truck driving by as an emergency siren wails and truck horn honks
    \item A wailing siren fades, a motor sputters, then the siren resumes accompanied by blaring horns
    \item An emergency siren ringing with car horn honking
    \item A fire truck engine runs and the siren is blowing but stops, traffic is present, the fire truck horn honks twice, and the siren begins again
    \item A fire engine with a siren fading then another loud siren
\end{itemize}

\textbf{\texttt{LoRA-MoE.}}\\
$\{$An emergency vehicle passes by and blows its horn followed by the siren of an emergency vehicle$\}$\\
$\{$An emergency vehicle passes and blows its horn followed by the siren of an emergency vehicle$\}$\\
$\{$A truck engine running followed by a loud horn and then an emergency siren$\}$\\
$\{$A truck engine running followed by a loud horn and an emergency siren$\}$\\
$\{$An emergency vehicle siren sounds as a truck passes by and then another emergency vehicle siren sounds$\}$\\

\textbf{\texttt{LoRA-MCL.}}\\
$\{$An emergency vehicle siren is going off$\}$\\
$\{$An emergency vehicle siren sounds, followed by a truck horn honking and an emergency vehicle siren$\}$\\
$\{$An emergency siren is triggered and a vehicle moves$\}$\\
$\{$Fire truck siren and engine revving$\}$\\
$\{$A fire truck siren sounds as a vehicle passes and then another fire truck siren sounds$\}$\\

\textit{Example 2.} References:
\begin{itemize}
    \item A man speaks as birds chirp and dogs bark
    \item A man is speaking as birds are squawking, and a dog barks
    \item A man talks while several animals make noises in the background
    \item A man speaking followed by dogs barking alongside chimps screaming and birds chirping
    \item A man speaking as monkeys scream and dogs bark followed by birds cawing in the distance
\end{itemize}

\textbf{\texttt{LoRA-MoE.}}\\
$\{$A man is speaking and dogs are barking$\}$\\
$\{$A man speaking with dogs barking in the background$\}$\\
$\{$A man is speaking and a dog is barking$\}$\\
$\{$A man is speaking and dogs are barking in the background$\}$\\
$\{$An adult male is speaking, and dogs are barking and whimpering$\}$\\

\textbf{\texttt{LoRA-MCL.}}\\
$\{$A man speaking and dogs barking$\}$\\
$\{$A man is narrating and a dog is barking in the background$\}$\\
$\{$A man is speaking and dogs are barking$\}$\\
$\{$Man speaking with dog barking in the background$\}$\\
$\{$A man speaks with dogs barking and birds chirping in the background$\}$\\

\textit{Example 3.} References:
\begin{itemize}
\item An engine running and wind with various speech in the background
\item A motorboat engine operating as a crowd of people talk followed by metal creaking and a man speaking
\item A large motor is running smoothly, water is splashing, people are talking in the background, and an adult male speaks in the distance
\item A ship engine running as a crowd of people talk followed by a ship hull creaking as wind blows heavily into a microphone
\item Outdoor noise from a water vehicle as people are talking
\end{itemize}

\textbf{\texttt{LoRA-MoE.}}\\
$\{$Humming of an idling engine followed by a horn honking$\}$\\
$\{$Humming of an idling engine followed by a horn sounding$\}$\\
$\{$An engine running followed by a horn sounding$\}$\\
$\{$An engine running followed by a horn honking$\}$\\
$\{$Humming of an idling engine followed by a honking horn$\}$\\

\textbf{\texttt{LoRA-MCL.}}\\
$\{$A boat motor is running and people are talking in the background$\}$\\
$\{$An aircraft engine running as people talk in the background$\}$\\
$\{$An engine is running and people are talking$\}$\\
$\{$An engine running consistently with people talking in the background$\}$\\
$\{$Humming of an engine with distant murmuring$\}$

\newpage

\subsection{Image Captioning Experiments}
\subsubsection{Experimental setup}
\label{apx:exp_details_image}

\begin{table}[h]
    \caption{\textbf{Quality and Diversity Evaluation on TextCaps with $3$ candidates}. For each of the presented metrics, higher is better $(\uparrow)$ except for mBLEU-4 $(\downarrow)$. \texttt{LoRA-MCL} is trained with $\varepsilon=0.1$, $r = 8$ and $\alpha = 32$. \texttt{LoRA-MLE} is trained with $r = 24$ and $\alpha = 96$. For completeness, we also trained \texttt{LoRA-MLE} with $r = 8$ and $\alpha = 32$ in the rows marked with $^\dag$. PaliGemma-3B rows are shown in \textcolor{gray}{gray}, as this model was fully fine-tuned and is therefore not directly comparable to LoRA-based methods.}
    \begin{center}
    \resizebox{\columnwidth}{!}{
        \begin{tabular}{lcccccccccccc}
\toprule
Training & Decoding & Beam & $\mathrm{Div}_{2}$ & $\mathrm{mBLEU}_{4}$ & $\mathrm{BLEU}_{1}$ & $\mathrm{BLEU}_{4}$ & $\mathrm{METEOR}$ & $\mathrm{ROUGE}_{L}$ & $\mathrm{sBERT}$ & $\mathrm{CIDEr}_{D}$ & $\mathrm{SPICE}$ & $\mathrm{SPIDEr}$ \\
\midrule
\texttt{LoRA-MLE} & BS & 3 & 0.509 & 0.688 & 0.802 & 0.318 & 0.315 & 0.580 & 0.670 & 1.517 & 0.244 & 0.873 \\
\texttt{LoRA-MLE} & BS & 6 & 0.457 & 0.786 & 0.795 & 0.338 & 0.326 & 0.583 & 0.671 & 1.557 & 0.246 & 0.895 \\
 \texttt{LoRA-MLE}$^\dag$ & BS & 3 & 0.421 & 0.833 & 0.788 & 0.315 & 0.317 & 0.573 & 0.670 & 1.517 & 0.241 & 0.874 \\
\texttt{LoRA-MLE}$^\dag$ & BS & 6 & 0.456 & 0.784 & 0.796 & 0.339 & 0.327 & 0.585 & 0.672 & 1.572 & 0.248 & 0.903 \\
\texttt{LoRA-MLE} & DBS ($\lambda=0.5$) & 3 & 0.600 & 0.529 & 0.818 & 0.345 & 0.325 & 0.596 & 0.684 & 1.571 & 0.253 & 0.902 \\
\texttt{LoRA-MLE} & DBS ($\lambda=0.8$) & 3& 0.655 & 0.437 & 0.824 & 0.349 & 0.327 & 0.601 & 0.686 & 1.590 & 0.251 & 0.909 \\
\texttt{LoRA-MLE} & DBS ($\lambda=1.0$) & 3 & \textbf{0.669} & \textbf{0.416} & 0.822 & 0.348 & 0.326 & 0.599 & 0.685 & 1.586 & 0.250 & 0.906 \\
\texttt{LoRA-MLE} & DBS ($\lambda=0.5$) & 6 & 0.532 & 0.694 & 0.813 & 0.344 & 0.328 & 0.595 & 0.681 & 1.580 & 0.253 & 0.908 \\
\texttt{LoRA-MLE} & DBS ($\lambda=0.8$) & 6 & 0.549 & 0.671 & 0.812 & 0.341 & 0.328 & 0.593 & 0.681 & 1.573 & 0.251 & 0.903 \\
\texttt{LoRA-MLE} & DBS ($\lambda=1.0$) & 6 & 0.553 & 0.666 & 0.812 & 0.340 & 0.328 & 0.592 & 0.680 & 1.577 & 0.250 & 0.904 \\
\texttt{LoRA-MLE}$^\dag$ & DBS ($\lambda=0.5$) & 3 & 0.597 & 0.531 & 0.821 & 0.346 & 0.326 & 0.596 & 0.685 & 1.589 & 0.255 & 0.912 \\
\texttt{LoRA-MLE}$^\dag$ & DBS ($\lambda=0.8$) & 3 & 0.597 & 0.531 & 0.821 & 0.346 & 0.326 & 0.596 & 0.685 & 1.589 & 0.255 & 0.912 \\
\texttt{LoRA-MLE}$^\dag$ & DBS ($\lambda=1.0$) & 3 & 0.665 & 0.425 & 0.827 & \underline{0.357} & 0.327 & 0.601 & 0.686 & 1.601 & 0.252 & 0.915 \\
\texttt{LoRA-MLE}$^\dag$ & DBS ($\lambda=0.5$) & 6 & 0.528 & 0.697 & 0.808 & 0.343 & 0.329 & 0.594 & 0.681 & 1.586 & 0.253 & 0.911 \\
\texttt{LoRA-MLE}$^\dag$ & DBS ($\lambda=0.8$) & 6 & 0.542 & 0.681 & 0.810 & 0.340 & 0.329 & 0.593 & 0.680 & 1.583 & 0.253 & 0.909 \\
\texttt{LoRA-MLE}$^\dag$ & DBS ($\lambda=1.0$) & 6 & 0.551 & 0.671 & 0.810 & 0.341 & 0.328 & 0.592 & 0.680 & 1.584 & 0.251 & 0.908 \\
\midrule
\midrule
\texttt{LoRA-MoE} & DBS ($\lambda=0.5$) & 3 & 0.600 & 0.529 & 0.822 & 0.347 & 0.327 & 0.598 & 0.684 & 1.610 & \textbf{0.258} & 0.923 \\
\texttt{LoRA-MoE} & DBS ($\lambda=0.8$) & 3 & 0.654 & 0.441 & \textbf{0.828} & 0.353 & 0.327 & \underline{0.603} & 0.685 & 1.616 & 0.254 & 0.924 \\
\texttt{LoRA-MoE} & DBS ($\lambda=1.0$) & 3 & \underline{0.666} & \underline{0.421} & \textbf{0.828} & 0.354 & 0.328 & 0.602 & 0.685 & 1.622 & 0.253 & 0.926 \\
\texttt{LoRA-MoE} & DBS ($\lambda=0.5$) & 6 & 0.530 & 0.698 & 0.814 & 0.348 & \underline{0.331} & 0.595 & 0.681 & 1.607 & 0.257 & 0.923 \\
\texttt{LoRA-MoE} & DBS ($\lambda=0.8$) & 6 & 0.545 & 0.678 & 0.813 & 0.349 & 0.330 & 0.596 & 0.680 & 1.608 & 0.255 & 0.922 \\
\texttt{LoRA-MoE} & DBS ($\lambda=1.0$) & 6 & 0.551 & 0.670 & 0.813 & 0.346 & 0.330 & 0.596 & 0.679 & 1.607 & 0.254 & 0.921 \\
\texttt{LoRA-MoE} & SR BS & 1 & 0.556 & 0.597 & 0.809 & 0.315 & 0.311 & 0.576 & 0.678 & 1.541 & 0.243 & 0.883 \\
\midrule
\midrule
\textcolor{gray}{PaliGemma-3B (ft)} & \textcolor{gray}{DBS ($\lambda=0.5$)} & \textcolor{gray}{3} & \textcolor{gray}{0.595} & \textcolor{gray}{0.532} & \textcolor{gray}{0.832} & \textcolor{gray}{0.369} & \textcolor{gray}{0.330} & \textcolor{gray}{0.608} & \textcolor{gray}{0.688} & \textcolor{gray}{1.658} & \textcolor{gray}{0.255} & \textcolor{gray}{0.947} \\
\textcolor{gray}{PaliGemma-3B (ft)} & \textcolor{gray}{DBS ($\lambda=0.8$)} & \textcolor{gray}{3} & \textcolor{gray}{0.635} & \textcolor{gray}{0.467} & \textcolor{gray}{0.835} & \textcolor{gray}{0.374} & \textcolor{gray}{0.332} & \textcolor{gray}{0.611} & \textcolor{gray}{0.689} & \textcolor{gray}{1.661} &\textcolor{gray}{0.257} & \textcolor{gray}{0.949} \\
\textcolor{gray}{PaliGemma-3B (ft)} & \textcolor{gray}{DBS ($\lambda=1.0$)} & \textcolor{gray}{3} & \textcolor{gray}{0.655} & \textcolor{gray}{0.439} & \textcolor{gray}{0.833} & \textcolor{gray}{0.368} & \textcolor{gray}{0.332} & \textcolor{gray}{0.609} & \textcolor{gray}{0.687} & \textcolor{gray}{1.654} & \textcolor{gray}{0.256} & \textcolor{gray}{0.944}\\
\midrule
\midrule
\texttt{LoRA-MCL} & BS & 1 & 0.599 & 0.520 & \textbf{0.828} & 0.344 & 0.330 & 0.597 & \textbf{0.690} & \textbf{1.674} & 0.255 & \textbf{0.955} \\
\texttt{LoRA-MCL} & BS & 2 & 0.618 & 0.490 & 0.824 & \textbf{0.360} & \textbf{0.333} & \textbf{0.604} & \underline{0.687} & \underline{1.627} & \textbf{0.258} & \underline{0.932} \\
\bottomrule
\end{tabular}
    }
    \end{center}
    \label{tab:full_results_textcaps}
    \end{table}

We used LLaVA 1.6 with Vicuna-7B \citep{zheng2023judging} for the LLM, as the base model, which features 7.1 billion parameters. We used the \href{https://github.com/haotian-liu/LLaVA?tab=readme-ov-file}{official codebase} for the implementation. We trained using \texttt{bfloat16} precision. We used LoRA adapters applied to the $Q, K, V$, up and down linear projections of each block of the language model. Fine-tuning hyperparameters follow those from the \href{https://github.com/haotian-liu/LLaVA/blob/main/scripts/v1_5/finetune_task_lora.sh}{recipe} provided by the authors, except that we used smaller batch sizes and a lower LoRA rank due to compute constraints. We used a rank $r$, with $r = 8$ and $\alpha = 32$ unless otherwise stated, with dropout equal to $0.1$ (enabled during training, and disabled during inference). We trained each model with a batch size of $8$, with Adam optimizer \citep{diederik2014adam} (with $\beta_{1}=0.9$, and $\beta_{2}=0.999$), using a cosine scheduler with maximum learning rate of $2 \times 10^{-4}$, with a warmup ratio of $0.03$. Maximum sequence length is set to $2048$. Gradient clipping is used with a maximum gradient norm of $1.0$. We used 1 epoch for training, where we duplicated the image as many times as the number of its captions, such that the model sees exactly one time each caption.

\subsubsection{Additional results}
\label{apx:add_results}
\textbf{Comparing with PaliGemma-3B.} Several works have explored improving caption diversity in image captioning \cite{wang2019describing, wang2020diversity, mahajan2020diverse}. Among those evaluating on TextCaps, \citet{zhang2022magic} and \citet{xu2021towards} report results, but their performance is far below that of recent VLMs, such as PaliGemma-3B \cite{beyer2024paligemma}. For example, the reported best corpus-level CIDEr on the TextCaps validation set are: 76.6 for \citet{zhang2022magic}, 95.5 for \citet{xu2021towards}, 127.48 for PaliGemma-3B (224x224). We therefore compare with PaliGemma-3B, which provides a publicly available \href{https://huggingface.co/google/paligemma-3b-ft-textcaps-224}{fine-tuned version} on Hugging Face. We evaluate it using DBS returning $K = 3$ hypotheses (Table \ref{tab:full_results_textcaps}). PaliGemma is a strong baseline, arguably the state-of-the-art open-weight model on TextCaps. Note, however, that the comparison is not entirely fair, as PaliGemma undergoes full-weight fine-tuning (instead of LoRA). Despite this, our method matches and even slightly improves its performance (Oracle SPIDEr of $0.955$ vs. $0.949$). We also expect that applying \texttt{LoRA-MCL} to PaliGemma could further improve the results.

The full results presented in Table \ref{tab:full_results_textcaps} show that \texttt{LoRA-MCL} tends to outperform \texttt{LoRA-MLE} with Beam Search (BS) and Diverse Beam Search (DBS) in terms of quality, although DBS produces more varied outputs. Depending on the rank ($r = 8$ or $r = 24$), we found that setting $\lambda$ to $1.0$ or $\lambda = 0.8$, respectively, yields the best quality scores for DBS. Similar to our experiments on Audio Captioning, increasing the rank of \texttt{LoRA-MLE} results in a slight degradation in performance while improving diversity. Additionally, increasing the number of beams in BS with \texttt{LoRA-MLE} results in slightly improved performance but reduced diversity. In contrast, increasing the number of beams in DBS with \texttt{LoRA-MLE} (with $\lambda$ fixed) leads to declines in both quality and diversity. Interestingly, with \texttt{LoRA-MCL}, increasing the number of beams enhances both performance and diversity here.

\subsubsection{Artificial multilingual dataset creation}

To evaluate the behaviour of \texttt{LoRA-MCL} under a multi-modal distribution, we simulated an artificial bi-modal dataset by automatically translating half of the captions from English to French using T5-small~\citep{T5}, while keeping the prompts in English. More specifically, we randomly sampled half of the images and translated their five associated captions.
All the training parameters are the same as those in the experiments on the original TextCaps dataset, except the learning rate, which we set at $2 \times 10^{-5}$ (as the maximum value in the scheduler) in both the \texttt{LoRA-MLE} and \texttt{LoRA-MCL}.

During evaluation, to assess which head is considered as the winner (for the head specialization analysis), we selected the one that maximizes the SPIDEr score over the references of the given sample.

\subsubsection{Specialization of the hypotheses}
\label{apx:specialization}

Understanding what each hypothesis learns is difficult. The founding paper of MCL by \citet{lee2016stochastic} already introduced the notion of specialization of the hypotheses in the context of classification (see \citet{lee2016stochastic} in Figure 4). We provide additional insights with the French/English controlled experiment of Section \ref{sec:specialization}, where we observed some ``unsupervised'' specialization of the hypotheses; one of the hypotheses learned English, and the other learned French. We tried to visualize the hypotheses predictions to understand what the hypotheses have learned in Figure \ref{fig:pca_toy}.

\textbf{Quantifying specialization.} To quantify specialization, we embed every generated caption on the evaluation set (for hypotheses $k \in \{1,\dots,K\}$ and each example $i \in \{1,\dots,N\}$) using a Sentence-BERT model (StyleDistance \cite{patel2025styledistance}). We trained a linear Support Vector Machine (SVM) \cite{cortes1995support} on this space to predict which head produced each caption, using a $70/30$ train-test split on captions. In the French/English experiment, the SVM achieves $99.8\%$ test accuracy for the two-head \texttt{LoRA-MCL} model of Section \ref{sec:specialization}, confirming specialization. Embeddings of the hypotheses are visualized using a Principal Components Analysis (PCA) with two components in Figure \ref{fig:pca_toy}. Outside this controlled setup, we repeated this analysis on captions from the \texttt{LoRA-MCL} trained on AudioCaps with $K=5$. The SVM reaches $63.7\%$ test-accuracy, over three times better than random choice, indicating a clear specialization of the heads. In contrast, applying the same procedure to captions from the \texttt{LoRA-MLE} baseline (with DBS decoding, $\lambda = 0.8$, and beam size = 5) yielded $23.2\%$ accuracy on the test set after SVM fitting, close to random choice and showing no evidence of specialization, likewise for \texttt{LoRA-MoE} (yielding $19.8\%$ accuracy in the same setup). We believe that further understanding of how specialization emerges represents a promising direction for future work.

\begin{figure}[H]
    \centering
    \begin{subfigure}{0.32\linewidth}
        \centering
        \includegraphics[width=\linewidth]{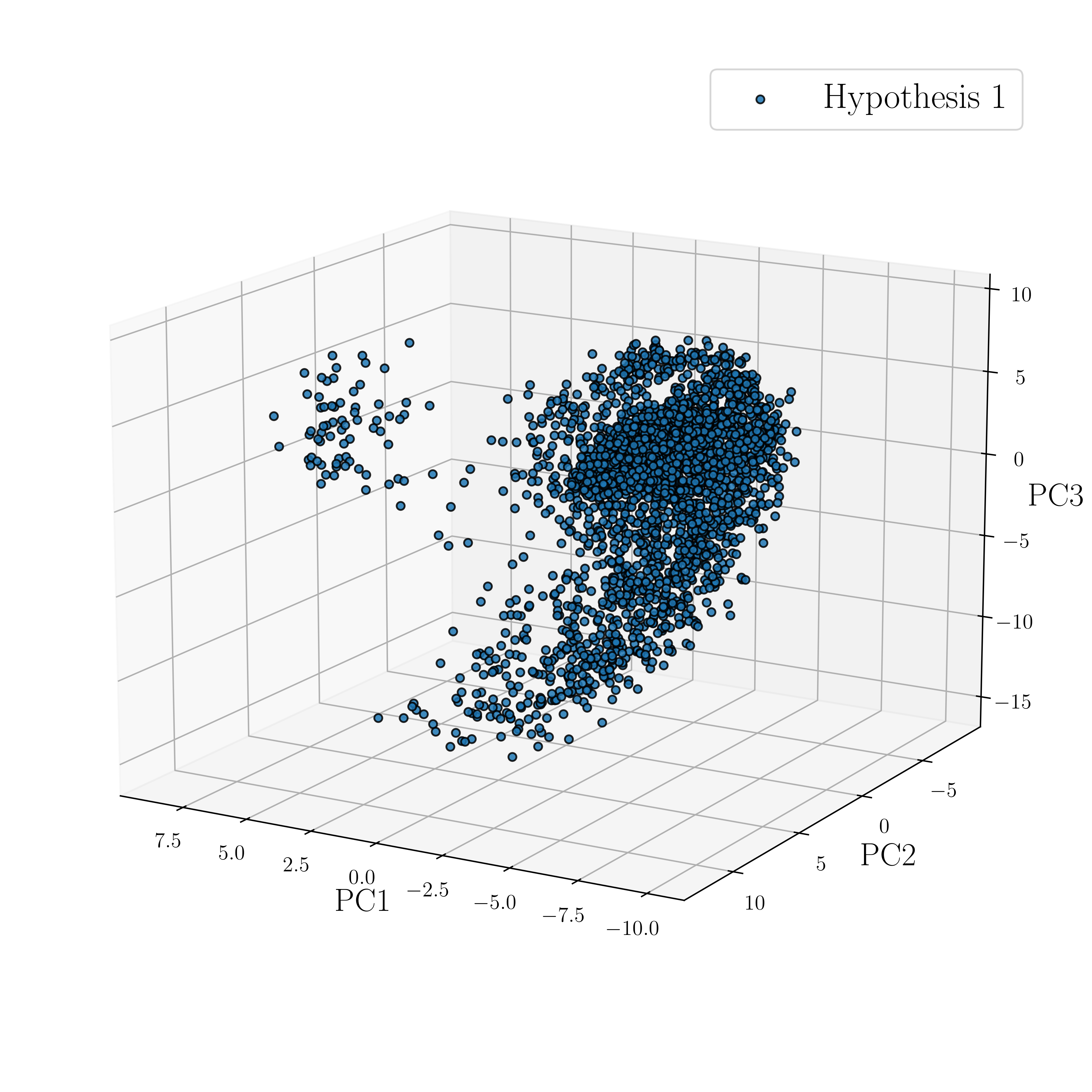}
        \caption{$k=1$}
    \end{subfigure}
    \hfill
    \begin{subfigure}{0.32\linewidth}
        \centering
        \includegraphics[width=\linewidth]{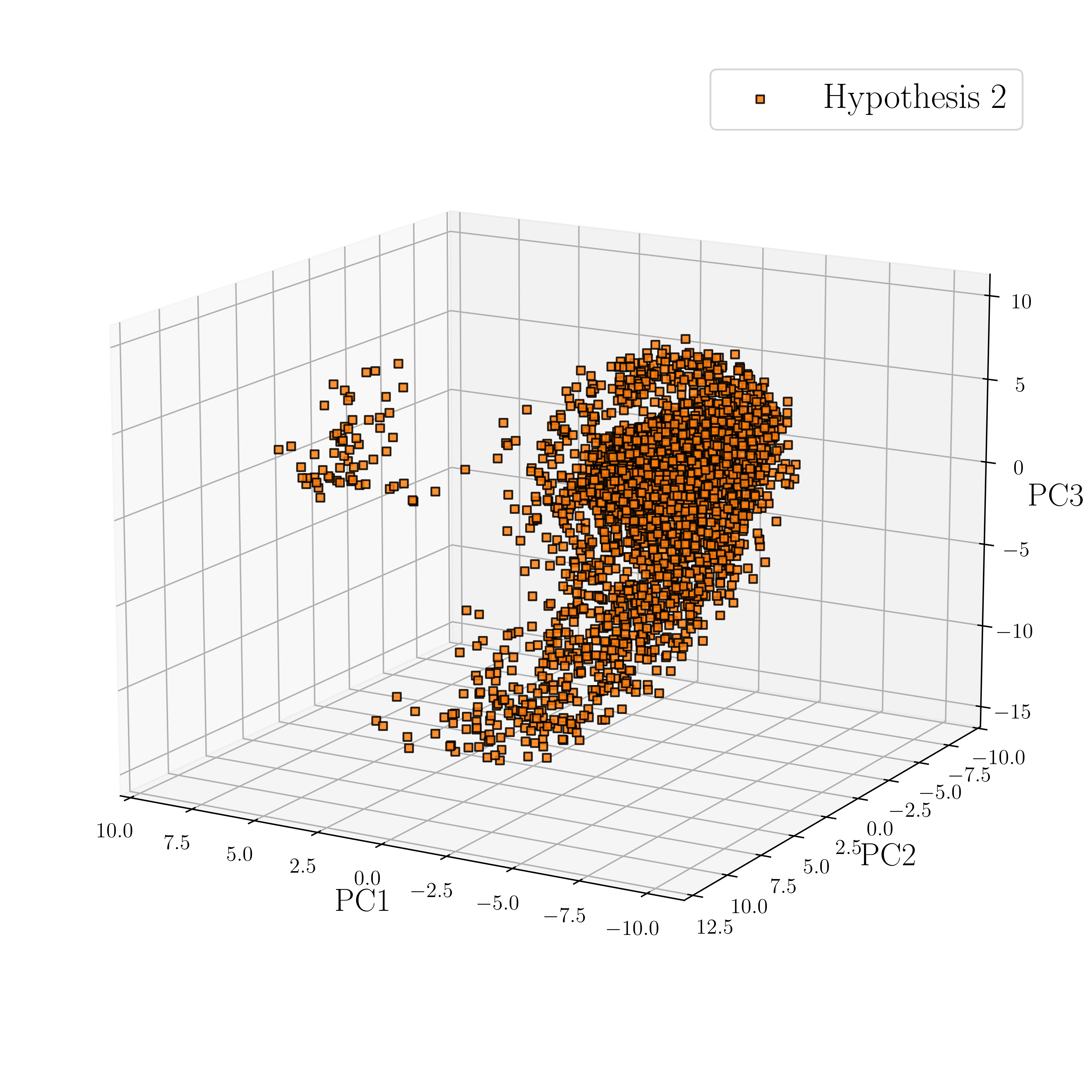}
        \caption{$k=2$}
    \end{subfigure}
    \hfill
    \begin{subfigure}{0.32\linewidth}
        \centering
        \includegraphics[width=\linewidth]{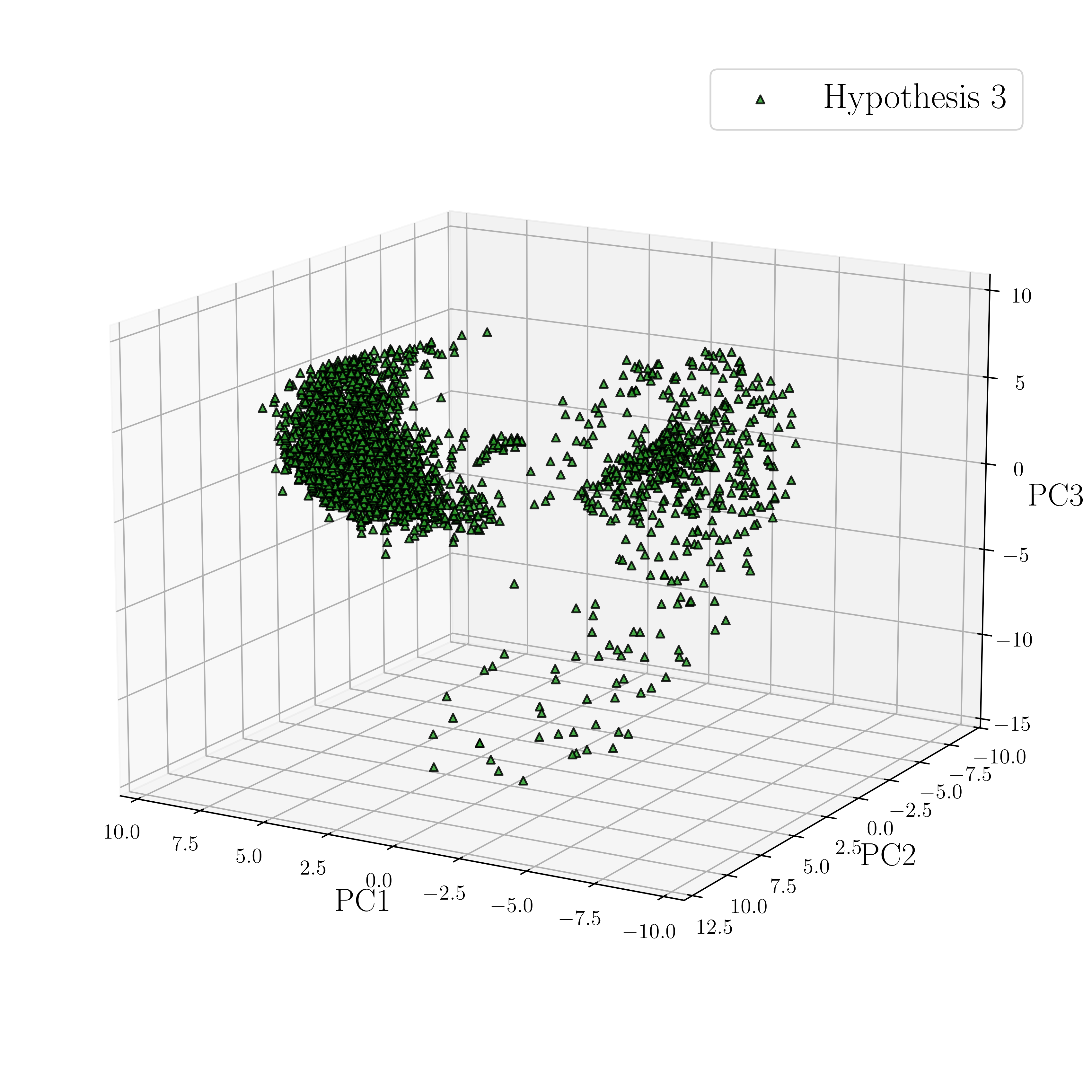}
        \caption{$k=3$}
    \end{subfigure}

    \vspace{0.5em}

    \begin{subfigure}{0.7\linewidth}
        \centering
        \includegraphics[width=\linewidth]{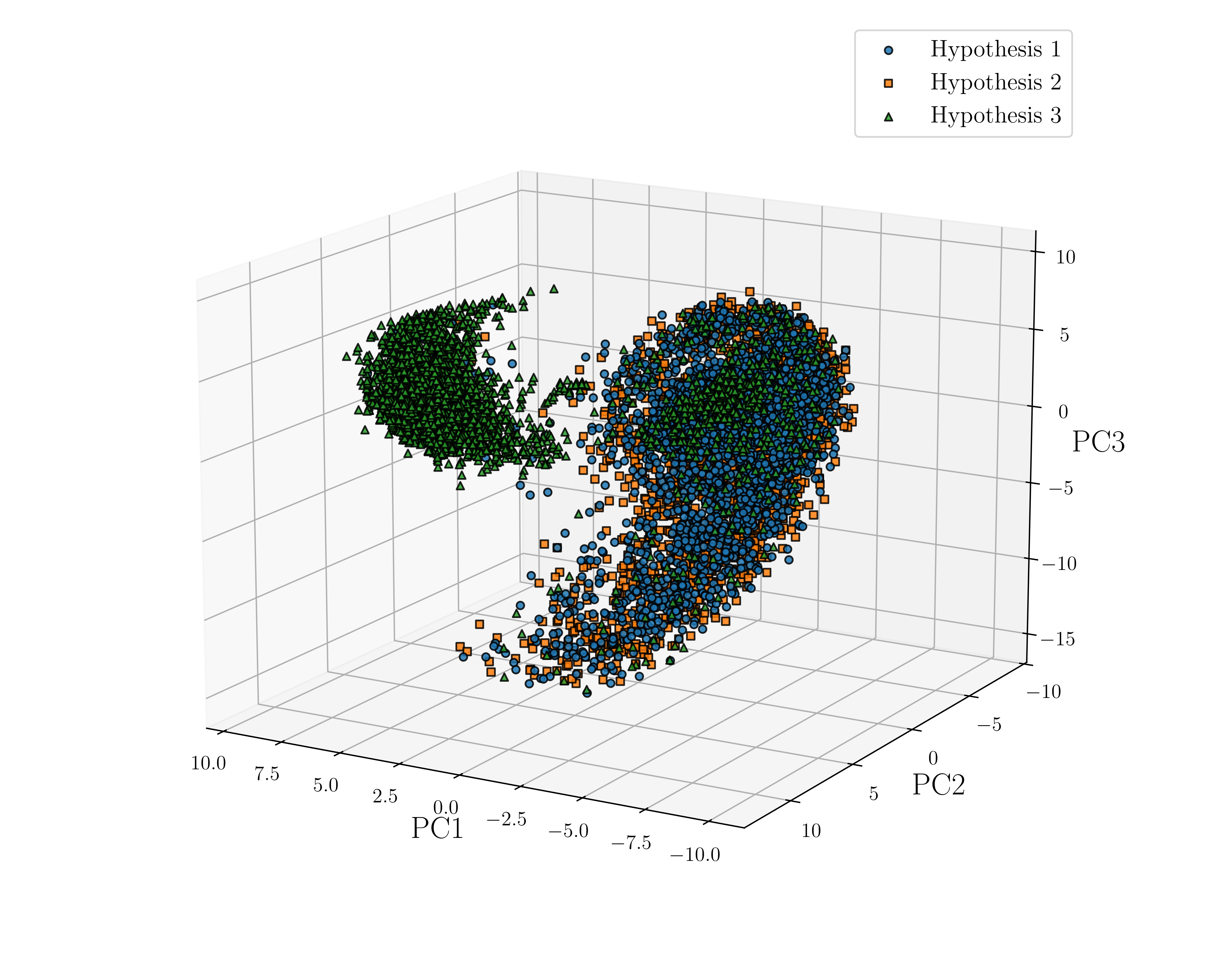}
        \caption{Full embedding space}
    \end{subfigure}
    \caption{\textbf{PCA of hypothesis candidate embeddings of \texttt{LoRA-MCL} with $K = 3$ on the bilingual experiment setup.} PC1, PC2 and PC3 correspond to the first three principal components, explaining $25.66\%$, $18.07\%$ and $13.61\%$ of the variance respectively. The three subplots (a), (b) and (c) visualization for each hypothesis $k$.}
    \label{fig:pca_combined-3d}
\end{figure}

\textbf{Repeating the experiment with $K = 3$.} To study $K$ larger than the true number of modes, we repeat the bilingual experiment with $K = 3$, using the same experimental setup otherwise. We observed that one hypothesis tends to specialize in French, while the other two specialize in English. Indeed, a PCA of the candidate embeddings (Figure \ref{fig:pca_combined-3d}) shows that hypothesis 3 occupies a largely distinct region from hypotheses 1 and 2, although there is a slight overlap. Table~\ref{tab:winning_rate} further confirms this via winning scores in each language.

\newpage

\begin{table}[h]
    \centering
    \small
    \begin{tabular}{lccc}
        \toprule
        & $k=1$ & $k=2$ & $k=3$ \\
        \midrule
        English & 54 & 39 & 7 \\
        French  & 20 & 13 & 67 \\
        \bottomrule
    \end{tabular}
    \vspace{4pt}
    \caption{\textbf{Winning rate (\%)} in the experiment of Figure~\ref{fig:pca_combined-3d}. 
    For each reference sentence (in either English or French), we select the hypothesis index maximizing the BLEU-4 score. We see that the hypothesis $k = 3$ tends to specialize in French.}
    \label{tab:winning_rate}
\end{table}

\newpage

\subsubsection{Qualitative Examples}
In this section, we show some qualitative examples (image–predicted caption pairs) that highlight the behavior of \texttt{LoRA-MCL} compared to a baseline model (\texttt{LoRA-MLE} with $r=24$). When the scene contains numerous objects or pieces of information, \texttt{LoRA-MCL} (through its different heads) covers a wider variety of descriptions than diverse beam search decoding. We performed inference with \texttt{LoRA-MCL} using greedy decoding, while we used diverse beam search ($\lambda=0.8$ and using $3$ beams) for the baseline model.
\begin{figure}[h!]
\centering
\begin{minipage}{0.35\textwidth}
    \includegraphics[width=\linewidth]{Figure9_shelf.png}
\end{minipage}
\hfill
\begin{minipage}{0.64\textwidth}
    \begin{tabular}{p{0.45\linewidth} p{0.45\linewidth}}
        \textbf{\texttt{LoRA-MLE.}} & \textbf{\texttt{LoRA-MCL.}}\\
        \{A display of Hugo Victor products on a table.\}
        &
        \{A display of Hugo Victor products is set up in a store.\} \\
        \{A display for Hug Victor with a sign that says Hug Victor.\} &
        \{A store display for Hugo Victor with a shelf full of items.\} \\
        \{a store display for Hugo Victor with many items\} &
        \{A store display for Hugo Victor features a variety of plates, bowls, and books.\}
    \end{tabular}
\end{minipage}
\end{figure}

\begin{figure}[h!]
\centering
\begin{minipage}{0.35\textwidth}
    \includegraphics[width=\linewidth]{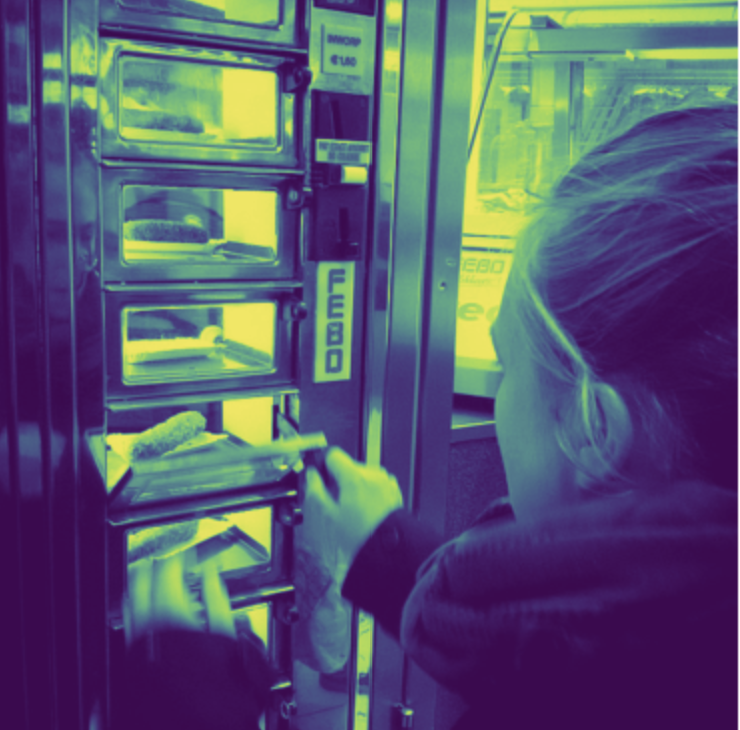}
\end{minipage}
\hfill
\begin{minipage}{0.64\textwidth}
    \begin{tabular}{p{0.45\linewidth} p{0.45\linewidth}}
         \textbf{\texttt{LoRA-MLE.}} & \textbf{\texttt{LoRA-MCL.}}\\
        \{a person getting food from a vending machine with the number 680 on it\}
        &
        \{A girl is putting food into a vending machine that says FEEDO.\}\\
        \{A woman is getting food from a vending machine that says "Fresh" on it.\} &
        \{A woman is putting money into a vending machine that says 100.\} \\
        \{A girl puts a sandwich into a vending machine that says "Fresh" on it.\}  &
        \{A girl is getting a hot dog from a vending machine.\}  \\
    \end{tabular}
\end{minipage}
\end{figure}

\begin{figure}[h!]
\centering
\begin{minipage}{0.35\textwidth}
    \includegraphics[width=\linewidth]{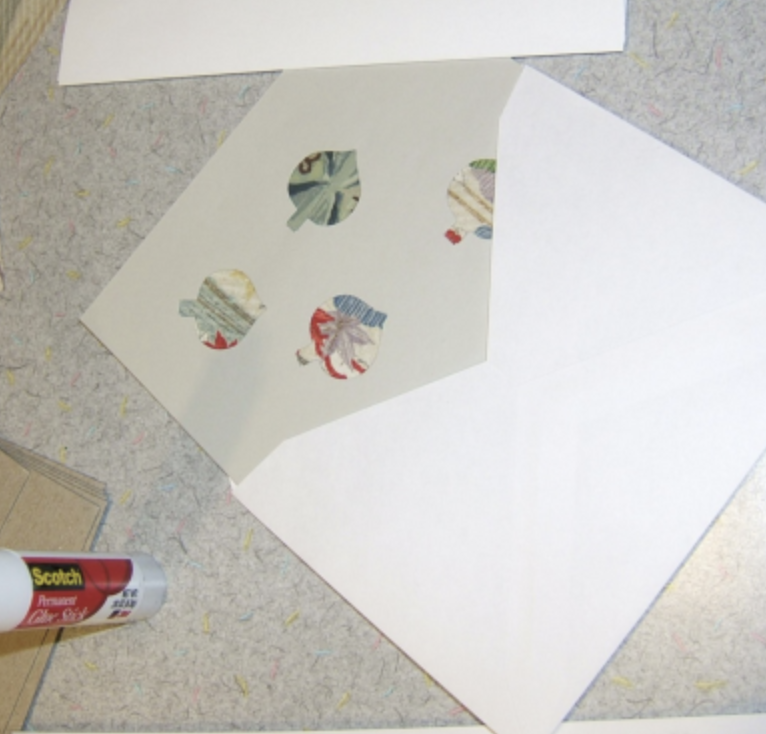}
\end{minipage}
\hfill
\begin{minipage}{0.64\textwidth}
    \begin{tabular}{p{0.45\linewidth} p{0.45\linewidth}}
         \textbf{\texttt{LoRA-MLE.}} & \textbf{\texttt{LoRA-MCL.}}\\
        \{
        A Scotch brand glue stick is on a table.\}
        &
        \{A bottle of Scotch tape sits on a table.\}\\
        \{A can of Scotch branded glue is on a table.\} &
        \{A bottle of Scotch tape sits on a table next to a piece of paper.\} \\
        \{a scotch tape that is on the ground\} &
        \{A Scotch tape is on the table next to a piece of paper.\} \\
    \end{tabular}
\end{minipage}
\end{figure}

\subsection{Diverse Machine Translation}
\label{apx:diverse_nmt}

\subsubsection{Experimental setup}

We fine-tuned ALMA-7B (based on LLaMA-2-7B~\cite{touvron2023llama}) for two epochs on the human-written data collected by \citet{xuparadigm}, which includes the WMT’17–WMT’20 test sets and the development and test sets from Flores-200~\cite{costa2022no}. Our fine-tuning setup follows \citet{xuparadigm}, except that we employ \texttt{bf16} precision instead of \texttt{fp16}. We use $r = \tfrac{\alpha}{2} = 16$, a maximum sequence length of $512$, an effective batch size of $256$, a warm-up ratio of $0.01$, and a peak learning rate of $2 \times 10^{-5}$ with an inverse square-root scheduler.

We follow the quality–diversity evaluation protocol of \citet{shen2019mixture}, computing BLEU scores with the \href{https://github.com/mjpost/sacrebleu}{SacreBLEU} \cite{post2018call} library, which uses the commonly used \href{https://github.com/moses-smt/mosesdecoder/blob/master/scripts/generic/mteval-v13a.pl}{Moses 13a Tokenizer} (standard in Machine Translation) instead of the Penn Treebank Tokenizer we used for Audio and Image Captioning. The reported metrics are Leave-One-Out BLEU (Loo-BLEU) and Pairwise-BLEU (see Section~4 of \citet{shen2019mixture}), using the \href{https://github.com/facebookresearch/fairseq/blob/main/examples/translation_moe/score.py}{official codebase}. Leave-One-Out BLEU (denoted simply as BLEU in \citep{shen2019mixture}) measures the overall quality of the hypothesis set by computing the corpus-level BLEU for each hypothesis against the reference set (higher is better). Pairwise-BLEU considers only the predicted hypotheses, computing BLEU scores for each pair of candidates (lower indicates greater diversity). We define hereafter formally these two metrics, as done in \cite{shen2019mixture}.

Let $\operatorname{BLEU}\left\{\left(\left[x^1, \cdots, x^R\right], \hat{x}^{k}\right)\right\}_{\{k \in [\![1,K]\!]\}}$ denote the \textit{corpus-level} BLEU when hypothesis $\hat{x}^{k}$ is evaluated against references $x_1, \cdots, x_R$ as references for each example in the test-set and for each $k \in [\![1,K]\!]$. We also define $[x^{-r}] = \{x^1, \dots, x^{r-1}, x^{r+1}, \dots, x^{R}\}$.

\textbf{Leave-One-Out BLEU.} Corresponds to an average of each corpus-level BLEU when using hypothesis $k$ against all possible sets of $M-1$ references 
$$\mathrm{Loo{-}BLEU} = \frac{1}{R} \sum_{r=1}^{R} \operatorname{BLEU}\left\{\left(\left[x^{-r}\right], \hat{x}^k\right)\right\}_{\{k \in [\![1,K]\!]\}}\;.$$

\textbf{Pairwise-BLEU.} Similar to $\mathrm{mBLEU}_{4}$, Pairwise-BLEU considers only the set of candidates as 
$$\mathrm{Pairwise{-}BLEU} = \operatorname{BLEU}\left\{\left(\left[\hat{x}^j\right], \hat{x}^k\right)\right\}_{\{j \in [\![1,K]\!], k \in [\![1,K]\!], j \neq k\}}\;.$$
We used it instead of $\mathrm{mBLEU}_{4}$ for consistency with the machine translation community.

\subsubsection{Qualitative Examples}

We provide some qualitative examples of the predictions on newstest2014. Here, \texttt{LoRA-MCL} uses $\varepsilon=0.05$, $B = 3$ and $K=3$, and \texttt{LoRA-MLE} ($r = 48$, DBS with $B = 6$ and $\lambda = 0.8$).

\textit{Example 1.} Input english sentence: $\{$As Reuters first reported in July, seat layout is exactly what drives the battle between the latest jets.$\}$ References:
\begin{itemize}
\item Wie Reuters im Juli erstmals berichtete, ist das Sitzlayout die treibende Kraft hinter der Auseinandersetzung um die neuen Jets.
\item Wie Reuters im Juli erstmals berichtete, ist die Sitzanordnung genau das, was den Kampf zwischen den neuesten Jets antreibt.
\item Wie Reuters erstmals im Juli berichtete, ist die Sitzanordnung genau das, was den Kampf zwischen den neuesten Jets antreibt.
\item Wie Reuters im Juli berichtete, ist es genau das Thema der Sitzanordnung, das den Kampf zwischen den neuesten Flugzeugmodellen antreibt.
\item Wie Reuters erstmals im Juli berichtet hatte, ist das Sitzkonzept genau der Punkt, der das Ringen zwischen den neuesten Jets beleb.
\item Wie Reuters bereits erstmals im Juli berichtete, ist es genau das Layout der Sitze, was den Konkurrenzkampf zwischen den neuesten Jets anschürt.
\item Wie Reuters erstmals im Juli berichtete, ist es das Sitz @-@ Layout, das den Wettbewerb zwischen den aktuellen Flugzeugen antreib.
\item Wie Reuters zuerst im Januar berichtete, ist der Aufbau der Sitzreihen genau das, was den Kampf zwischen den neuesten Jets antreibt.
\end{itemize}
\textbf{\texttt{LoRA-MLE.}}\\
$\{$Wie Reuters zuerst im Juli berichtete, ist die Sitzanordnung genau das, was den Kampf zwischen den neuesten Jets antreibt .$\}$\\
$\{$Wie Reuters zuerst im Juli berichtete, ist die Sitzanordnung genau das, was den Kampf zwischen den neuesten Jets antreibt.$\}$\\
$\{$Nachdem Reuters im Juli berichtet hatte, ist die Sitzanordnung das, was den Kampf zwischen den neuesten Jets antreibt.$\}$\\

\textbf{\texttt{LoRA-MCL.}}\\
$\{$Als Reuters im Juli berichtete, ist die Sitzanordnung genau das, was den Kampf zwischen den neuesten Jets ausmacht.$\}$\\
$\{$Als Reuters zuerst im Juli berichtete, ist die Sitzanordnung genau das, was den Kampf zwischen den neuesten Jets antreibt.$\}$\\
$\{$Wie Reuters bereits im Juli berichtete, ist die Sitzanordnung das entscheidende Kriterium im Kampf um die neuesten Jets.$\}$\\

\textit{Example 2.} Input english sentence: $\{$``I was vocal about looking into a whole bunch of things, and this was one of them,''Daley told the paper.$\}$ References:
\begin{itemize}
\item``Ich äußerte eine ganze Reihe von Dingen, die man in Erwägung ziehen sollte, und das war eines davon'', erklärte Daley dem Magazin.
Ich habe mich freimütig darüber geäußert eine ganze Reihe an Sachen ausprobieren zu wollen und das war eine davon, teilte Daley der Zeitung mit.
\item``Ich war lautstark dabei, in eine ganze Reihe von Dingen zu schauen, und das war eines von ihnen'', Daley sagte der Zeitung.
\item``Ich äußerte mich lautstark darüber, dass ich mir eine ganze Reihe von Dingen ansehen wollte. Und das war eines davon", sagte Daley der Zeitung..
\item``Ich verkündete lautstark, mir verschiedene Dinge ansehen zu wollen, und dies war eines davon''Daley hat der Zeitung davon erzählt.
\item``Ich war entschlossen, mich mit einer ganzen Reihe von Dingen zu befassen, und dies war eins davon'', sagte Daley der Zeitung.
\item``Ich habe mündlich geäußert, mich mit einer ganzen Menge Dinge befasst zu haben und das hier gehörte dazu'', erzählte Daley der Zeitung.
\item``Ich habe es deutlich gemacht, dass ich eine ganze Reihe von Sachen unter die Lupe nehmen werde, und das war eine von ihnen'', erzählte Daley der Zeitung.
\item``Ich habe mich dazu geäußert, mir mehrere Dinge ansehen zu wollen, und das war eines davon'', sagte Daley der Zeitung.
\item``Ich hatte geäußert, eine ganze Reihe an Dingen zu untersuchen, und dies war eine von ihnen'', sagte Daley der Zeitung.
\end{itemize}
\textbf{\texttt{LoRA-MLE.}}\\
$\{$„Ich war laut darüber, dass wir uns mit einer ganzen Reihe von Dingen beschäftigen sollten, und dies war eine davon“, sagte Daley der Zeitung.$\}$\\
$\{$„Ich war laut darüber, dass wir uns mit einer ganzen Reihe von Dingen beschäftigen sollten, und das war eine davon“, sagte Daley der Zeitung.$\}$\\
$\{$``Ich habe mich laut darüber geäußert, dass ich mich in eine ganze Reihe von Dingen einlesen wollte, und das war eine davon'', sagte Daley der Zeitung.$\}$\\

\textbf{\texttt{LoRA-MCL.}}\\
$\{$``Ich war laut darüber, dass ich mich in eine ganze Reihe von Dingen einlesen wollte, und das war eine davon,'' Daley sagte der Zeitung.$\}$\\
$\{$``Ich war laut über die Untersuchung einer ganzen Reihe von Dingen, und dies war eines von ihnen'', sagte Daley der Zeitung.$\}$\\
$\{$Daley sagte der Zeitung: ``Ich habe mich laut darüber geäußert, eine ganze Reihe von Dingen zu untersuchen, und das war eine von ihnen.''$\}$

\subsection{Computation details}
\label{apx:computation}
We run the experiments mostly on H100 NVIDIA GPUs with $80$ GB of RAM. Training and inference were launched on a single GPU for the Audio Captioning experiments, and up to $8$ GPUs for the Image Captioning experiments. The total computing resources used for this project, including failed experiments, amount to approximately $25{,}000$ GPU hours.

\subsection{Use of Large Language Models}

We used LLM assistants in the preparation of this work. They helped to polish the writing (improving clarity, grammar, and style without altering the content) and to serve as a coding assistant (visualization, debugging).

\end{document}